\documentclass{article}

\usepackage{microtype}
\usepackage{booktabs} 
\usepackage{hyperref}
\usepackage{tabularx}

\usepackage[accepted]{icml2025}
\usepackage{amsmath}
\usepackage{amssymb}
\usepackage{mathtools}
\usepackage{amsthm}
\usepackage{graphicx}
\usepackage{caption}
\usepackage{subcaption}
\usepackage{wrapfig}
\usepackage{bm}
\usepackage{bbm}
\usepackage{url}
\usepackage{multirow}
\usepackage{enumitem}
\usepackage{diagbox}
\usepackage{color,colortbl}
\usepackage{xcolor}
\usepackage{makecell}
\usepackage{tcolorbox}
\definecolor{COLOR_MEAN}{HTML}{f0f0f0}

\usepackage[capitalize,noabbrev]{cleveref}

\theoremstyle{plain}

\theoremstyle{definition}

\theoremstyle{remark}

\usepackage[textsize=tiny]{todonotes}

\icmltitlerunning{When Do LLMs Help With Node Classification? A Comprehensive Analysis}

\begin{document}

\twocolumn[
\icmltitle{When Do LLMs Help With Node Classification? A Comprehensive Analysis
}




\begin{icmlauthorlist}
\icmlauthor{Xixi Wu}{cuhk}
\icmlauthor{Yifei Shen}{msra}
\icmlauthor{Fangzhou Ge}{cuhk}
\icmlauthor{Caihua Shan}{msra}
\icmlauthor{Yizhu Jiao}{uiuc}
\icmlauthor{Xiangguo Sun}{cuhk}
\icmlauthor{Hong Cheng}{cuhk}
\end{icmlauthorlist}

\icmlaffiliation{cuhk}{Department of Systems Engineering and Engineering Management, and Shun Hing Institute of Advanced Engineering, The Chinese University of Hong Kong}
\icmlaffiliation{msra}{Microsoft Research Asia}
\icmlaffiliation{uiuc}{University of Illinois Urbana-Champaign}

\icmlcorrespondingauthor{Yifei Shen}{\texttt{yifeishen@microsoft.com}}
\icmlcorrespondingauthor{Hong Cheng}{\texttt{hcheng@se.cuhk.edu.hk}}

\icmlkeywords{Large Language Models, Graph Neural Networks, Node Classification}

\vskip 0.3in
]



\printAffiliationsAndNotice{} 

\begin{abstract}
Node classification is a fundamental task in graph analysis, with broad applications across various fields. Recent breakthroughs in Large Language Models (LLMs) have enabled LLM-based approaches for this task. Although many studies demonstrate the impressive performance of LLM-based methods, the lack of clear design guidelines may hinder their practical application.  In this work, we aim to establish such guidelines through a fair and systematic comparison of these algorithms. As a first step, we developed LLMNodeBed, a comprehensive codebase and testbed for node classification using LLMs. It includes 10 homophilic datasets, 4 heterophilic datasets, 8 LLM-based algorithms, 8 classic baselines, and 3 learning paradigms. Subsequently, we conducted extensive experiments, training and evaluating over 2,700 models, to determine the key settings (e.g., learning paradigms and homophily) and components (e.g., model size and prompt) that affect performance. Our findings uncover 8 insights, e.g., (1) LLM-based methods can significantly outperform traditional methods in a semi-supervised setting, while the advantage is marginal in a supervised setting; (2) Graph Foundation Models can beat open-source LLMs but still fall short of strong LLMs like GPT-4o in a zero-shot setting. We hope that the release of LLMNodeBed, along with our insights, will facilitate reproducible research and inspire future studies in this field. Codes and datasets are released at \href{https://llmnodebed.github.io/}{\texttt{https://llmnodebed.github.io/}}.

\end{abstract}

\section{Introduction}

Node classification is a fundamental task in graph analysis, with a wide range of applications such as item tagging \cite{Mao2020ItemTF}, user profiling \cite{Yan2021RelationawareHG}, and financial fraud detection \cite{Zhang2022eFraudComAE}. Developing effective algorithms for node classification is crucial, as they can significantly impact commercial success. For instance, US banks lost 6 billion USD to fraudsters in 2016. Therefore, even a marginal improvement in fraud detection accuracy could result in substantial financial savings.

Given its practical importance, node classification has been a long-standing research focus in both academia and industry. The earliest attempts to address this task adopted techniques such as Laplacian regularization \cite{belkin2006manifold}, graph embeddings \cite{yang2016revisiting}, and label propagation \cite{zhu2003semi}. Over the past decade, GNN-based methods have been developed and have quickly become prominent due to their superior performance, as demonstrated by works such as \citet{kipf2017GCN}, \citet{velickovic2018GAT}, and \citet{hamilton2017SAGE}. Additionally, the incorporation of encoded textual information has been shown to further complement GNNs' node features, enhancing their effectiveness \cite{jin2023patton, zhao2022GLEM}.

\begin{figure*}[!t]
    \centering
    \includegraphics[width=0.86\linewidth]{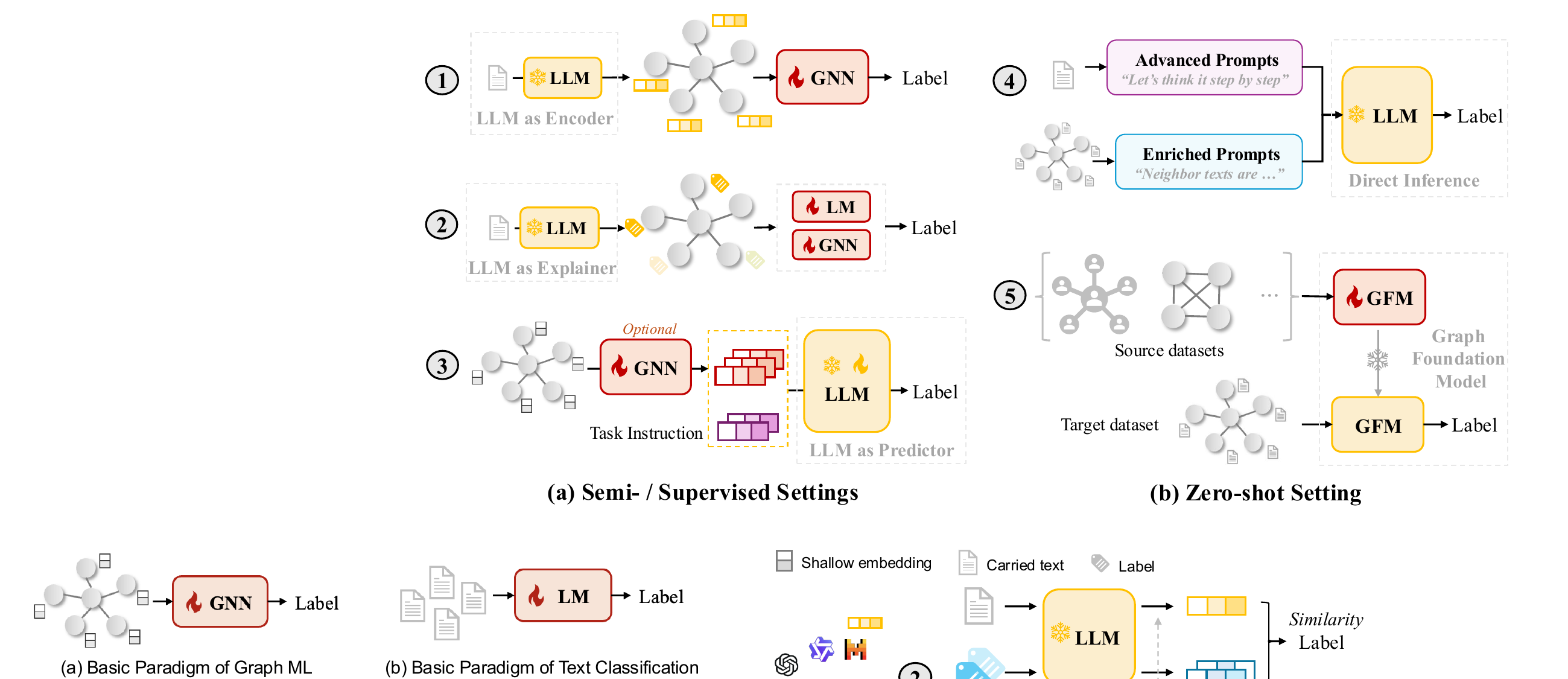}
    \vspace*{-0.1cm}
    \caption{\textbf{Illustrations of LLM-based node classification algorithms under supervised and zero-shot settings.}}
    \vspace*{-0.2cm}
    \label{fig:llm_role}
\end{figure*}

Inspired by the recent success of LLMs, there has been a surge of interest in leveraging LLMs for node classification \cite{li2023survey}. LLMs, pre-trained on extensive text corpora, possess context-aware knowledge and superior semantic comprehension, overcoming the limitations of the non-contextualized shallow embeddings used by traditional GNNs. Typically, supervised methods fall into three categories: Encoder, Explainer, and Predictor. In the Encoder paradigm, LLMs employ their vast parameters to encode nodes' textual information, producing more expressive features that surpass shallow embeddings \cite{Zhu2024ENGINE}. The Explainer approach utilizes LLMs' generative capability to enhance node attributes and the task descriptions with a more detailed text \cite{chen2024exploring, he2023TAPE}. This generated text augments the nodes' original information, thereby enriching their attributes. Lastly, the Predictor role involves LLMs integrating graph context through graph encoders, enabling direct text-based predictions  \cite{chen23llaga,tang2023graphgpt,chai2023graphllm,Huang2024GraphAdapter}. For zero-shot learning with LLMs, methods can be categorized into two types: Direct Inference and Graph Foundation Models (GFMs). Direct Inference involves guiding LLMs to directly perform classification tasks via crafted prompts \cite{Huang2023CanLE}. In contrast, GFMs entail pre-training on extensive graph corpora before applying the model to target graphs, thereby equipping the model with specialized graph intelligence \cite{li2024zerog}. An illustration of these methods is shown in Figure \ref{fig:llm_role}. 

Despite tremendous efforts and promising results, the design principles for LLM-based node classification algorithms remain elusive. Given the significant training and inference costs associated with LLMs, practitioners may opt to deploy these algorithms only when they provide substantial performance enhancements compared to costs. This study, therefore, seeks to identify \textbf{(1) the most suitable settings for each algorithm category, and (2) the scenarios where LLMs surpass traditional LMs such as BERT}. While recent work like GLBench \cite{Li2024GLBench} has evaluated various methods using consistent data splits in semi-supervised and zero-shot settings, differences in backbone architectures and implementation codebases still hinder fair comparisons and rigorous conclusions. To address these limitations, we introduce a new benchmark that further standardizes backbones and codebases. Additionally, we extend GLBench by incorporating 3 E-Commerce datasets, and 4 heterophilic datasets, while also expanding the evaluation settings. Specifically, we assess the impact of supervision signals (e.g., supervised, semi-supervised), different language model backbones (e.g., RoBERTa, Mistral, LLaMA, Qwen, GPT-4o), and various prompt types (e.g., CoT, ToT, ReAct). These enhancements enable a more detailed and reliable analysis of LLM-based node classification methods. In summary, our contributions to the field of LLMs for graph analysis are as follows:

\begin{itemize}
    \item \textbf{A Testbed:} We release LLMNodeBed, a PyG-based testbed designed to facilitate reproducible and rigorous research in LLM-based node classification algorithms. The initial release includes 14 datasets, 8 LLM-based algorithms, and 3 learning configurations. LLMNodeBed allows for easy addition of new algorithms or datasets, and a single command to run all experiments, and to automatically generate all tables included in this work.
    
    \item \textbf{Comprehensive Experiments:} By training and evaluating over 2,700 models, we analyzed how the learning paradigm, homophily, language model type and size, and prompt design impact the performance of each algorithm category.
    
    \item \textbf{Insights and Tips:} Detailed experiments were conducted to analyze each influencing factor. We identified the settings where each algorithm category performs best and the key components for achieving this performance. Our work provides intuitive explanations, practical tips, and insights about the strengths and limitations of each algorithm category.
\end{itemize}

\begin{table*}[!t]
    \centering
    \caption{\textbf{Statistics of supported datasets in LLMNodeBed.}}
    \vspace*{-8pt}
    \resizebox{\linewidth}{!}{
    \begin{tabular}{c|cccc|c|cc|ccc|cccc}
       \toprule
       \rowcolor{COLOR_MEAN}  &  \multicolumn{4}{c|}{\textbf{Academic}} & \textbf{Web Link} & \multicolumn{2}{c|}{\textbf{Social}} & \multicolumn{3}{c|}{\textbf{E-Commerce}} & \multicolumn{4}{c}{\textbf{Heterophilic}}  \\ 
       \rowcolor{COLOR_MEAN} \multirow{-2}{*}{\textbf{Statistics}} & Cora & Citeseer & Pubmed & arXiv & WikiCS & Instagram & Reddit & Books & Photo & Computer & Cornell & Texas & Wisconsin & Washington \\ \midrule
        \# Classes & 7 & 6 & 3 & 40 & 10 & 2 & 2 & 12 & 12 & 10 & 5 & 5 & 5 & 5 \\
       \# Nodes & 2,708 & 3,186 & 19,717 & 169,343 & 11,701 & 11,339 & 33,434 & 41,551 & 48,362 & 87,229 & 191 & 187 & 265 & 229 \\ 
       \# Edges & 5,429 & 4,277 & 44,338 & 1,166,243 & 215,863 & 144,010 & 198,448 & 358,574 & 500,928 & 721,081 & 292 & 310 & 510 & 394 \\ 
       Avg. \# Token & 183.4 & 210.0  & 446.5 & 239.8 & 629.9 & 56.2 & 197.3 & 337.0 & 201.5 & 123.1 & 594.6 & 453.2 & 639.1 & 469.0 \\
       Homophily (\%) & 82.52 & 72.93 & 79.24 & 63.53 & 68.67 & 63.35 & 55.52 & 78.05 & 78.50 & 85.28 & 11.55 & 6.69 & 16.27 & 17.07 \\
       \bottomrule
    \end{tabular}
    }
    \label{tab:dataset}
\end{table*}

\begin{figure}[!t]
    \centering
    \includegraphics[width=0.98\linewidth]{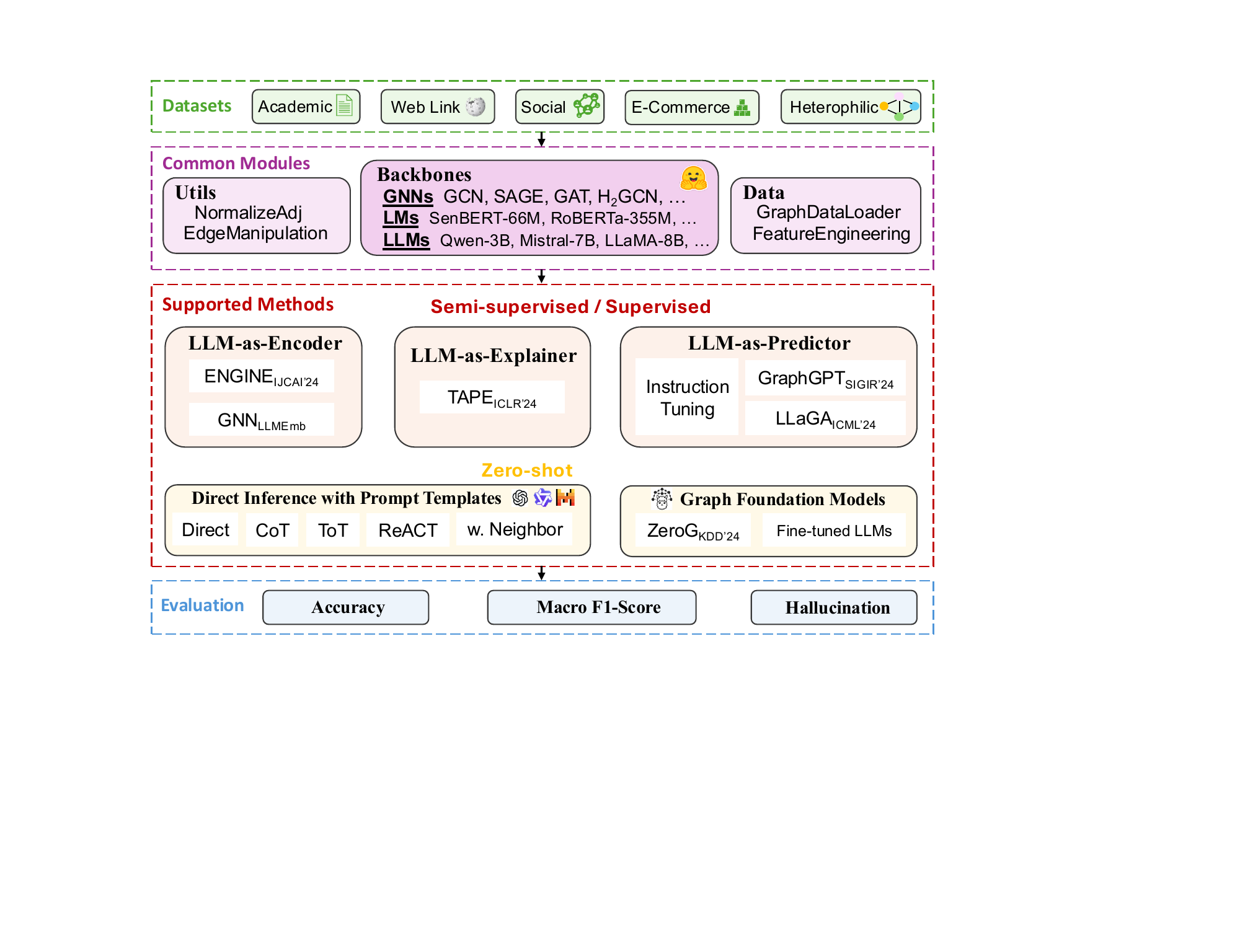}
    \vspace*{-0.1cm}
    \caption{\textbf{Overview of LLMNodeBed.}}
    \vspace*{-10pt}
    \label{fig:system_implementation}
\end{figure}

\section{Preliminaries on Node Classification}
To leverage the language abilities of LLMs, we study the node classification task within the context of text-attributed graphs (TAGs) \cite{ma2021deep}. TAGs are represented as $\mathcal{G} = (\mathcal{V}, \mathcal{E}, \mathcal{S})$, where $\mathcal{V}$ represents the set of nodes, $\mathcal{E}$ the set of edges, and $\mathcal{S}$ the collection of textual descriptions associated with each node $v \in \mathcal{V}$. Some of the nodes are associated with labels, represented as $\mathcal{V}_l \subset \mathcal{V}$. The remaining nodes do not have labels, and are denoted as $\mathcal{V}_u$. The goal of node classification is to train a neural network based on the graph $\mathcal{G}$ and the labels of $\mathcal{V}_l$, which can predict the labels of unlabeled nodes in $\mathcal{V}_u$.

Traditionally, the textual attributes of nodes can be encoded into shallow embeddings as $\bm{X} = [\bm{x}_1, \ldots, \bm{x}_{|\mathcal{V}|}] \in \mathbb{R}^{|\mathcal{V}| \times d}$ using naive methods like bag-of-words or TF-IDF \cite{Salton1988TermWeightingAI}, where $d$ represents the dimensionality of the embeddings. Such transformation is adopted in most GNN papers. Instead, the input for LLM-based approaches is the raw text and one may expect that the pre-trained knowledge in LLMs can improve performance.

\section{LLMNodeBed: A Testbed for LLM-based Node Classification}
In this section, we present the datasets, baselines, and learning paradigms within LLMNodeBed (Figure \ref{fig:system_implementation}).

\subsection{Datasets}

To provide guidelines for applying algorithms across diverse real-world applications, the selection of datasets in LLMNodeBed considers several key factors: (1) \textbf{Multi-domain Diversity} to reflect different contexts, (2) \textbf{Varying Scales} to examine algorithm scalability and the associated costs of leveraging LLMs, and (3) \textbf{Diverse Levels of Homophily} to understand its impact on performance. Therefore, LLMNodeBed comprises $14$ datasets spanning various domains. These datasets vary significantly in scale, ranging from thousands of nodes to millions of edges, and exhibit differing levels of homophily. Such diversity in domain, scale, and homophily enables the assessment of algorithms across a wide range of contexts. 

For datasets where raw text has been preprocessed into vector embeddings using bag-of-words or TF-IDF techniques, we utilize collected versions including Cora and Pubmed \cite{he2023TAPE}, Citeseer \cite{chen2024exploring}, and WikiCS \cite{liu2023one}. The remaining datasets already include text attributes in their official releases, including arXiv \cite{hu2020open}, Instagram and Reddit \cite{Huang2024GraphAdapter}, Books, Computer, and Photo \cite{yan2023comprehensive}, Cornell, Texas, Wisconsin, and Washington \cite{wang2025modelgeneralization}. Detailed statistics and information for these datasets are provided in Table \ref{tab:dataset} and Table \ref{tab:dataset_detail} in the Appendix.

\subsection{Baselines} 
The initial release of LLMNodeBed includes eight LLM-based baseline algorithms alongside classic methods. We selected these LLM-based algorithms based on three key criteria: (1) \textbf{Diverse Roles of LLMs} to thoroughly evaluate their effectiveness, (2) \textbf{Straightforward Design} to facilitate clear comparisons by avoiding complex and intertwined architectures, and (3) \textbf{Representativeness} to ensure benchmark relevance by including widely recognized methods. Therefore, the LLM-based baselines include:

\textbf{LLM-as-Encoder: }We include \textbf{ENGINE} \cite{Zhu2024ENGINE} and introduce \textbf{GNN\textsubscript{LLMEmb}}. ENGINE aggregates hidden embeddings from each LLM layer to create comprehensive node representations. In contrast, GNN\textsubscript{LLMEmb} initializes node embeddings using the LLM's last hidden layer before feeding them into GNNs for classification.

\textbf{LLM-as-Explainer: }We select \textbf{TAPE} \cite{he2023TAPE}, a representative LLM-as-Explainer method. TAPE prompts the LLM to think over nodes by generating predictions along with explanations, thereby enriching the node's text attributes and enhancing classification performance.

Both Encoder and Explainer methods require processing the entire dataset, either by encoding each node's text or generating explanations for nodes, which introduces additional processing time before the actual model training begins.

\textbf{LLM-as-Predictor: }We select \textbf{GraphGPT} \cite{tang2023graphgpt}, \textbf{LLaGA} \cite{chen23llaga}, and implement \textbf{LLM Instruction Tuning} (\textbf{LLM\textsubscript{IT}}). GraphGPT employs a multi-stage pre-training and instruction tuning process to classify nodes based on both text and graph context. \textbf{LLaGA} integrates tokenized task instructions and graph context into LLMs to generate predictions.  We implement LLM\textsubscript{IT} to evaluate whether LLMs alone can function as effective predictors. This involves fine-tuning the LLM using task prompts and ground-truth labels formatted as $\langle \texttt{Question}, \texttt{Answer} \rangle$ pairs. 

\textbf{LLM Direct Inference: } This category refers to LLMs generating prediction labels directly from a node's text without additional training or labels. We employ two types of prompt templates: (1) \textbf{Advanced Prompts} that improve LLMs' reasoning abilities such as Chain-of-Thought (CoT) \cite{Wei2022ChainOT} and Tree-of-Thought (ToT) \cite{yao2023tree}, and (2) \textbf{Enriched Prompts} that incorporate neighboring node information to provide structural context.

\textbf{GFMs: } GFMs are foundation models trained on large-scale source graph datasets to acquire general classification knowledge, which can then be seamlessly applied to target graphs. We include \textbf{ZeroG} \cite{li2024zerog} as a representative GFM due to its superior performance in zero-shot settings. Additionally, LLM-as-Predictor methods trained with extensive graph corpora are also considered within this category for zero-shot applications.

Besides LLM-based methods, LLMNodeBed also integrates classic algorithms, including \textbf{MLPs}, \textbf{GNNs}, and \textbf{LMs}. MLPs generate predicted label matrices from node embeddings, while GNNs combine shallow embeddings with graph structures for label prediction. LMs process a node's text through hidden layers and use a classification head to produce label distributions. Furthermore, we include \textbf{GLEM} \cite{zhao2022GLEM}, a hybrid LM+GNN method that combines graph and text modalities. Further discussions of existing algorithms are provided in Appendix \ref{sec:related_works}.

Prompt templates for LLM-as-Predictor and Direct Inference are listed in Appendix \ref{sec:predictor_prompt} and \ref{sec:zeroshot_prompt}, respectively. Appendix \ref{sec:hyperparam} describes the implementation details, backbone selections, and hyperparameter search spaces for all algorithms. Additionally, we highlight the distinctions of LLMNodeBed in Appendix \ref{sec:distinct_llmnodebed}.

\begin{table*}[!t]
    \centering
    \caption{\textbf{Performance comparison under semi-supervised and supervised settings with Accuracy ($\%$) reported.} \\\small{The \colorbox{orange!25}{\textbf{best}} and \colorbox{orange!10}{second-best} results are highlighted. Results of Macro-F1 are shown in Table \ref{tab:mainexp_f1} in the Appendix. LLM\textsubscript{IT} on the arXiv dataset requires 30+ hours per run, preventing multiple executions.}} 
    \vspace*{-8pt}
    \resizebox{\linewidth}{!}{
    \begin{tabular}{cc|cccccccccc}
      \toprule
     \rowcolor{COLOR_MEAN} \multicolumn{2}{c|}{\textbf{Semi-supervised}}  & \textbf{Cora} & \textbf{Citeseer} & \textbf{Pubmed} & \textbf{WikiCS} & \textbf{Instagram} & \textbf{Reddit} & \textbf{Books} & \textbf{Photo} & \textbf{Computer} & \textbf{Avg.} \\ \midrule
       \multirow{6}{*}{\textbf{Classic}} & {GCN\tiny{ShallowEmb}} & 82.30$_{\pm \text{0.19}}$ & 70.55$_{\pm \text{0.32}}$ & 78.94$_{\pm \text{0.27}}$ & 79.86$_{\pm \text{0.19}}$ & 63.50$_{\pm \text{0.11}}$ & 61.44$_{\pm \text{0.38}}$ & 68.79$_{\pm \text{0.46}}$ & 69.25$_{\pm \text{0.81}}$ & 71.44$_{\pm \text{1.19}}$ & 71.79 \\ 
        & SAGE\tiny{ShallowEmb} & 82.27$_{\pm \text{0.37}}$ & 69.56$_{\pm \text{0.43}}$ & 77.88$_{\pm \text{0.44}}$ & 79.67$_{\pm \text{0.25}}$ & 63.57$_{\pm \text{0.10}}$ & 56.65$_{\pm \text{0.33}}$ & 72.01$_{\pm \text{0.33}}$& 78.50$_{\pm \text{0.15}}$ & 81.43$_{\pm \text{0.27}}$ & 73.50 \\ 
         & {GAT\tiny{ShallowEmb}} & 81.30$_{\pm \text{0.67}}$ & 69.94$_{\pm \text{0.74}}$ & 78.49$_{\pm \text{0.70}}$ & 79.99$_{\pm \text{0.65}}$ & 63.56$_{\pm \text{0.04}}$ & 60.60$_{\pm \text{1.17}}$ & 74.35$_{\pm \text{0.35}}$ & 80.40$_{\pm \text{0.45}}$ & 83.39$_{\pm \text{0.22}}$ & 74.67 \\ 
         & SenBERT-66M & 66.66$_{\pm \text{1.42}}$ & 60.52$_{\pm \text{1.62}}$ & 36.04$_{\pm \text{2.92}}$ & 77.77$_{\pm \text{0.75}}$ & 59.00$_{\pm \text{1.17}}$ & 56.05$_{\pm \text{0.41}}$ & 83.68$_{\pm \text{0.19}}$ & 73.89$_{\pm \text{0.31}}$ & 70.76$_{\pm \text{0.15}}$ & 64.93 \\
         & {RoBERTa-355M} & 72.24$_{\pm \text{1.14}}$ & 66.68$_{\pm \text{2.03}}$ & 42.32$_{\pm \text{1.56}}$ & 76.81$_{\pm \text{1.04}}$ & 63.52$_{\pm \text{0.44}}$ & 59.27$_{\pm \text{0.34}}$ & \cellcolor{orange!10} 84.62$_{\pm \text{0.16}}$ & 74.79$_{\pm \text{1.13}}$ & 72.31$_{\pm \text{0.37}}$ & 68.06 \\
         & GLEM & 81.30$_{\pm \text{0.88}}$ & 68.80$_{\pm \text{2.46}}$ & \cellcolor{orange!25} \textbf{81.70$_{\pm \text{1.07}}$} & 76.43$_{\pm \text{0.55}}$ & 60.25$_{\pm \text{3.66}}$ & 55.13$_{\pm \text{1.41}}$ & 83.28$_{\pm \text{0.39}}$ & 76.93$_{\pm \text{0.49}}$ & 80.46$_{\pm \text{1.45}}$  & 73.81  \\ 
         \midrule
         
        \multirow{2}{*}{\textbf{Encoder}} 
       & $\text{GCN}_{\text{LLMEmb}}$ & 83.33$_{\pm \text{0.75}}$ & 71.39$_{\pm \text{0.90}}$ & 78.71$_{\pm \text{0.45}}$ & \cellcolor{orange!10} 80.94$_{\pm \text{0.16}}$ & \cellcolor{orange!25} \textbf{67.49$_{\pm \text{0.43}}$} & 68.65$_{\pm \text{0.75}}$ &  83.03$_{\pm \text{0.34}}$ & \cellcolor{orange!10} 84.84$_{\pm \text{0.47}}$ & \cellcolor{orange!10} 88.22$_{\pm \text{0.16}}$ & \cellcolor{orange!25} \textbf{78.51} \\ 
       & ENGINE & \cellcolor{orange!25} \textbf{84.22$_{\pm \text{0.46}}$} & \cellcolor{orange!25} \textbf{72.14$_{\pm \text{0.74}}$}
        & 77.84$_{\pm \text{0.27}}$ & 80.94$_{\pm \text{0.19}}$ & \cellcolor{orange!10} 67.14$_{\pm \text{0.46}}$ & \cellcolor{orange!10} 69.67$_{\pm \text{0.16}}$ & 82.89$_{\pm \text{0.14}}$ & 84.33$_{\pm \text{0.57}}$ & 86.42$_{\pm \text{0.23}}$ & 78.40  \\ \midrule
        
       \textbf{Explainer} & TAPE &  \cellcolor{orange!10} 84.04$_{\pm \text{0.24}}$ & \cellcolor{orange!10} 71.87$_{\pm \text{0.35}}$ & 78.61$_{\pm \text{1.23}}$ & \cellcolor{orange!25} \textbf{81.94$_{\pm \text{0.16}}$} & 66.07$_{\pm \text{0.10}}$ & 62.43$_{\pm \text{0.47}}$ & \cellcolor{orange!25} \textbf{84.92$_{\pm \text{0.26}}$} & \cellcolor{orange!25} \textbf{86.46$_{\pm \text{0.12}}$} & \cellcolor{orange!25} \textbf{89.52$_{\pm \text{0.04}}$} & \cellcolor{orange!10} 78.43  \\  \midrule
       
      \multirow{3}{*}{\textbf{Predictor}} & $\text{LLM}_{\text{IT}}$  &  67.00$_{\pm \text{0.16}}$ & 54.26$_{\pm \text{0.22}}$ & \cellcolor{orange!10} \textbf{80.99$_{\pm \text{0.43}}$} & 75.02$_{\pm \text{0.16}}$ & 41.83$_{\pm \text{0.47}}$ & 54.09$_{\pm \text{1.02}}$ & 80.92$_{\pm \text{1.38}}$ & 71.28$_{\pm \text{1.81}}$ & 66.99$_{\pm \text{2.02}}$ & 65.76 \\ 
       & GraphGPT & 64.72$_{\pm \text{1.50}}$ & 64.58$_{\pm \text{1.55}}$ & 70.34$_{\pm \text{2.27}}$ & 70.71$_{\pm \text{0.37}}$ & 62.88$_{\pm \text{2.14}}$ & 58.25$_{\pm \text{0.37}}$ & 81.13$_{\pm \text{1.52}}$ & 77.48$_{\pm \text{0.78}}$ & 80.10$_{\pm \text{0.76}}$ & 70.02 \\ 
       & LLaGA & 78.94$_{\pm \text{1.14}}$ & 62.61$_{\pm \text{3.63}}$ & 65.91$_{\pm \text{2.09}}$ & 76.47$_{\pm \text{2.20}}$ & 65.84$_{\pm \text{0.72}}$ &  \cellcolor{orange!25} \textbf{70.10$_{\pm \text{0.38}}$} & 83.47$_{\pm \text{0.45}}$ & 84.44$_{\pm \text{0.90}}$ & 87.82$_{\pm \text{0.53}}$  & 75.07 \\
       \bottomrule
    \end{tabular}
    }

   \vspace*{5pt}
    \resizebox{\linewidth}{!}{
    \begin{tabular}{cc|ccccccccccc}
      \toprule
      \rowcolor{COLOR_MEAN} \multicolumn{2}{c|}{\textbf{Supervised}} & \textbf{Cora} & \textbf{Citeseer} & \textbf{Pubmed} & \textbf{arXiv} & \textbf{WikiCS} & \textbf{Instagram} & \textbf{Reddit} & \textbf{Books} & \textbf{Photo} & \textbf{Computer} & \textbf{Avg.} \\ \midrule

     \multirow{6}{*}{\textbf{Classic}} & {GCN\tiny{ShallowEmb}} & 87.41$_{\pm \text{2.08}}$ & 75.74$_{\pm \text{1.20}}$ & 89.01$_{\pm \text{0.59}}$ & 71.39$_{\pm \text{0.28}}$ & 83.67$_{\pm \text{0.63}}$ & 63.94$_{\pm \text{0.61}}$ & 65.07$_{\pm \text{0.38}}$ & 76.94$_{\pm \text{0.26}}$ & 73.34$_{\pm \text{1.34}}$ & 77.16$_{\pm \text{3.80}}$ & 76.37 \\ 
     & {SAGE\tiny{ShallowEmb}} & 87.44$_{\pm \text{1.74}}$ & 74.96$_{\pm \text{1.20}}$ & 90.47$_{\pm \text{0.25}}$ & 71.21$_{\pm \text{0.18}}$ & 84.86$_{\pm \text{0.91}}$ & 64.14$_{\pm \text{0.47}}$ & 61.52$_{\pm \text{0.60}}$ & 79.40$_{\pm \text{0.45}}$ & 84.59$_{\pm \text{0.32}}$ & 87.77$_{\pm \text{0.34}}$ & 78.64 \\

     & {GAT\tiny{ShallowEmb}} & 86.68$_{\pm \text{1.12}}$ & 73.73$_{\pm \text{0.94}}$ & 88.25$_{\pm \text{0.47}}$ & 71.57$_{\pm \text{0.25}}$ & 83.94$_{\pm \text{0.61}}$ & 64.93$_{\pm \text{0.75}}$ & 64.16$_{\pm \text{1.05}}$ & 80.61$_{\pm \text{0.49}}$ & 84.84$_{\pm \text{0.69}}$ & 88.32$_{\pm \text{0.24}}$ & 78.70 \\ 
     & SenBERT-66M & 79.61$_{\pm \text{1.40}}$ & 74.06$_{\pm \text{1.26}}$ & \cellcolor{orange!10} 94.47$_{\pm \text{0.33}}$ & 72.66$_{\pm \text{0.24}}$ & 86.51$_{\pm \text{0.86}}$ & 60.11$_{\pm \text{0.93}}$ & 58.70$_{\pm \text{0.54}}$ & \cellcolor{orange!10} 85.99$_{\pm \text{0.58}}$ & 77.72$_{\pm \text{0.35}}$ & 74.22$_{\pm \text{0.21}}$ & 76.40 \\ 
     & {RoBERTa-355M} & 83.17$_{\pm \text{0.84}}$ & 75.90$_{\pm \text{1.69}}$ & \cellcolor{orange!25} \textbf{94.84$_{\pm \text{0.06}}$} & 74.12$_{\pm \text{0.12}}$ & \cellcolor{orange!25}\textbf{87.47$_{\pm \text{0.83}}$} & 63.75$_{\pm \text{1.13}}$ & 60.61$_{\pm \text{0.24}}$ & 
      \cellcolor{orange!25} \textbf{86.65$_{\pm \text{0.38}}$} & 79.45$_{\pm \text{0.37}}$ & 75.76$_{\pm \text{0.30}}$ & 78.17 \\
      
     & GLEM & 86.81$_{\pm \text{1.19}}$ & 73.24$_{\pm \text{1.55}}$ & 93.98$_{\pm \text{0.32}}$ & 73.55$_{\pm \text{0.22}}$ & 79.81$_{\pm \text{0.45}}$ & 67.39$_{\pm \text{1.73}}$ & 53.11$_{\pm \text{2.96}}$ & 83.98$_{\pm \text{0.97}}$ & 78.16$_{\pm \text{0.45}}$ & 81.63$_{\pm \text{0.46}}$ & 77.17 \\ 
      \midrule
      
      \multirow{2}{*}{\textbf{Encoder}} & $\text{GCN}_{\text{LLMEmb}}$ & 
      \cellcolor{orange!25} \textbf{88.15$_{\pm \text{1.79}}$} & \cellcolor{orange!10} 76.45$_{\pm \text{1.19}}$ & 88.38$_{\pm \text{0.68}}$ & 74.39$_{\pm \text{0.31}}$ & 84.78$_{\pm \text{0.86}}$ & 68.27$_{\pm \text{0.45}}$ & 70.65$_{\pm \text{0.75}}$ & 84.23$_{\pm \text{0.20}}$ & 86.07$_{\pm \text{0.20}}$ & 89.52$_{\pm \text{0.31}}$ & 81.09 \\ 
      & ENGINE & 87.00$_{\pm \text{1.60}}$ & 75.82$_{\pm \text{1.52}}$ & 90.08$_{\pm \text{0.16}}$ & 74.69$_{\pm \text{0.36}}$ & 85.44$_{\pm \text{0.53}}$ & \cellcolor{orange!10} 68.87$_{\pm \text{0.25}}$ & \cellcolor{orange!25} \textbf{71.21$_{\pm \text{0.77}}$} & 84.09$_{\pm \text{0.09}}$ & 86.98$_{\pm \text{0.06}}$ & 89.05$_{\pm \text{0.13}}$ & 81.32 \\  \midrule
      \textbf{Explainer} & TAPE &  \cellcolor{orange!10} 88.05$_{\pm \text{1.76}}$ & 76.45$_{\pm \text{1.60}}$ & 93.00$_{\pm \text{0.13}}$ & 74.96$_{\pm \text{0.14}}$ & \cellcolor{orange!10} 87.11$_{\pm \text{0.66}}$ & 68.11$_{\pm \text{0.54}}$ & 66.22$_{\pm \text{0.83}}$ &  85.95$_{\pm \text{0.59}}$ & \cellcolor{orange!25} \textbf{87.72$_{\pm \text{0.28}}$} & \cellcolor{orange!25} \textbf{90.46$_{\pm \text{0.18}}$} & \cellcolor{orange!25}\textbf{81.80} \\  \midrule
      \multirow{3}{*}{\textbf{Predictor}} & $\text{LLM}_{\text{IT}}$ & 71.93$_{\pm \text{1.47}}$ & 60.97$_{\pm \text{3.97}}$ &  94.16$_{\pm \text{0.19}}$ & \cellcolor{orange!25} \textbf{76.08} & 80.61$_{\pm \text{0.47}}$ & 44.20$_{\pm \text{3.06}}$ & 58.30$_{\pm \text{0.48}}$ & 84.80$_{\pm \text{0.13}}$ & 78.27$_{\pm \text{0.54}}$ & 74.51$_{\pm \text{0.53}}$ & 72.38 \\ 
      & GraphGPT & 82.29$_{\pm \text{0.26}}$ & 74.67$_{\pm \text{1.15}}$ & 93.54$_{\pm \text{0.22}}$ & \cellcolor{orange!10} 75.15$_{\pm \text{0.14}}$ & 82.54$_{\pm \text{0.23}}$ & 67.00$_{\pm \text{1.22}}$  & 60.72$_{\pm \text{1.47}}$ & 85.38$_{\pm \text{0.72}}$ & 84.46$_{\pm \text{0.36}}$ & 86.78$_{\pm \text{1.14}}$ & 79.25  \\ 
      & LLaGA & 87.55$_{\pm \text{1.15}}$ & \cellcolor{orange!25} \textbf{76.73$_{\pm \text{1.70}}$} & 90.28$_{\pm \text{0.91}}$ & 74.49$_{\pm \text{0.23}}$ & 84.03$_{\pm \text{1.10}}$ & \cellcolor{orange!25} \textbf{69.16$_{\pm \text{0.72}}$} & \cellcolor{orange!10} 71.06$_{\pm \text{0.38}}$ & 85.56$_{\pm \text{0.30}}$ & \cellcolor{orange!10} 87.62$_{\pm \text{0.30}}$ & \cellcolor{orange!10} 90.41$_{\pm \text{0.12}}$ & \cellcolor{orange!10}81.69 \\ \bottomrule
    \end{tabular}
    }

    \label{tab:mainexp}
\end{table*}

\subsection{Learning Paradigms}

We evaluate the baselines under three learning configurations: Semi-supervised, Supervised, and Zero-shot. These configurations are defined as follows:

\begin{itemize}
    \item \textbf{Semi-supervised Learning:} A small subset of nodes $\mathcal{V}_l \subseteq \mathcal{V}$ with known labels $\mathcal{Y}_l$ is provided. This setting assesses the model's ability to effectively utilize limited labeled data, reflecting real-world scenarios where labeling is scarce. For experimental datasets, we adopt the official splits designed for semi-supervised settings to ensure standardized evaluation.
   
    \item \textbf{Supervised Learning:} A larger subset of nodes $\mathcal{V}_l$ with known labels is provided, assessing the model's performance with abundant supervision. Specifically, we use a 60\% training, 20\% validation, and 20\% testing split for most datasets. This consistent split facilitates fair comparisons across baselines. Detailed data splits are provided in Table \ref{tab:dataset_detail} in the Appendix.
    
    \item \textbf{Zero-shot Learning:} No labeled data is provided for training. The model predicts labels solely based on node textual descriptions and the graph structure, assessing its ability to generalize to new, unseen data. For test samples, we follow existing literature \cite{Zhu2024GraphCLIPET} by selecting one smaller dataset from each domain and using 20\% of its nodes as test samples.
\end{itemize}

\section{Comparisons among Algorithm Categories}
In this section, we present empirical results among algorithm categories, along with key insights derived from them.

\subsection{Semi-supervised and Supervised Performance}
\textbf{Settings: }To ensure a fair comparison of baseline algorithms, all methods are implemented with consistent components: GCN \cite{kipf2017GCN} for GNNs, RoBERTa-355M \cite{Liu2019roberta} for LMs, and Mistral-7B \cite{Jiang2023Mistral7B} for LLM components where applicable. This uniformity guarantees that performance differences are attributable to model designs rather than underlying architectures. Each experiment was conducted over \textbf{4 runs}. Based on the results of Accuracy (Table \ref{tab:mainexp}) and Macro-F1 (Table \ref{tab:mainexp_f1} in Appendix), we summarize the following takeaways:

\textbf{Takeaway 1: Appropriately incorporating LLMs consistently improves the performance.} According to the table, the best performance is often achieved by LLM-based methods compared to classic methods. It suggests that using LLM to exploit the textual information is useful. This conclusion extends to \textbf{large-scale graphs} as well, e.g., ogbn-product dataset with millions of nodes. Detailed results and analysis for this dataset are provided in Appendix \ref{sec:exp_on_product}.

\textbf{Takeaway 2: LLM-based methods provide greater improvements in semi-supervised settings than in supervised settings.} By comparing the tables, we observe that performance gains are more significant in semi-supervised scenarios. From an information-theoretic perspective, the node classification task with cross-entropy loss aims to maximize the mutual information between the graph and the provided labels, denoted as $I(\mathcal{G}; \mathcal{Y}_{l})$. If we consider graph $\mathcal{G}$ as a joint distribution of node attributes $\bm{X}$ and structure $\mathcal{E}$, we have: 
\begin{equation}
\label{eq:mutual_info}
    \begin{aligned}
    I(\mathcal{G}; \mathcal{Y}_{l}) = I(\bm{X}, \mathcal{E}; \mathcal{Y}_{l}) = I(\mathcal{E}; \mathcal{Y}_{l}) + I(\bm{X}; \mathcal{Y}_{l} | \mathcal{E}).
\end{aligned}
\end{equation}

The first term represents the information encoded in the graph structure, utilized by classic GNNs, while the second term represents information from node features, leveraged by LLMs. In semi-supervised settings, the mutual information between structure and labels is relatively low, allowing LLMs to contribute more significantly to performance.

\textbf{Takeaway 3: LLM-as-Explainer methods are highly effective when labels heavily depend on text.} TAPE achieves top or runner-up performance on academic and web link datasets like Cora and WikiCS, where structural information is less relevant to labels \cite{zhang2021graphless}. However, TAPE struggles with social networks that require deeper structural understanding, such as predicting popular users (high-degree nodes) on Reddit.

\textbf{Takeaway 4: LLM-as-Encoder methods balance computational cost and accuracy effectively.} LLM-as-Encoder methods perform satisfactorily across all datasets. Further experiments in Section \ref{exp:encoder_comp} reveal that \textbf{LLM-as-Encoder methods are more effective than their LM counterparts when graphs are less informative.} Regarding cost-effectiveness, LLM-as-Explainer should generate long explanatory text, which is far more time-consuming than encoding texts in LLM-as-Encoder (see Appendix \ref{sec:detail_cost}). Therefore, LLM-as-Encoder methods strike a balance between computational efficiency and accuracy.

\textbf{Takeaway 5: LLM-as-Predictor methods are more effective when labeled data is abundant.} In supervised scenarios, LLM-as-Predictor methods enhance performance across most datasets. Especially, the LLaGA method achieves superior results among 5 of 10 datasets. Conversely, in semi-supervised settings, LLM-as-Predictor methods exhibit unstable performance, evidenced by low Macro-F1 scores, and imbalanced output distributions (detailed discussion in Appendix \ref{sec:llm_bias_pred}). These findings indicate that LLM-as-Predictor methods are most effective when ample supervision is available, with LLaGA being an especially strong choice. Furthermore, within the predictor methods, LLM Instruction Tuning typically falls behind the other two methods and incurs substantial time costs (Table \ref{tab:timecost_supervised} in the Appendix). This shows that standalone LLMs are weak predictors and incorporating graph context is essential for achieving satisfactory performance.

\begin{table*}[!t]
    \centering
    \caption{\textbf{Performance comparison under zero-shot setting with Accuracy ($\%$) and Macro-F1 ($\%$) reported.} \\ \small{Numbers in brackets represent the dataset's homophily ratio (\%). Results of other LLMs are shown in Table \ref{tab:zeroshot_supple} in Appendix.}}
    \vspace*{-8pt}
    \resizebox{0.92\linewidth}{!}{
    \begin{tabular}{cc|cc|cc|cc|cc|cc}
       \toprule
       \rowcolor{COLOR_MEAN} & &  \multicolumn{2}{c|}{\textbf{Cora} (82.52)} & \multicolumn{2}{c|}{\textbf{WikiCS} (68.67)} & \multicolumn{2}{c|}{\textbf{Instagram} (63.35)}  & \multicolumn{2}{c|}{\textbf{Photo} (78.50)} & \multicolumn{2}{c}{\textbf{Avg.}}  \\ 
       \rowcolor{COLOR_MEAN} \multirow{-2}{*}{\textbf{Type \& LLM}} &  \multirow{-2}{*}{\textbf{Method}} & Acc & Macro-F1 & Acc & Macro-F1 & Acc & Macro-F1 &  Acc & Macro-F1  & Acc & Macro-F1 \\  \midrule
       \multirow{6}{*}{\begin{tabular}{c}
            \textbf{LLM} \\ GPT-4o
       \end{tabular}} & Direct  & 68.08  & 69.25  & 68.59  & 63.21  & 44.53  &  42.77 & 63.99  & 61.09 & 61.30 & 59.08 \\
       & CoT  & 68.89 &	69.86 &	70.75 &	\textbf{66.23} &	\textbf{47.87} &	\textbf{47.57} &	61.61 &	60.62 & 62.28 & 61.07 \\ 
       & ToT  & 68.29 &	69.13 &	70.78 &	65.69 &	44.16 &	42.68 &	60.84 &	59.16 & 61.02 & 59.16  \\
       & ReAct & 68.21 & 69.28  & 69.45 &	66.03 & 44.49 &	43.16 &	63.63 &	60.82 & 61.44 & 59.82 \\ 
       & w. Neighbor & 70.30 & 71.44 & 69.69 & 64.51 & 42.42 & 39.79 & 69.93 & 68.55 & 63.09 & 61.07 \\ 
       & w. Summary & \textbf{71.40} &	\textbf{72.13}  &	\textbf{70.90} &	65.42 &	45.02 &	44.62 &	\textbf{72.63} & \textbf{70.84} & \textbf{64.99} & \textbf{63.25} \\ \midrule


       \multirow{6}{*}{\begin{tabular}{c}
            \textbf{LLM} \\ LLaMA-8B 
       \end{tabular}} & Direct & 62.64 & 63.02 & 56.77 & 53.04 & 37.58 & 29.70 & 41.23 & 44.26 & 49.56 & 47.50 \\ 
       & CoT & 62.04 & 62.61 & 58.88 & 56.00 & 42.00 & 39.06 & 44.22 & 47.13 & 51.78 & 51.20 \\ 
       & ToT & 34.06 & 33.30 & 40.35 & 41.15 & \textbf{45.33} & \textbf{45.27} & 31.31 & 34.00 & 37.76 & 38.43 \\ 
       & ReAct & 36.55 & 38.04 & 22.40 & 25.76 & 44.67 & 44.42 & 27.03 & 28.96  & 32.66 & 34.30 \\ 
       & w. Neighbor & 64.55 & 64.41 & 59.43 & 54.16 & 36.98 & 28.32 & 45.49 & 50.44 & 51.61 & 49.33 \\ 
       & w. Summary & \textbf{64.69} & \textbf{64.62} & \textbf{62.69} & \textbf{56.40}  & 37.59 & 30.91 & \textbf{48.11} & \textbf{52.20} & \textbf{53.27} & \textbf{51.03} \\ \midrule

       \multirow{3}{*}{\begin{tabular}{c}
            \textbf{GFM} 
       \end{tabular}} & ZeroG & \textbf{62.55} & \textbf{57.56}  & \textbf{62.71} & \textbf{57.87} & \textbf{50.71} & \textbf{50.43} & 46.27 & \textbf{51.52} & \textbf{55.56} & \textbf{54.35} \\ 
       & LLM\textsubscript{IT} & 52.58 & 51.89 & 60.83 & 53.59 & 41.58 & 26.26 & \textbf{49.23} & 44.85  & 51.06 & 44.15 \\ 
       & LLaGA & 18.82 & 8.49 & 8.20 & 8.29 & 47.93 & 47.70 & 39.18 & 4.71 & 28.53 & 17.30 \\ 
       \bottomrule
    \end{tabular}
    }
    \label{tab:zeroshot}
\end{table*}

\subsection{Zero-shot Performance}
\textbf{Settings:} For LLM Direct Inference, we utilize both closed-source and open-source LLMs, including GPT-4o \cite{Achiam2023GPT4TR}, DeepSeek-V3 \cite{deepseekai2024deepseekv3}, LLaMA3.1-8B \cite{llama3modelcard}, and Mistral-7B. The prompt templates include \textbf{Direct}, \textbf{CoT}, \textbf{ToT}, and \textbf{ReAct} \cite{Yao2022ReActSR}. Additionally, we incorporate a node's neighboring information into extended prompts, referred to as ``\textbf{w. Neighbor}'', and have the LLMs first reason over neighbors to generate a summary that facilitates the subsequent classification task, referred to as ``\textbf{w. Summary}''. Prompt templates are listed in Appendix \ref{sec:zeroshot_prompt}. Besides, we assess the transferability of GFMs by evaluating \textbf{ZeroG} \cite{li2024zerog}, \textbf{LLM Instruction Tuning}, and \textbf{LLaGA}. GFMs are applied following the intra-domain manner: each model is pre-trained on a larger dataset within the same domain (e.g., arXiv from the academic domain) before being evaluated on the target dataset. Results for Accuracy and Macro-F1 are shown in Table \ref{tab:zeroshot} and Table \ref{tab:zeroshot_supple} in the Appendix, where we have the following takeaways:

\textbf{Takeaway 6: GFMs can outperform open-source LLMs but still fall short of strong LLMs like GPT-4o.} ZeroG outperforms LLaMA-8B in most cases, achieving up to a 6\% average improvement in accuracy. However, it still falls short of GPT-4o and DeepSeek-V3. Among GFMs, LLaGA performs poorly because it uses a projector to align the source graph's tokens with LLM input tokens. This projector may be dataset-specific, leading to reduced performance on different datasets, as also observed in \citet{Zhu2024GraphCLIPET}. These findings highlight the need for further research to improve the generalization of GFMs to match the performance of more powerful LLMs.

\textbf{Takeaway 7: LLM direct inference can be improved by appropriately incorporating structural information.} Our results reveal that advanced prompt templates such as CoT, ToT, and ReAct, offer only minor performance improvements. Specifically, models like LLaMA exhibit limited instruction-following abilities, often producing unexpected and over-length outputs when encountering complex prompts such as ReAct. This makes parsing classification results challenging and leads to suboptimal performance. The advanced prompts are generally designed for broad reasoning tasks and lack graph- or classification-specific knowledge, thereby limiting their benefits for the node classification task. In contrast, enriched prompts that incorporate structural information, i.e., ``w. Neighbor'' and ``w. Summary'', demonstrate performance enhancements across LLMs. The performance gains are particularly evident on homophilic datasets such as Cora and Photo (3\%-10\%), where neighboring nodes are likely to share the same labels as the central node. High homophily means that information from neighboring nodes provides crucial clues about a central node's label, thereby improving classification performance. Among these enriched prompts, ``\textbf{w. Summary}'' is especially effective as it not only provides structural context but also leverages the self-reflection abilities of LLMs to further utilize structural information.

\begin{table*}[!t]
    \centering
    \caption{\textbf{Comparison of LLM- and LM-as-Encoder with Accuracy ($\%$) reported under semi-supervised setting.}\\ \small{The \colorbox{blue!10}{\textbf{best encoder}} within each method on a dataset is highlighted. Results in supervised settings are shown in Table \ref{tab:encoder_comp_fullysupervised} in Appendix.}}
    \vspace*{-6pt}
    \resizebox{\linewidth}{!}{
    \begin{tabular}{cc|ccccccccc}
      \toprule
     \rowcolor{COLOR_MEAN} \textbf{Method}  & \textbf{Encoder}  & \textbf{Computer} & \textbf{Cora} & \textbf{Pubmed} & \textbf{Photo} & \textbf{Books} & \textbf{Citeseer} & \textbf{WikiCS} & \textbf{Instagram} & \textbf{Reddit} \\ \midrule
     \multicolumn{2}{c}{Homophily Ratio (\%)} & 85.28 & 82.52 & 79.24 & 78.50 & 78.05 & 72.93 & {68.67} & {63.35} & {55.52} \\ \midrule

      \multirow{4}{*}{MLP} & SenBERT & \cellcolor{blue!10}\textbf{69.57$_{\pm \text{0.18}}$} & 64.61$_{\pm \text{1.34}}$ & 74.67$_{\pm \text{0.63}}$ & 72.28$_{\pm \text{0.36}}$ & \cellcolor{blue!10}\textbf{81.93$_{\pm \text{0.08}}$} & 66.83$_{\pm \text{0.58}}$ & 71.48$_{\pm \text{0.33}}$ & 64.98$_{\pm \text{0.38}}$ & 57.23$_{\pm \text{0.51}}$ \\
    &  RoBERTa & 69.42$_{\pm \text{0.10}}$ & 73.84$_{\pm \text{0.55}}$ & 73.21$_{\pm \text{0.78}}$ & 72.95$_{\pm \text{0.34}}$ & 81.71$_{\pm \text{0.14}}$ & \cellcolor{blue!10}\textbf{70.59$_{\pm \text{0.31}}$} & 75.82$_{\pm \text{0.13}}$ & 66.39$_{\pm \text{0.24}}$ & 59.66$_{\pm \text{0.53}}$  \\
    &  Qwen-3B & 67.54$_{\pm \text{0.29}}$ & \cellcolor{blue!10}\textbf{74.03$_{\pm \text{0.57}}$} & 75.30$_{\pm \text{0.72}}$ & 72.72$_{\pm \text{0.23}}$ & 81.60$_{\pm \text{0.53}}$ & 68.26$_{\pm \text{0.79}}$ & 78.64$_{\pm \text{0.37}}$ & 66.53$_{\pm \text{0.37}}$ & 60.49$_{\pm \text{0.17}}$  \\ 
    &  Mistral-7B & 69.37$_{\pm \text{0.28}}$ & 73.90$_{\pm \text{0.59}}$ & \cellcolor{blue!10}\textbf{75.70$_{\pm \text{1.00}}$} & \cellcolor{blue!10} \textbf{74.16$_{\pm \text{0.24}}$} & 81.91$_{\pm \text{0.25}}$ & 69.66$_{\pm \text{0.38}}$ & \cellcolor{blue!10}\textbf{79.56$_{\pm \text{0.41}}$} & \cellcolor{blue!10}\textbf{66.68$_{\pm \text{0.24}}$} & \cellcolor{blue!10}\textbf{61.91$_{\pm \text{0.21}}$}  \\ \midrule
    
     \multirow{4}{*}{GCN} & SenBERT & \cellcolor{blue!10}\textbf{88.92$_{\pm \text{0.19}}$} & 81.76$_{\pm \text{0.75}}$ & 78.24$_{\pm \text{0.66}}$ & 85.18$_{\pm \text{0.16}}$ & \cellcolor{blue!10} \textbf{83.47$_{\pm \text{0.20}}$} & 70.97$_{\pm \text{0.81}}$ & 80.41$_{\pm \text{0.18}}$ & 65.78$_{\pm \text{0.14}}$ & 64.97$_{\pm \text{0.82}}$ \\ 
    &  RoBERTa & 88.90$_{\pm \text{0.14}}$ & \cellcolor{blue!10}\textbf{84.56$_{\pm \text{0.41}}$} & 78.08$_{\pm \text{0.52}}$ & \cellcolor{blue!10} \textbf{85.19$_{\pm \text{0.17}}$} & 83.22$_{\pm \text{0.29}}$ & \cellcolor{blue!10}\textbf{73.52$_{\pm \text{0.58}}$} & 80.97$_{\pm \text{0.22}}$ & 66.64$_{\pm \text{0.21}}$ & 65.69$_{\pm \text{1.01}}$ \\ 
    &  Qwen-3B & 87.55$_{\pm \text{0.14}}$ & 83.62$_{\pm \text{0.41}}$ & 78.50$_{\pm \text{0.80}}$ & 84.26$_{\pm \text{0.33}}$ & 82.83$_{\pm \text{0.24}}$ & 71.50$_{\pm \text{0.92}}$ & \cellcolor{blue!10} \textbf{81.02$_{\pm \text{0.33}}$} & 66.69$_{\pm \text{0.59}}$ & \cellcolor{blue!10} \textbf{69.40$_{\pm \text{0.56}}$} \\ 
    &  Mistral-7B & 88.22$_{\pm \text{0.16}}$ & 83.33$_{\pm \text{0.75}}$ & \cellcolor{blue!10} \textbf{78.71$_{\pm \text{0.45}}$} & 84.84$_{\pm \text{0.47}}$ & 83.03$_{\pm \text{0.34}}$ & 71.39$_{\pm \text{0.90}}$ & 80.94$_{\pm \text{0.16}}$ & \cellcolor{blue!10}\textbf{67.49$_{\pm \text{0.43}}$} & 68.65$_{\pm \text{0.75}}$ \\ \midrule

     \multirow{4}{*}{SAGE} & SenBERT & \cellcolor{blue!10}\textbf{89.08$_{\pm \text{0.06}}$} & 80.45$_{\pm \text{0.79}}$ & 77.29$_{\pm \text{0.45}}$ & 85.54$_{\pm \text{0.16}}$ & \cellcolor{blue!10} \textbf{83.93$_{\pm \text{0.17}}$} & 69.42$_{\pm \text{1.42}}$ & 80.02$_{\pm \text{0.24}}$ & 65.34$_{\pm \text{0.44}}$ & 61.65$_{\pm \text{0.17}}$ \\ 
    & RoBERTa & 88.97$_{\pm \text{0.09}}$ & \cellcolor{blue!10} \textbf{84.06$_{\pm \text{0.52}}$} & 75.82$_{\pm \text{0.59}}$ & \cellcolor{blue!10} \textbf{85.57$_{\pm \text{0.17}}$} & 83.74$_{\pm \text{0.22}}$ & \cellcolor{blue!10}\textbf{72.58$_{\pm \text{0.45}}$} & 80.77$_{\pm \text{0.29}}$ & 66.53$_{\pm \text{0.50}}$ & 63.65$_{\pm \text{0.32}}$ \\ 
    & Qwen-3B & 86.24$_{\pm \text{0.32}}$ & 83.31$_{\pm \text{0.63}}$ & 76.76$_{\pm \text{0.35}}$ & 84.28$_{\pm \text{0.42}}$ & 82.84$_{\pm \text{0.31}}$ & 71.11$_{\pm \text{0.98}}$ & 80.85$_{\pm \text{0.22}}$ & 66.73$_{\pm \text{0.37}}$ & 63.82$_{\pm \text{0.38}}$ \\ 
    & Mistral-7B & 88.48$_{\pm \text{0.20}}$ & 82.73$_{\pm \text{0.99}}$ & \cellcolor{blue!10} \textbf{77.64$_{\pm \text{1.73}}$} & 85.50$_{\pm \text{0.20}}$ & 83.32$_{\pm \text{0.16}}$ & 71.42$_{\pm \text{0.47}}$ & \cellcolor{blue!10} \textbf{81.47$_{\pm \text{0.32}}$} & \cellcolor{blue!10}\textbf{67.44$_{\pm \text{0.06}}$} &  \cellcolor{blue!10} \textbf{65.02$_{\pm \text{0.13}}$} \\ 
    \bottomrule
    \end{tabular}
    }
    \label{tab:encoder_comp}
\end{table*}

\subsection{Computational Cost Analysis}
 
\textbf{Time Cost:}  We evaluate the training and inference times of various methods in supervised settings. Detailed training times are provided in Table \ref{tab:timecost_supervised} in the Appendix, while inference times are presented in Table \ref{tab:inference_cost}. All measurements were conducted on a single NVIDIA H100-80G GPU to ensure consistency.

Based on the results, we can conclude that classic methods are highly efficient, with GNNs typically converging within seconds (e.g., 5.2 seconds for GCN\textsubscript{ShallowEmb} on Pubmed) and LMs fine-tuning completed within minutes. In contrast, LLM-as-Explainer approaches are the most time-consuming (e.g., 5.9 hours for TAPE on Pubmed) because they require generating explanatory text for each node and subsequently processing this augmented text through both an LM and a GNN. This three-stage computational process significantly extends the overall computation time. LLM-as-Encoder methods are the most efficient among LLM-based approaches (e.g., 13.4 minutes for GCN$_{\text{LLMEmb}}$ on Pubmed), utilizing LLMs solely for feature encoding, which allows GNN training to remain efficient and complete within minutes. Although LLM-as-Predictor methods are more efficient than Explainer approaches, they still require hours for effective model training. Among predictor methods, LLaGA is the most efficient (e.g., 25.6 minutes on Pubmed) as it encodes both the node's textual and structural information into embeddings instead of processing raw text.

During inference, a significant efficiency gap remains between LLM-based and classic methods. Classic methods can complete the entire inference process for thousands of cases within milliseconds, making them suitable for industrial deployments that demand real-time responses. In contrast, LLM-based methods are limited to processing one case within the same timeframe, highlighting the urgent need to improve their efficiency.

\textbf{Memory Cost:} We also analyze the memory costs of different methods during both the training and inference stages to assess their resource usage. We select three datasets, Cora, WikiCS, and arXiv, spanning a range from 2,708 to 169,343 nodes. The results, presented in Table \ref{tab:memory_usage} in the Appendix, were measured using a single H100-80GB GPU to ensure consistency and comparability. The results show that LLM-as-Encoder methods demonstrate memory costs comparable to traditional GNNs, making them practical choices for deployment. In contrast, methods involving fine-tuning an LM or LLM-as-Predictor methods are significantly more memory-intensive, largely due to the high parameterization of language models. These findings underscore the importance of addressing efficiency challenges to enable the practical deployment of LLM-based methods.

\section{Fine-grained Analysis Within Each Category}
In this section, we present empirical results within each category. For LLM-as-Encoder, we explore the conditions under which LLMs outperform traditional LMs. Additionally, we examine how key components (e.g., model type and size) influence the effectiveness of LLM-as-Explainer and LLM-as-Predictor. 

\subsection{LLM-as-Encoder: Compared with LMs}\label{exp:encoder_comp}
\begin{table*}[!t]
    \centering
     \caption{\textbf{Comparison of LLM- and LM-as-Encoder with Accuracy ($\%$) reported on heterophilic graphs.} The \colorbox{blue!10}{\textbf{best encoder}} within each method on a dataset is highlighted. }
     \vspace*{-8pt}
    \resizebox{0.98\linewidth}{!}{

     \begin{tabular}{cc|cccc|cccc}
       \toprule
       \rowcolor{COLOR_MEAN} & & \multicolumn{4}{c|}{\textbf{Semi-supervised}} & \multicolumn{4}{c}{\textbf{Supervised}} \\ 
       \rowcolor{COLOR_MEAN} \multirow{-2}{*}{\textbf{Method}}  & \multirow{-2}{*}{ \textbf{Encoder}} & \textbf{Cornell} & \textbf{Texas} &  \textbf{Wisconsin} & \textbf{Washington} & \textbf{Cornell} & \textbf{Texas} & \textbf{Wisconsin} & \textbf{Washington}  \\ \midrule
       \multicolumn{2}{c|}{Homophily Ratio (\%)} & 11.55 & 6.69 & 16.27 & 17.07 & 11.55 & 6.69 & 16.27 & 17.07 \\ \midrule 

       \multirow{4}{*}{MLP} & SenBERT & 50.59$_{\pm \text{3.14}}$ & 56.67$_{\pm \text{2.15}}$ & 71.98$_{\pm \text{1.59}}$ & 63.26$_{\pm \text{2.89}}$  & 66.15$_{\pm \text{1.92}}$ & 76.32$_{\pm \text{3.72}}$ & 81.51$_{\pm \text{7.00}}$ & 70.44$_{\pm \text{8.65}}$ \\
       & RoBERTa & 59.08$_{\pm \text{2.57}}$ & 67.47$_{\pm \text{1.29}}$ & 73.87$_{\pm \text{1.62}}$ & 65.43$_{\pm \text{3.44}}$ & 66.67$_{\pm \text{8.88}}$ & 74.21$_{\pm \text{6.09}}$ & 80.00$_{\pm \text{9.88}}$ & 76.96$_{\pm \text{7.48}}$ \\ 
       & Qwen-3B & 57.78$_{\pm \text{3.24}}$ & 76.27$_{\pm \text{1.61}}$ & 82.36$_{\pm \text{1.62}}$ & 
       \cellcolor{blue!10} \textbf{75.11$_{\pm \text{1.92}}$} & 77.95$_{\pm \text{4.76}}$ & 88.95$_{\pm \text{3.07}}$ & 88.68$_{\pm \text{6.64}}$ & 83.48$_{\pm \text{1.74}}$ \\ 
       & Mistral-7B & \cellcolor{blue!10} \textbf{59.87$_{\pm \text{6.72}}$} & \cellcolor{blue!10} \textbf{76.27$_{\pm \text{1.08}}$} & \cellcolor{blue!10} \textbf{83.30$_{\pm \text{1.42}}$} & 74.24$_{\pm \text{0.88}}$ & \cellcolor{blue!10} \textbf{78.46$_{\pm \text{4.17}}$} & \cellcolor{blue!10} \textbf{90.53$_{\pm \text{3.16}}$} & \cellcolor{blue!10} \textbf{89.43$_{\pm \text{5.15}}$} & \cellcolor{blue!10} \textbf{83.91$_{\pm \text{5.60}}$} \\ \midrule 

       \multirow{4}{*}{GCN} & SenBERT & 46.80$_{\pm \text{2.13}}$ & 54.93$_{\pm \text{0.68}}$ & 58.30$_{\pm \text{2.56}}$ & 52.61$_{\pm \text{1.35}}$ & 50.77$_{\pm \text{10.18}}$ & 59.47$_{\pm \text{5.16}}$ & 61.13$_{\pm \text{8.65}}$ & 61.30$_{\pm \text{1.62}}$  \\ 
       & RoBERTa & 47.06$_{\pm \text{2.19}}$ & 55.20$_{\pm \text{2.78}}$ & 54.91$_{\pm \text{3.40}}$ & 54.89$_{\pm \text{1.50}}$ & 51.79$_{\pm \text{7.68}}$ & 58.42$_{\pm \text{7.33}}$ & 59.24$_{\pm \text{8.82}}$ & 61.31$_{\pm \text{5.39}}$ \\ 
       & Qwen-3B & 53.59$_{\pm \text{2.07}}$ & 56.80$_{\pm \text{4.29}}$ & \cellcolor{blue!10} \textbf{63.02$_{\pm \text{2.16}}$} & \cellcolor{blue!10} \textbf{64.56$_{\pm \text{4.06}}$} & 58.46$_{\pm \text{10.56}}$ & 64.74$_{\pm \text{7.37}}$ & \cellcolor{blue!10} \textbf{65.28$_{\pm \text{6.82}}$} & \cellcolor{blue!10} \textbf{67.83$_{\pm \text{3.74}}$} \\ 
       & Mistral-7B  & \cellcolor{blue!10} \textbf{54.64$_{\pm \text{1.52}}$} & \cellcolor{blue!10} \textbf{58.67$_{\pm \text{3.60}}$} & 62.08$_{\pm \text{2.61}}$ & 61.52$_{\pm \text{3.61}}$ & \cellcolor{blue!10} \textbf{59.49$_{\pm \text{6.96}}$} & \cellcolor{blue!10} \textbf{65.79$_{\pm \text{6.66}}$} & 64.90$_{\pm \text{5.67}}$ & 66.96$_{\pm \text{4.84}}$ \\  \midrule 

       \multirow{4}{*}{SAGE} & SenBERT & 52.55$_{\pm \text{1.58}}$ & 61.73$_{\pm \text{1.37}}$ & 70.47$_{\pm \text{1.75}}$ & 65.54$_{\pm \text{2.44}}$ & 68.72$_{\pm \text{4.97}}$ & 80.00$_{\pm \text{5.91}}$ & 83.02$_{\pm \text{6.31}}$ & 76.96$_{\pm \text{4.88}}$ \\ 
       & RoBERTa & 55.55$_{\pm \text{3.44}}$ & 64.26$_{\pm \text{6.26}}$ & 73.59$_{\pm \text{2.72}}$ & 66.08$_{\pm \text{1.60}}$ & 70.26$_{\pm \text{8.37}}$ & 80.53$_{\pm \text{2.68}}$ & 81.89$_{\pm \text{7.42}}$ & 74.35$_{\pm \text{7.95}}$ \\ 
       & Qwen-3B & \cellcolor{blue!10} \textbf{57.13$_{\pm \text{2.29}}$} & \cellcolor{blue!10} \textbf{78.53$_{\pm \text{1.76}}$} & 83.21$_{\pm \text{1.39}}$ & 72.18$_{\pm \text{3.66}}$ & 74.87$_{\pm \text{2.99}}$ & 89.47$_{\pm \text{1.67}}$ & \cellcolor{blue!10} \textbf{91.32$_{\pm \text{2.82}}$} & \cellcolor{blue!10} \textbf{83.48$_{\pm \text{3.25}}$}  \\ 
       & Mistral-7B & 56.86$_{\pm \text{1.37}}$ & 76.53$_{\pm \text{2.40}}$ & \cellcolor{blue!10} \textbf{83.96$_{\pm \text{1.55}}$} & \cellcolor{blue!10} \textbf{73.91$_{\pm \text{0.97}}$} & \cellcolor{blue!10} \textbf{77.44$_{\pm \text{2.99}}$} & \cellcolor{blue!10} \textbf{91.05$_{\pm \text{2.69}}$} & 89.44$_{\pm \text{4.24}}$ & 81.74$_{\pm \text{4.48}}$ \\  \midrule 

       \multirow{4}{*}{H$_2$GCN} & SenBERT & 56.34$_{\pm \text{1.67}}$ & 66.67$_{\pm \text{2.95}}$ & 73.40$_{\pm \text{1.68}}$ & 70.55$_{\pm \text{4.95}}$ & 73.85$_{\pm \text{7.14}}$ & 84.21$_{\pm \text{4.40}}$ & 86.42$_{\pm \text{6.01}}$ & 77.83$_{\pm \text{7.20}}$ \\ 
       & RoBERTa & 60.00$_{\pm \text{3.54}}$ & 68.13$_{\pm \text{2.93}}$ & 75.66$_{\pm \text{2.12}}$ & 71.52$_{\pm \text{1.22}}$ & 74.87$_{\pm \text{7.68}}$ & 83.16$_{\pm \text{6.14}}$ & 84.53$_{\pm \text{9.04}}$ & 79.13$_{\pm \text{5.43}}$ \\ 
       & Qwen-3B & \cellcolor{blue!10} \textbf{61.57$_{\pm \text{3.89}}$} & \cellcolor{blue!10} \textbf{80.13$_{\pm \text{6.45}}$} & \cellcolor{blue!10} \textbf{84.53$_{\pm \text{0.70}}$} & \cellcolor{blue!10} \textbf{74.67$_{\pm \text{1.77}}$} & \cellcolor{blue!10} \textbf{76.41$_{\pm \text{2.99}}$} & \cellcolor{blue!10} \textbf{92.11$_{\pm \text{2.88}}$} & \cellcolor{blue!10} \textbf{89.81$_{\pm \text{3.29}}$} & 85.22$_{\pm \text{3.99}}$ \\ 
       & Mistral-7B & 59.22$_{\pm \text{4.54}}$ & 72.93$_{\pm \text{8.21}}$ & 81.89$_{\pm \text{1.51}}$ & 68.59$_{\pm \text{4.46}}$ & 75.89$_{\pm \text{3.84}}$ & 89.47$_{\pm \text{3.72}}$ & 89.43$_{\pm \text{5.42}}$ & \cellcolor{blue!10} \textbf{86.09$_{\pm \text{3.25}}$} \\  
       \bottomrule
    \end{tabular}
    
    }
    \label{tab:encoder_comp_heterophilic}
\end{table*}

\begin{table*}[!t]
    \centering
     \caption{\textbf{Performance ($\%$) of TAPE with different LLM backbones under semi-supervised setting}.}
     \vspace*{-8pt}
    \resizebox{\linewidth}{!}{
    \begin{tabular}{cc|cccccccccc}
       \toprule
      \rowcolor{COLOR_MEAN} \textbf{Metrics} & \textbf{LLM}  &  \textbf{Cora} & \textbf{Citeseer} & \textbf{Pubmed} & \textbf{WikiCS} & \textbf{Instagram} & \textbf{Reddit} & \textbf{Books} & \textbf{Photo} & \textbf{Computer} & \textbf{Avg.} \\ \midrule
      \multirow{2}{*}{\textbf{Acc}} & Mistral  &  84.04$_{\pm \text{0.24}}$ & 71.87$_{\pm \text{0.35}}$ & 78.61 $_{\pm \text{1.23}}$ & \textbf{81.94$_{\pm \text{0.16}}$} & 66.07$_{\pm \text{0.10}}$ & \textbf{62.43$_{\pm \text{0.47}}$} & 84.92$_{\pm \text{0.26}}$ & 86.46$_{\pm \text{0.12}}$ & 89.52$_{\pm \text{0.04}}$ & 78.43 \\  
      & GPT-4o & \textbf{84.30$_{\pm \text{0.36}}$} & \textbf{73.75$_{\pm \text{0.67}}$} & \textbf{82.70$_{\pm \text{1.78}}$} & 81.93$_{\pm \text{0.33}}$ & \textbf{66.25$_{\pm \text{0.38}}$} & 62.22$_{\pm \text{1.24}}$ & \textbf{85.08$_{\pm \text{0.17}}$} & \textbf{86.65$_{\pm \text{0.17}}$} & \textbf{89.62$_{\pm \text{0.13}}$} & \textbf{79.17} \\  \midrule

     \multirow{2}{*}{\textbf{F1}} & Mistral & 81.89$_{\pm \text{0.31}}$ & 66.80$_{\pm \text{0.33}}$ & 78.46$_{\pm \text{1.13}}$ & 80.03$_{\pm \text{0.23}}$ & 50.01$_{\pm \text{1.60}}$ & \textbf{61.23$_{\pm \text{0.69}}$} & 47.12$_{\pm \text{3.26}}$ & 82.31$_{\pm \text{0.19}}$ & \textbf{84.90$_{\pm \text{1.14}}$} & 70.31 \\ 
      & GPT-4o & \textbf{82.62$_{\pm \text{0.60}}$} & \textbf{67.41$_{\pm \text{0.82}}$} & \textbf{82.45$_{\pm \text{1.65}}$} & \textbf{80.27$_{\pm \text{0.34}}$} & \textbf{51.16$_{\pm \text{3.54}}$} & 61.11$_{\pm \text{1.52}}$ & \textbf{47.51$_{\pm \text{2.92}}$} & \textbf{82.54$_{\pm \text{0.18}}$} & 84.28$_{\pm \text{2.98}}$ & \textbf{71.04} \\ 
      
       \bottomrule
    \end{tabular}
    }
    \label{tab:tape_llm_semi}
\end{table*}

\begin{table*}[!t]
    \centering
     \caption{\textbf{Performance ($\%$) of TAPE with different LLM backbones under supervised setting}.}
    \vspace*{-8pt}
    \resizebox{\linewidth}{!}{
    \begin{tabular}{cc|ccccccccccc}
       \toprule
      \rowcolor{COLOR_MEAN} \textbf{Metrics} & \textbf{LLM}  &  \textbf{Cora} & \textbf{Citeseer} & \textbf{Pubmed} & \textbf{arXiv} & \textbf{WikiCS} & \textbf{Instagram} & \textbf{Reddit} & \textbf{Books} & \textbf{Photo} & \textbf{Computer} & \textbf{Avg.} \\ \midrule

       \multirow{2}{*}{\textbf{Acc}} & Mistral  & 88.05$_{\pm \text{1.76}}$ & \textbf{76.45$_{\pm \text{1.60}}$} & 93.00$_{\pm \text{0.13}}$ & 74.96$_{\pm \text{0.14}}$ &
        \textbf{87.11$_{\pm \text{0.66}}$} & \textbf{68.11$_{\pm \text{0.54}}$} & 66.22$_{\pm \text{0.83}}$ & 85.95$_{\pm \text{0.59}}$ & \textbf{87.72$_{\pm \text{0.28}}$} & 90.46$_{\pm \text{0.18}}$ & 81.80 \\  

      & GPT-4o & \textbf{88.24$_{\pm \text{1.23}}$} & 76.41$_{\pm \text{1.38}}$ & \textbf{94.12$_{\pm \text{0.03}}$} & \textbf{75.08$_{\pm \text{0.08}}$} & 87.10$_{\pm \text{0.62}}$ & 67.99$_{\pm \text{0.51}}$ & \textbf{66.33$_{\pm \text{0.89}}$} & \textbf{86.19$_{\pm \text{0.60}}$} & 87.65$_{\pm \text{0.47}}$ & \textbf{90.56$_{\pm \text{0.21}}$} & \textbf{81.97} \\   \midrule
    
      \multirow{2}{*}{\textbf{F1}} & Mistral & 87.21$_{\pm \text{1.60}}$ & \textbf{73.33$_{\pm \text{1.57}}$} & 92.39$_{\pm \text{0.02}}$ & \textbf{57.79$_{\pm \text{0.45}}$} & \textbf{86.03$_{\pm \text{1.14}}$} & \textbf{58.31$_{\pm \text{1.15}}$} & 65.91$_{\pm \text{0.71}}$ & 54.07$_{\pm \text{2.01}}$ & \textbf{83.41$_{\pm \text{0.42}}$} & 86.78$_{\pm \text{0.53}}$ & 74.52 \\ 
      & GPT-4o & \textbf{87.34$_{\pm \text{1.06}}$} & 73.17$_{\pm \text{2.00}}$ & \textbf{93.58$_{\pm \text{0.09}}$} & 57.69$_{\pm \text{0.23}}$ & 85.93$_{\pm \text{1.05}}$ & 57.49$_{\pm \text{1.93}}$ & \textbf{66.09$_{\pm \text{0.80}}$} & \textbf{54.32$_{\pm \text{3.30}}$} & 83.40$_{\pm \text{0.41}}$ & \textbf{86.91$_{\pm \text{0.55}}$} & \textbf{74.59} \\ 
      
       \bottomrule
    \end{tabular}
    }
    \label{tab:tape_llm_fully}
\end{table*}

\textbf{Motivation and Settings: }Both LLMs and small-scale LMs can encode nodes' associated texts. This raises the question: When do LLMs surpass LMs as encoders? To address this, we evaluate various methods using node features derived from LLMs and LMs, observing the resulting performance differences. For LMs, we select SenBERT-66M \cite{reimers-2019-sentence-bert} and RoBERTa-355M \cite{Liu2019roberta}. For LLMs, we choose Qwen2.5-3B \cite{Yang2024Qwen2TR} and Mistral-7B \cite{Jiang2023Mistral7B}. The evaluated datasets include both homophilic and heterophilic graphs to ensure generality. For heterophilic graphs, we use the released datasets collected by \citet{wang2025modelgeneralization}\footnote{\href{https://huggingface.co/datasets/Graph-COM/Text-Attributed-Graphs}{https://huggingface.co/datasets/Graph-COM/Text-Attributed-Graphs}} with processing details explained in Appendix \ref{sec:dataset}. The considered methods include: (1) \textbf{MLP}, which solely utilizes node features as input to predict labels without incorporating any graph information, (2) \textbf{GCN},  (3) \textbf{GraphSAGE}, and (4) \textbf{H$_2$GCN} \cite{zhu2020beyond}, a representative GNN explicitly designed for heterophilic settings, included specifically to accommodate heterophilic graphs. For each method, we initialize node features using embeddings derived from various LM or LLM backbones, ensuring that all other components remain consistent across evaluations.

From the results shown in Table \ref{tab:encoder_comp} (homophilic graphs) and Table \ref{tab:encoder_comp_heterophilic} (heterophilic graphs), we can observe that: \textbf{Takeaway 8: LLM-as-Encoder significantly outperforms LMs in less informative graphs, e.g., heterophilic ones.} 

For homophilic graphs, the performance gap between LLMs and LMs is relatively small and becomes noticeable only in graphs with lower homophily, such as WikiCS and Reddit. For instance, on Reddit, LLM-based encoders achieve 4\% higher accuracy than their LM counterparts. The performance gap is substantially larger for heterophilic graphs, with differences reaching up to 10\% across methods in both semi-supervised and supervised settings. For example, on the Texas dataset, LLM-based encoders achieve a 12\% improvement over their LM counterparts for the GCN, SAGE, and H$_2$GCN methods in the semi-supervised setting. We can also explain these results using mutual information in Equation \eqref{eq:mutual_info}. In homophilic graphs, edges often connect nodes with the same labels. Consequently, the first term in \eqref{eq:mutual_info} dominates, leaving limited room for improved encoders like LLMs to enhance performance. In contrast, this property does not hold for less-informative heterophilic graphs, where neighboring nodes are less likely to share the same labels, making the choice of encoder significantly more impactful.


\subsection{LLM-as-Explainer: Impact of LLM Capabilities}

\textbf{Motivation and Settings: }In the LLM-as-Explainer paradigm, as the language model should generate explanatory texts, the adopted language models should be auto-regressive and the model size should be large. To investigate how the advanced reasoning capabilities of LLMs influence overall performance, we replace the default Mistral-7B model in the TAPE method with the more powerful GPT-4o model, keeping all other components unchanged. 

\begin{table*}[!t]
    \centering
    \caption{\textbf{Accuracy ($\%$) of LLaGA to different LLM backbones under supervised settings}.\\ \small{The best LLM backbone within \colorbox{red!10}{\textbf{each series}} and \colorbox{yellow!20}{\textbf{at similar scales}} is highlighted. Semi-supervised performance is shown in Table \ref{tab:llaga_llm}.}}
    \vspace*{-8pt}
    \resizebox{\linewidth}{!}{
    \begin{tabular}{cc|ccccccccccc}
      \toprule
     \rowcolor{COLOR_MEAN}  & \textbf{LLM} & \textbf{Cora} & \textbf{Citeseer} & \textbf{Pubmed} & \textbf{arXiv} & \textbf{WikiCS} & \textbf{Instagram} & \textbf{Reddit} & \textbf{Books} & \textbf{Photo} & \textbf{Computer} & \textbf{Avg.} \\ \midrule
      \multirow{4}{*}{\rotatebox[origin=c]{90}{\small \begin{tabular}{c}
           \textbf{Same} \\ \textbf{series}
      \end{tabular}}}  & Qwen-3B & 84.91$_{\pm \text{2.19}}$ & 74.83$_{\pm \text{2.46}}$ & 88.61$_{\pm \text{1.24}}$ & 71.82$_{\pm \text{1.37}}$ & 82.23$_{\pm \text{3.14}}$ & 62.49$_{\pm \text{0.98}}$ & 67.96$_{\pm \text{0.90}}$ & 83.56$_{\pm \text{1.86}}$ & \cellcolor{red!10} \textbf{85.20$_{\pm \text{1.63}}$} & \cellcolor{red!10} \textbf{89.37$_{\pm \text{0.29}}$} & 79.10 \\ 
     & Qwen-7B & 85.33$_{\pm \text{1.50}}$ & 70.75$_{\pm \text{5.18}}$ & \cellcolor{red!10} \textbf{90.53$_{\pm \text{0.49}}$} & 71.60$_{\pm \text{1.59}}$ & 82.57$_{\pm \text{1.67}}$ & 63.86$_{\pm \text{2.76}}$ & \cellcolor{red!10} \textbf{68.62$_{\pm \text{0.53}}$} & \cellcolor{red!10} \textbf{84.23$_{\pm \text{0.51}}$} & 83.55$_{\pm \text{1.35}}$ & 87.21$_{\pm \text{1.88}}$  & 78.82 \\ 
    & Qwen-14B & \cellcolor{red!10} \textbf{87.25$_{\pm \text{1.63}}$} & \cellcolor{red!10} \textbf{75.49$_{\pm \text{2.03}}$} & 89.93$_{\pm \text{0.27}}$ & \cellcolor{red!10} \textbf{73.15$_{\pm \text{0.74}}$} & 82.26$_{\pm \text{1.51}}$ & 63.88$_{\pm \text{2.49}}$ & 67.60$_{\pm \text{1.77}}$ & 83.94$_{\pm \text{0.41}}$ & 84.83$_{\pm \text{0.77}}$ & 87.06$_{\pm \text{0.80}}$ & 79.54 \\  
    & Qwen-32B & 85.93$_{\pm \text{0.99}}$ & 75.39$_{\pm \text{1.90}}$ & 89.97$_{\pm \text{0.26}}$ & 72.84$_{\pm \text{0.67}}$ & \cellcolor{red!10} \textbf{83.49$_{\pm \text{0.91}}$} & \cellcolor{red!10} \textbf{64.33$_{\pm \text{1.69}}$} & 68.47$_{\pm \text{0.09}}$ & 84.18$_{\pm \text{0.29}}$ & 84.77$_{\pm \text{0.23}}$ & 88.49$_{\pm \text{0.49}}$ & \cellcolor{red!10}\textbf{79.79} \\ 
    \midrule
    
    \multirow{3}{*}{\rotatebox[origin=c]{90}{\small \begin{tabular}{c}
           \textbf{Similar} \\ \textbf{scales}
      \end{tabular}}}  & Mistral-7B & \cellcolor{yellow!10}\textbf{87.55$_{\pm \text{1.15}}$} &\cellcolor{yellow!10} \textbf{76.73$_{\pm \text{1.70}}$} & 90.28$_{\pm \text{0.91}}$ & \cellcolor{yellow!10}\textbf{74.49$_{\pm \text{0.23}}$} & \cellcolor{yellow!10}\textbf{84.03$_{\pm \text{1.10}}$} & \cellcolor{yellow!10}\textbf{69.16$_{\pm \text{0.72}}$} & \cellcolor{yellow!10}\textbf{71.06$_{\pm \text{0.38}}$} & \cellcolor{yellow!10}\textbf{85.56$_{\pm \text{0.30}}$} &  \cellcolor{yellow!10}\textbf{87.62$_{\pm \text{0.30}}$} &\cellcolor{yellow!10} \textbf{90.41$_{\pm \text{0.12}}$} & \cellcolor{yellow!10} \textbf{81.69} \\ 
     & Qwen-7B & 85.33$_{\pm \text{1.50}}$ & 70.75$_{\pm \text{5.18}}$ & \cellcolor{yellow!10}\textbf{90.53$_{\pm \text{0.49}}$} & 70.47$_{\pm \text{1.12}}$ & 82.57$_{\pm \text{1.67}}$ & 63.86$_{\pm \text{2.76}}$ & 68.62$_{\pm \text{0.53}}$ & 84.23$_{\pm \text{0.51}}$ & 83.55$_{\pm \text{1.35}}$ & 87.21$_{\pm \text{1.88}}$  & 78.71\\ 
    & LLaMA-8B & 85.77$_{\pm \text{1.34}}$ & 74.84$_{\pm \text{1.09}}$ & 89.57$_{\pm \text{0.24}}$ & 72.72$_{\pm \text{0.26}}$  & 82.25$_{\pm \text{1.65}}$ & 61.12$_{\pm \text{0.45}}$ & 67.70$_{\pm \text{0.44}}$ & 84.05$_{\pm \text{0.26}}$ & 85.57$_{\pm \text{0.41}}$ & 89.42$_{\pm \text{0.12}}$ & 79.30 \\ 
      \bottomrule
    \end{tabular}
    }
    \label{tab:llaga_llm_s}
\end{table*}

\begin{table*}[!t]
    \centering
    \caption{\textbf{Macro-F1($\%$) of LLaGA to different LLM backbones under supervised settings}.}
    \vspace*{-8pt}
    \resizebox{\linewidth}{!}{
    \begin{tabular}{cc|ccccccccccc}
      \toprule
     \rowcolor{COLOR_MEAN}  & \textbf{LLM} & \textbf{Cora} & \textbf{Citeseer} & \textbf{Pubmed} & \textbf{arXiv} & \textbf{WikiCS} & \textbf{Instagram} & \textbf{Reddit} & \textbf{Books} & \textbf{Photo} & \textbf{Computer} & \textbf{Avg.} \\ \midrule
      \multirow{4}{*}{\rotatebox[origin=c]{90}{\begin{tabular}{c}
           \textbf{Same-} \\ \textbf{series}
      \end{tabular}}}  & Qwen-3B & 77.92$_{\pm \text{6.14}}$ & 66.52$_{\pm \text{5.69}}$ & 78.88$_{\pm \text{10.43}}$ & 51.30$_{\pm \text{0.83}}$ & 78.81$_{\pm \text{7.68}}$ & 50.93$_{\pm \text{7.72}}$ & 65.77$_{\pm \text{1.38}}$ & \cellcolor{red!10} \textbf{49.87$_{\pm \text{1.52}}$} & 77.51$_{\pm \text{3.24}}$ & \cellcolor{red!10} \textbf{80.77$_{\pm \text{3.27}}$} & 67.83 \\
    & Qwen-7B & 82.50$_{\pm \text{4.12}}$ & 64.03$_{\pm \text{4.86}}$ & \cellcolor{red!10}\textbf{90.29$_{\pm \text{0.52}}$} & 51.97$_{\pm \text{0.83}}$ & 77.35$_{\pm \text{4.26}}$ & 56.50$_{\pm \text{1.15}}$ & \cellcolor{red!10}\textbf{68.55$_{\pm \text{0.60}}$} & 46.21$_{\pm \text{1.78}}$ & 75.76$_{\pm \text{5.34}}$ & 78.86$_{\pm \text{5.90}}$ & 69.20 \\  
   & Qwen-14B &\cellcolor{red!10} \textbf{85.64$_{\pm \text{1.89}}$} & \cellcolor{red!10}\textbf{69.92$_{\pm \text{3.95}}$} & 89.69$_{\pm \text{0.39}}$ & \cellcolor{red!10}\textbf{53.32$_{\pm \text{0.38}}$} & \cellcolor{red!10}\textbf{79.13$_{\pm \text{1.78}}$} & \cellcolor{red!10}\textbf{57.58$_{\pm \text{0.83}}$} & 67.10$_{\pm \text{2.18}}$ & 44.26$_{\pm \text{2.27}}$ & 76.09$_{\pm \text{2.71}}$ & 80.17$_{\pm \text{4.74}}$ & \cellcolor{red!10} \textbf{70.29} \\   
    & Qwen-32B & 82.85$_{\pm \text{4.10}}$ & 68.11$_{\pm \text{3.69}}$ & 89.57$_{\pm \text{0.37}}$ & 52.52$_{\pm \text{0.65}}$ & 77.31$_{\pm \text{3.89}}$ & 57.28$_{\pm \text{2.21}}$ & 68.22$_{\pm \text{0.02}}$ & 48.25$_{\pm \text{1.29}}$ & \cellcolor{red!10}\textbf{79.51$_{\pm \text{0.87}}$} & 77.15$_{\pm \text{7.53}}$ & 70.08 \\ 
    \midrule
    
    \multirow{3}{*}{\rotatebox[origin=c]{90}{\small \begin{tabular}{c}
           \textbf{Similar} \\ \textbf{scales}
      \end{tabular}}}  & Mistral-7B & \cellcolor{yellow!10}\textbf{84.97$_{\pm \text{3.97}}$} & \cellcolor{yellow!10}\textbf{72.59$_{\pm \text{1.70}}$} & 90.00$_{\pm \text{0.80}}$ & \cellcolor{yellow!10}\textbf{58.08$_{\pm \text{0.29}}$} & \cellcolor{yellow!10}\textbf{82.37$_{\pm \text{1.73}}$} & \cellcolor{yellow!10} \textbf{57.96$_{\pm \text{2.40}}$} & 62.14$_{\pm \text{15.59}}$ & \cellcolor{yellow!10} \textbf{54.89$_{\pm \text{2.29}}$} & \cellcolor{yellow!10} \textbf{83.56$_{\pm \text{0.40}}$} & \cellcolor{yellow!10} \textbf{86.97$_{\pm \text{0.34}}$} &  \cellcolor{yellow!10}\textbf{73.35} \\ 
     & Qwen-7B & 82.50$_{\pm \text{4.12}}$ & 64.03$_{\pm \text{4.86}}$ & \cellcolor{yellow!10} \textbf{90.29$_{\pm \text{0.52}}$} & 45.74$_{\pm \text{9.78}}$ & 77.35$_{\pm \text{4.26}}$ & 56.50$_{\pm \text{1.15}}$ & \cellcolor{yellow!10} \textbf{68.55$_{\pm \text{0.60}}$} & 46.21$_{\pm \text{1.78}}$ & 75.76$_{\pm \text{5.34}}$ & 78.86$_{\pm \text{5.90}}$ & 69.09 \\  
    & LLaMA-8B & 81.40$_{\pm \text{5.46}}$ & 69.87$_{\pm \text{3.68}}$ & 89.30$_{\pm \text{0.23}}$ & 55.23$_{\pm \text{0.59}}$ & 80.14$_{\pm \text{2.09}}$ & 54.58$_{\pm \text{1.24}}$ & 67.40$_{\pm \text{0.61}}$ & 51.65$_{\pm \text{0.17}}$ & 78.87$_{\pm \text{2.38}}$ & 85.54$_{\pm \text{0.59}}$ & 71.40 \\
      \bottomrule
    \end{tabular}
    }
    \label{tab:llaga_llm_s_f1}
\end{table*}

\textbf{Results and Analysis: } As presented in Table \ref{tab:tape_llm_semi} (semi-supervised settings) and Table \ref{tab:tape_llm_fully} (supervised settings), the effectiveness of LLM-as-Explainer methods positively correlates with the strength of the underlying LLMs. In semi-supervised settings, TAPE utilizing GPT-4o consistently outperforms its Mistral-7B counterpart, achieving performance gains of up to 4\% on the Pubmed dataset. However, in supervised scenarios, the performance gap between GPT-4o and Mistral-7B narrows. This reduction is attributed to the abundance of labeled data, which increases the mutual information $I(\mathcal{E}; \mathcal{Y}_l)$ in \eqref{eq:mutual_info}. Consequently, the dependency on node attributes decreases, thereby diminishing the relative advantages of more powerful LLMs. 

Based on these findings, we recommend that when abundant supervision is available, practitioners may opt for open-source LLMs instead of more powerful and costlier models for the LLM-as-Explainer method. This practice can achieve comparable performance without incurring additional costs.

\subsection{LLM-as-Predictor: Sensitivity to LLM Backbones}

\textbf{Motivation and Settings: } For most LLM-as-Predictor methods, only open-source LLMs are compatible. Given the diverse choices and varying scales of these models, we aim to investigate the sensitivity of performance to different LLM backbones. This examination seeks to identify potential scaling laws and determine which LLMs excel at the node classification task. Therefore, we choose the best predictor method LLaGA as the baseline, include models of different sizes within the same series, i.e., Qwen2.5-series \cite{Yang2024Qwen2TR}. Additionally, we consider similar-scaled models to identify the most suitable for this task, including Qwen2.5-7B, Mistral-7B, and LLaMA3.1-8B. All experiments maintain consistency by only varying the backbone LLMs while keeping other components, training configurations, and hyperparameters unchanged.

\textbf{Results and Analysis: } \textbf{(1) Scaling within the same series: }Comparing Qwen-3B to Qwen-32B (performance shown in Tables \ref{tab:llaga_llm_s} and \ref{tab:llaga_llm_s_f1}, efficiency in Table \ref{tab:qwen_cost}, and performance trends in Figures \ref{fig:llaga_scaling} and \ref{fig:llaga_scaling_s} in the Appendix), we observe that performance generally improves with larger model sizes. However, beyond Qwen-7B and Qwen-14B, the performance gains become marginal while training and inference times increase significantly. For instance, Qwen-32B takes over 200 milliseconds per sample for inference, which is five times longer than Qwen-7B. Therefore, Qwen-7B or Qwen-14B are recommended as practical choices balancing performance and efficiency. \textbf{(2) Model selection at similar scales: }When comparing models of similar sizes (Tables \ref{tab:llaga_llm_s} and \ref{tab:llaga_llm_s_f1}, and Table \ref{tab:llaga_llm} in the Appendix), Mistral-7B outperforms other LLMs of comparable scale. Its superior performance makes Mistral-7B the recommended backbone LLM for node classification tasks.

\section{Conclusion}

This paper provides guidelines for leveraging LLMs to enhance node classification tasks across diverse real-world applications. We introduce LLMNodeBed, a codebase and testbed for systematic comparisons, featuring 14 datasets, 8 LLM-based algorithms, 8 classic algorithms, and 3 learning paradigms. Through extensive experiments involving 2,700 models, we uncover key insights: In supervised settings, each category offers unique advantages, but LLM-based approaches deliver marginal improvements over classic methods when ample supervision is available. In zero-shot scenarios, directing powerful LLMs to perform inference with integrated structural context yields the best performance.

Our findings offer practical guidance for practitioners applying LLMs to node classification tasks and highlight research gaps, e.g., the limited exploration of LLMs on heterophilic graphs and the scarcity of such text-rich datasets. We hope that LLMNodeBed will inspire and serve as a valuable toolkit for further research.

\clearpage 
\newpage

\section*{Acknowledgments}
 This research is supported in part by project \#MMT-p2-23 of the Shun Hing Institute of Advanced Engineering, The Chinese University of Hong Kong, by grants from the Research Grants Council of the Hong Kong SAR, China (No. CUHK 14217622). 
 The authors would like to express their gratitude to the reviewers for their feedback, which has improved the clarity and contribution of the paper.

\section*{Impact Statement}
In this paper, we did not use any non-public data, unauthorized software, or APIs, and there are no privacy or other related ethical concerns associated with our work. Similar to other machine learning models designed for node classification, our LLM-based algorithms have the potential to be misused for malicious purposes, such as unauthorized surveillance, manipulation of social networks, or exploitation of financial systems. We pledge to restrict the usage of our models exclusively to research settings to prevent such misuse and encourage the responsible development and deployment of these technologies in real-world applications.

\bibliography{reference}
\bibliographystyle{icml2025}

\newpage
\appendix
\onecolumn

\section{Related Works and Discussion}\label{sec:related_works}

In this section, we present a comprehensive taxonomy of node classification methods, ranging from classic approaches to those leveraging LLMs.

\subsection{Classic Methods}

Early approaches for node classification tasks relied on structural techniques such as Laplacian regularization \cite{belkin2006manifold}, graph embeddings \cite{yang2016revisiting}, and label propagation \cite{zhu2003semi}. These methods infer node labels by leveraging the connectivity and similarity among nodes within the graph. 

Over the past decade, GNNs have emerged as the dominant paradigm for node classification, demonstrating superior performance across various benchmarks \cite{kipf2017GCN, velickovic2018GAT, hamilton2017SAGE, xu2018GIN}. GNNs enhance node representations by aggregating and transforming feature information from their local neighborhoods. Formally, given a graph's feature matrix $\bm{X}$ and structure $\mathcal{E}$, a GNN produces the predicted label matrix as $\bm{Y} = \text{GNN}_{\Theta}(\bm{X}, \mathcal{E}) \in \mathbb{R}^{|\mathcal{V}| \times C}$, where $C$ is the number of classes, and $\Theta$ represents the learned parameters. 

Beyond structural methods, node classification can also be approached using LMs by treating each node as a text entity. Fine-tuning LMs \cite{Liu2019roberta, reimers-2019-sentence-bert, Wang2022e5-large} allows these models to map textual information directly to node labels, leveraging their strong language understanding capabilities to predict labels. To harness the complementary strengths of GNNs and LMs, i.e., structural and textual information, hybrid LM+GNN architectures have been developed \cite{jin2023patton, zhao2022GLEM, Wen2023G2P2, li2023grenade}. These models integrate LM-encoded textual features with GNN-processed structural features, enhancing node classification performance by combining both modalities.

\subsection{LLM-based Methods}
\textbf{LLM as Encoder: }LLMs possess an extensive number of parameters, enabling them to generate highly expressive representations. These representations can replace shallow node embeddings, promising to enhance expressiveness and improve the performance of downstream task. A notable approach is ENGINE \cite{Zhu2024ENGINE}, which utilizes hidden embeddings from LLMs to construct node embeddings. ENGINE integrates these embeddings with GNNs to propagate and update representations. Specifically, it aggregates hidden embeddings from each LLM layer that processes a node's text and incorporates them into a cascaded GNN structure.

\textbf{LLM as Explainer: }A notable advantage of LLMs lies in their generative capabilities, enabling them to intelligently perform a wide range of downstream tasks, from creative generation to reasoning and planning \cite{luo2024rog, wu2024graph}. Motivated by this strength, \citet{he2023TAPE} introduced TAPE, a method that leverages LLM as Explainer for node classification. Specifically, TAPE prompts the LLM to generate predictions along with a chain-of-thought reasoning process that includes explanations, denoted as $s_v^{\text{exp}} = \text{LLM}(s_v^{\text{orig}}, p)$, where $p$ denotes the textual prompt, and $s_v^{\text{orig}}$ and $s_v^{\text{exp}}$ represent the original and generated texts for node $v$, respectively. Both the original and generated texts are processed by an LM to produce embeddings as $\bm{x}_v^{\text{orig}} = \text{LM}(s_v^{\text{orig}})$ and $\bm{x}_v^{\text{exp}} = \text{LM}(s_v^{\text{exp}})$, which are subsequently processed by GNNs for the classification task as: 

\begin{equation*}
    \bm{Y} = \text{Ensemble}( \text{GNN}_{\Theta_1}(\bm{X}^{\text{orig}}, \mathcal{E}),  \text{GNN}_{\Theta_2}(\bm{X}^{\text{exp}}, \mathcal{E})).
\end{equation*}

Subsequent works have enhanced the reliability of LLM-generated explanations \cite{fang2024gaugl}. For example, KEA \cite{chen2024exploring} prompts LLMs to extract and explain specific technical terms from a node's original text instead of making direct predictions, thereby mitigating potential misguidance. Overall, the LLM-as-Explainer paradigm harnesses the reasoning and generative capabilities of LLMs to produce reliable explanations, thereby enriching the original graph data and enhancing downstream tasks.

\textbf{LLM as Predictor: }LLMs' strong reasoning abilities make them effective for direct downstream classification tasks. In the LLM-as-Predictor paradigm, a node's textual and structural information, along with task-specific instructions, are tokenized and input into an LLM for prediction. A notable method in this category is LLaGA \cite{chen23llaga}. Firstly, the original text $s_v$ of node $v$ is encoded via LM as $\bm{x}_v^{\text{LM}} = \text{LM}(s_v)$. Then, a parameter-free GNN, i.e., SGC \cite{Wu2019SimplifyingGC}, updates the node embeddings based on the graph structure, initializing with $\bm{h}_v^{(0)} = \bm{x}_v^{\text{LM}}$. The embeddings from each SGC layer are concatenated into $\bm{H}_v = [ \bm{h}_v^{(0)}, \ldots, \bm{h}_v^{(L)} ] $ and further projected into the LLM's dimensionality using a projection layer $\phi_{\theta}$. These projected embeddings are then combined with tokenized instructions $\bm{T}$ and input into the LLM to generate the predicted label as:

\begin{equation*}
    \ell_v = \text{LLM}( [  \phi_{\theta}(\bm{H}_v ) \; \| \;  \bm{T}  ] ).
\end{equation*}

In the LLaGA framework, only the parameters of the projection layer are tuned, utilizing the next-token-prediction loss based on ground-truth labels and generated outputs. GraphGPT employs a more complex framework with three distinct pre-training and instruction tuning stages. Other LLM-as-Predictor methods \cite{chai2023graphllm, perozzi2024graphtoken, Kong2024GOFAAG, Huang2024GraphAdapter, Zhao2023GraphTextGR, Ji2024NTLLMAN} share similar frameworks with LLaGA but vary in integration approaches, training objectives, and tackled tasks.

\subsection{Zero-shot Learning with LLMs}
Supervised learning approaches, which rely on labeled data, often struggle to keep pace with the rapid evolution of real-world graph data. Zero-shot learning methods address this limitation by enabling models to generalize to unseen data without requiring explicit labels. These methods can be broadly categorized into two approaches: LLM Direct Inference and GFMs.

\textbf{LLM Direct Inference} involves using LLMs to make predictions directly on the node's information through various prompt engineering techniques. Advanced prompt templates for reasoning tasks include Chain-of-Thought \cite{Wei2022ChainOT}, ReAct \cite{Yao2022ReActSR}, and Tree-of-Thought \cite{yao2023tree}. Besides, structural information can also be integrated into extended prompts \cite{tang2023graphgpt, wang2023can, Huang2023CanLE}, enriching the input provided to LLMs and facilitating more accurate predictions.

On the other hand, \textbf{GFMs} are foundation models pre-trained on extensive graph corpora to achieve general graph intelligence. Approaches such as ZeroG \cite{li2024zerog} and OFA \cite{liu2023one} fine-tune LMs or GNNs on multiple graphs, enabling these models to generalize to unseen graph datasets without extensive retraining. There also exist other zero-shot learning methods utilizing LLMs. For example, \citet{chen2024label} leverage LLMs as annotators to generate pseudo-labels for GNN training, enabling classification tasks. These approaches are not considered in this work, as our focus is primarily on LLMs directly solving node classification tasks in zero-shot scenarios.

\subsection{Benchmarks of LLMs for Graphs}
In addition to developments in node classification algorithms, we discuss existing benchmarks that leverage LLMs for graph-related tasks. These benchmarks can be categorized based on the type of tasks they address.

The first category primarily utilizes LLMs for basic graph reasoning tasks, e.g., shortest path and connectivity. For instance, NLGraph \cite{wang2023can} is a pioneering benchmark that encompasses eight different graph reasoning tasks presented in natural language. LLM4DyG \cite{zhang2023LLM4DyG} further extends these reasoning tasks to dynamic graph settings. GraphArena \cite{tang2024GraphArena} deals with more complex graph computational problems, with the complexity of tasks ranging from polynomial to NP-Complete like the Traveling Salesman Problem.  ProGraph \cite{li2024prograph} evaluates the scalability of LLMs by handling large graphs with up to $10^6$ nodes, necessitating the use of Python APIs for graph analysis rather than relying solely on direct reasoning of LLMs. Additionally, \citet{dai2024llm4pattern} investigates whether LLMs can recognize graph patterns, e.g., triangles or squares, based on terminological or topological descriptions.

The second category focuses on the potential of LLMs for node classification tasks. While numerous surveys discuss the progress in this area \cite{li2023survey, jin@llmgraph}, benchmarks that systematically evaluate LLM-based node classification methods remain limited. In the preliminary work by \citet{chen2024exploring}, the exploration is confined to a narrow range of LLM-as-Encoder and LLM-as-Explainer approaches, primarily focusing on a limited set of language models. GLBench \cite{Li2024GLBench} emerges as the first comprehensive benchmark for LLM-based node classification, offering consistent data splits to evaluate representative methods in both semi-supervised and zero-shot settings. However, variations in backbone models and implemented codebases impede fair and rigorous comparisons. 

Our benchmark, LLMNodeBed, distinguishes itself from existing benchmarks like GLBench by \textbf{standardizing implementations of baselines}, \textbf{extending learning paradigms and datasets} to encompass more real-world contexts, and \textbf{incorporating influential factors} like model type and size, homophily, and prompt design. This comprehensive approach provides more practical guidelines for effectively leveraging LLMs to enhance node classification tasks.

\clearpage
\newpage

\section{Prompts}

\subsection{Prompt in LLM-as-Predictor Methods}\label{sec:predictor_prompt}
For GraphGPT \cite{tang2023graphgpt} and LLaGA \cite{chen23llaga}, we utilize the prompt templates provided in their original papers for several datasets, including Cora and arXiv. For datasets not originally addressed, such as Photo, we adapt their prompt designs to create similarly formatted prompts. For LLM Instruction Tuning, we carefully craft prompt templates tailored to each dataset to directly guide the LLMs in performing classification tasks. Below is a summary of these prompt templates using Cora as an example: \textcolor{purple}{\textbf{$\langle$labels$\rangle$}} denotes the dataset-specific label space (Table \ref{tab:dataset_detail}), \textcolor{brown}{\textbf{$\langle$graph$\rangle$}} represents the tokenized graph context, and \textcolor{blue}{\textbf{$\langle$raw\_text$\rangle$}} refers to the node's original raw text.

\begin{tcolorbox}[colback=gray!10, colframe=black, boxrule=1pt, arc=2pt, left=5pt, right=5pt]

\textbf{Illustration of Prompts Utilized by LLM-as-Predictor Methods on the Cora Dataset} \vspace*{5pt}
\small

\textbf{LLM Instruction Tuning:} { 
Given a node-centered graph with centric node description: \textcolor{blue}{\textbf{$\langle$raw\_text$\rangle$}}, each node represents a paper, we need to classify the center node into 7 classes: \textcolor{purple}{\textbf{$\langle$labels$\rangle$}}, please tell me which class the center node belongs to?
} \vspace*{3pt}

\textbf{GraphGPT:} {
Given a citation graph: \textcolor{brown}{\textbf{$\langle$graph$\rangle$}}, where the 0-th node is the target paper, with the following information: \textcolor{blue}{\textbf{$\langle$raw\_text$\rangle$}}. Question: Which of the following specific research does this paper belong to: \textcolor{purple}{\textbf{$\langle$labels$\rangle$}}. Directly give the full name of the most likely category of this paper.
} \vspace*{3pt}

\textbf{LLaGA:} {
  Given a node-centered graph: \textcolor{brown}{\textbf{$\langle$graph$\rangle$}}, each node represents a paper, we need to classify the center node into 7 classes: \textcolor{purple}{\textbf{$\langle$labels$\rangle$}}, please tell me which class the center node belongs to?
}  
\end{tcolorbox}

\subsection{Prompt in Zero-shot Scenarios}\label{sec:zeroshot_prompt}

For LLM Direct Inference, we consider prompt templates including Direct, Chain-of-Thought \cite{Wei2022ChainOT}, Tree-of-Thought \cite{yao2023tree}, and ReACT \cite{Yao2022ReActSR}. The latter three methods are effective in enhancing LLM reasoning abilities across different tasks. Therefore, we adopt these advanced prompts for node classification to assess their continued effectiveness. Illustrations of these prompts on the Cora dataset are shown below: 
\begin{tcolorbox}[colback=gray!10, colframe=black, boxrule=1pt, arc=2pt, left=5pt, right=5pt]
\textbf{Illustration of Advanced Prompts Utilized by LLM Inference on the Cora Dataset}
\vspace*{5pt}

\small

\textbf{Direct:}
{Given the information of the node: \textcolor{blue}{\textbf{$\langle$raw\_text$\rangle$}}. Question: Which of the following categories does this paper belong to? Here are the categories: \textcolor{purple}{\textbf{$\langle$labels$\rangle$}}. Reply only with one category that you think this paper might belong to. Only reply with the category name without any other words. 
} \vspace*{3pt}

 \textbf{Chain-of-Thought:} 
{
Given the information of the node: \textcolor{blue}{\textbf{$\langle$raw\_text$\rangle$}}. 
Question: Which of the following types does this paper belong to? 
Here are the 7 categories: \textcolor{purple}{\textbf{$\langle$labels$\rangle$}}. \textcolor{red}{\textbf{Let's think about it step by step.}} Analyze the content of the node and choose one appropriate category.
Output format: $\langle$reason: $\rangle$, $\langle$classification: $\rangle$
 } \vspace*{3pt}

\textbf{Tree-of-Thought: }{Given the information of the node: \textcolor{blue}{\textbf{$\langle$raw\_text$\rangle$}}. 
Imagine three different experts answering this question. 
All experts will write down 1 step of their thinking, and then share it with the group. 
Then all experts will go on to the next step, etc. If any expert realizes they're wrong at any point then they leave.
Question: Based on this information, which of the following categories does this paper belong to? 
Here are 7 categories: \textcolor{purple}{\textbf{$\langle$labels$\rangle$}}. \textcolor{red}{\textbf{Let's think through this using a tree of thought approach. }}
Output format: $\langle$discussion: $\rangle$, $\langle$classification: $\rangle$. The classification should only consist of one of the category names listed.
} \vspace*{3pt}

 \textbf{ReACT: }{
Given the information of the node: \textcolor{blue}{\textbf{$\langle$raw\_text$\rangle$}}. Your task is to determine which of the following categories this paper belongs to. 
Here are the 7 categories: \textcolor{purple}{\textbf{$\langle$labels$\rangle$}}.
\textcolor{red}{\textbf{Solve this question by interleaving the Thought, Action, and Observation steps.}} Thought can reason about the current situation, and Action can be one of the following:
(1) Search[entity], which searches the exact entity on Wikipedia and returns the first paragraph if it exists. If not, it will return some similar entities to search.
(2) Lookup[keyword], which returns the next sentence containing the keyword in the current passage.
(3) Finish[answer], which returns the answer and finishes the task.
The output format must be $\langle$process: $\rangle$, $\langle$classification: $\rangle$. The classification should only consist of one of the category names listed.
}
\end{tcolorbox}

Additionally, we incorporate the central node's structural information into extended prompts to evaluate performance. We propose two variants for integrating neighbor information: (1) ``w. Neighbor'': we concatenate the texts of all $1$-hop neighbors of the central node as enriched context, and (2) ``w. Summary'': we first provide all neighbors' information to LLMs to generate a summary that highlights the common points among the neighbors. Then, we feed both the generated summary and the node's text to LLMs to facilitate prediction. The prompt templates for these methods on the Cora dataset are illustrated below:

\begin{tcolorbox}[colback=gray!10, colframe=black, boxrule=1pt, arc=2pt, left=5pt, right=5pt]

\textbf{Illustration of Structure-enriched Prompts Utilized by LLM Inference on the Cora Dataset} \vspace*{5pt}
 \small
 
\textbf{w. Neighbor: }{Given the information of the node: \textcolor{blue}{\textbf{$\langle$raw\_text$\rangle$}}. Given the information of its neighbors \textcolor{blue}{\textbf{$\langle$raw\_text$\rangle$}}.
Here I give you the content of the node itself and the information of its 1-hop neighbors. The relation between the node and its neighbors is `citation'. Question: Based on this information, which of the following sub-categories of AI does this paper belong to? Here are the 7 categories: \textcolor{purple}{\textbf{$\langle$labels$\rangle$}}. Reply only one category that you think this paper might belong to. Only reply with the category name without any other words. } \vspace*{4pt}

\textbf{w. Summary (Step 1): }{
The following list records papers related to the current one, with the relationship being `citation': \textcolor{blue}{\textbf{$\langle$raw\_text$\rangle$}}. Please summarize the information above with a short paragraph, and find some common points that can reflect the category of the paper.
} \vspace*{2pt}

\textbf{w. Summary (Step 2): }{Given the information of the node: \textcolor{blue}{\textbf{$\langle$raw\_text$\rangle$}}, \textcolor{teal}{\textbf{$\langle$summary$\rangle$}}.
Here I give you the content of the node itself and the summary information of its 1-hop neighbors. The relation between the node and its neighbors is `citation'. Question: Based on this information, which of the following sub-categories of AI does this paper belong to? Here are the 7 categories: \textcolor{purple}{\textbf{$\langle$labels$\rangle$}}.
 Reply only one category that you think this paper might belong to. Only reply with the category name without any other words. } 

\end{tcolorbox}

\clearpage
\newpage

\section{Supplementary Materials for LLMNodeBed}

\subsection{Datasets}\label{sec:dataset}

\begin{table}[!t]
    \centering
    \caption{\textbf{Details of datasets: label space and training data percentages with supervision}.}
    \resizebox{0.95\linewidth}{!}{
    \begin{tabular}{c|cc} 
      \toprule
      \rowcolor{COLOR_MEAN} \textbf{Domain} &  \textbf{Dataset}  & \textbf{Label Space} \\ \midrule
       \multirow{9}{*}{\textbf{Academic}} & Cora  & \begin{tabular}{c}
          Rule\_Learning, Neural\_Networks, Case\_Based, Genetic\_Algorithms,\\ Theory, Reinforcement\_Learning, Probabilistic\_Methods
       \end{tabular} \\
       & Citeseer & \begin{tabular}{c}
            Agents, ML (Machine Learning), IR (Information Retrieval), DB (Databases), \\ HCI (Human-Computer Interaction), AI (Artificial Intelligence)
       \end{tabular} \\ 
       & Pubmed & Experimentally induced diabetes, Type 1 diabetes, Type 2 diabetes \\ 
      &  arXiv & \begin{tabular}{c}
           cs.NA, cs.MM, cs.LO, cs.CY, cs.CR, cs.DC, cs.HC, cs.CE, cs.NI, cs.CC,\\ cs.AI, cs.MA, cs.GL, cs.NE, cs.SC, cs.AR, cs.CV, cs.GR, cs.ET, cs.SY,\\ cs.CG, cs.OH, cs.PL, cs.SE, cs.LG, cs.SD, cs.SI, cs.RO, cs.IT, cs.PF,\\ cs.CL, cs.IR, cs.MS, cs.FL, cs.DS, cs.OS, cs.GT, cs.DB, cs.DL, cs.DM
      \end{tabular}\\ \midrule 

      \textbf{Web Link} & WikiCS & \begin{tabular}{c} 
           Computational Linguistics, Databases, Operating Systems, Computer Architecture,\\ Computer Security, Internet Protocols, Computer File Systems,\\ Distributed Computing Architecture, Web Technology, Programming Language Topics
      \end{tabular} \\ \midrule

      \multirow{2}{*}{\textbf{Social}} & Instagram & Normal User, Commercial User \\ 
      & Reddit & Normal User, Popular User \\ \midrule

      \multirow{6}{*}{\textbf{E-Commerce}} & Books & \begin{tabular}{c} 
          World, Americas, Asia, Military, Europe, Russia, Africa, \\ Ancient Civilizations,  Middle East, Historical Study \& Educational Resources,\\ Australia \& Oceania, Arctic \& Antarctica
      \end{tabular} \\ 
      & Photo & \begin{tabular}{c} \small
           Video Surveillance, Accessories, Binoculars \& Scopes, Video,\\ Lighting \& Studio, Bags \& Cases,  Tripods \& Monopods, Flashes, \\ Digital Cameras, Film Photography, Lenses, Underwater Photography
      \end{tabular} \\ 
      & Computer & \begin{tabular}{c} 
          Computer Accessories \& Peripherals, Tablet Accessories, Laptop Accessories, \\ Computers \& Tablets,  Computer Components, Data Storage, Networking Products,\\ Monitors, Servers, Tablet Replacement Parts \\ 
      \end{tabular} 
      
      \\ \midrule 

      \multicolumn{2}{c}{\textbf{Heterophilic}} & Student, Faculty, Staff, Course, Project \\

      \bottomrule 
        
    \end{tabular}
    }
    \vspace*{10pt}
    \resizebox{0.95\linewidth}{!}{
     \begin{tabular}{c|ccccccccccc} 
         \toprule
        \rowcolor{COLOR_MEAN} \textbf{Setting} &  \textbf{Cora} & \textbf{Citeseer} & \textbf{Pubmed} & \textbf{arXiv} & \textbf{WikiCS} & \textbf{Instagram} & \textbf{Reddit} & \textbf{Books} & \textbf{Photo} & \textbf{Computer} & \textbf{Heterophilic} \\ \midrule
         Semi-supervised & 5.17\% & 3.77\% & 0.30\% & - & 4.96\% & 10.00\% & 10.00\% & 10.00\% & 10.00\% & 10.00\% & 10.00\% \\
         Supervised & 60.0\% & 60.0\% & 60.0\% & 53.7\% & 60.0\% & 60.0\% & 60.0\% & 60.0\% & 60.0\% & 60.0\% & 60.0\% \\
        \bottomrule
     \end{tabular}
    }
    \label{tab:dataset_detail}
\end{table}

We selected 14 datasets from academic, web link, social, and e-commerce domains to create a diverse graph database. Within LLMNodeBed, each dataset is stored in \texttt{.pt} format using PyTorch, which includes shallow embeddings, raw text of nodes, edge indices, labels, and data splits for convenient loading. The processed data is publicly available at \href{https://huggingface.co/datasets/xxwu/LLMNodeBed}{\texttt{https://huggingface.co/datasets/xxwu/LLMNodeBed}}. A description of the datasets is provided below, with their statistics and additional details summarized in Table \ref{tab:dataset} and Table \ref{tab:dataset_detail}, respectively.

\begin{itemize} 
    \item \textbf{Academic Networks: }The \textbf{Cora} \cite{Sen2008CollectiveCora}, \textbf{Citeseer} \cite{Giles1998CiteSeerAA}, \textbf{Pubmed} \cite{yang2016revisiting}, and \textbf{ogbn-arXiv} (abbreviated as ``arXiv'') \cite{hu2020open} datasets consist of nodes representing papers, with edges indicating citation relationships. The associated text attributes include each paper's title and abstract, which we use the collected version as follows: Cora and Pubmed from \citet{he2023TAPE}, Citeseer from \citet{chen2024exploring}. Within the dataset, each node is labeled according to its category. For example, the arXiv dataset includes 40 CS sub-categories such as cs.AI (Artificial Intelligence) and cs.DB (Databases).

    \item  \textbf{Web Link Network: }In the \textbf{WikiCS} dataset \cite{Mernyei2020WikiCSAW}, each node represents a Wikipedia page, and edges indicate reference links between pages. The raw text for each node includes the page name and content, which was collected by \citet{liu2023one}. The classification goal is to categorize each entity into different Wikipedia categories. 
    
    \item \textbf{Social Networks: }The \textbf{Reddit} and \textbf{Instagram} datasets, originally released in \citet{Huang2024GraphAdapter}, feature nodes representing users, with edges denoting social connections like following relationships. For Reddit, each user's associated text consists of their historically published sub-reddits, while for Instagram, it includes the user’s profile page introduction. In Reddit, nodes are labeled to indicate whether the user is popular or normal, while in Instagram, labels specify whether a user is commercial or normal.  

    \item \textbf{E-Commerce Networks: }The \textbf{Ele-Photo} (abbreviated as ``Photo'') and \textbf{Ele-Computer} (abbreviated as ``Computer'') datasets are derived from the Amazon Electronics dataset \cite{Ni2019Amazon}, where each node represents an item in the Photo or Computer category. The \textbf{Books-History} (abbreviated as ``Books'') dataset comes from the Amazon Books dataset, where each node corresponds to a book in the history category. We utilize the processed datasets released in \citet{yan2023comprehensive}. In these e-commerce networks, edges indicate co-purchase or co-view relationships. The associated text for each item includes descriptions, e.g., book titles and summaries, or user reviews. The classification task involves categorizing these products into fine-grained sub-categories.

    \item \textbf{Heterophilic Datasets: }The \textbf{Cornell}, \textbf{Texas}, \textbf{Wisconsin}, and \textbf{Washington}  datasets are collected from \citet{wang2025modelgeneralization}, which are the only available heterophilic datasets that have text attributes. Since the original release does not provide \textbf{shallow embeddings}, we generate fixed 300-dimensional embeddings for each node using Node2Vec \cite{Grover2016node2vecSF}. For \textbf{dataset splits}, we assign semi-supervised and supervised settings with 1:1:8 and 6:2:2 splits for training, validation, and test sets, respectively.
\end{itemize}

\subsection{Implementation Details and Hyperparameters Setting}\label{sec:hyperparam}

\begin{itemize}
    \item For \textbf{GNNs} with arbitrary input embeddings, either from shallow embeddings or those generated by LMs or LLMs, we perform a grid-search on the hyperparameters as follows:
    
    \texttt{num\_layers} in $[2, 3, 4]$, \texttt{hidden\_dimension} in $[32, 64, 128, 256]$, and \texttt{dropout} in $[0.3, 0.5, 0.7]$.

    Additionally, we explore the design space by considering the inclusion or exclusion of \texttt{batch\_normalization} and \texttt{residual\_connection}. 
    
    For \textbf{shallow embeddings}, the Cora, Citeseer, Pubmed, WikiCS, and arXiv datasets provide initialized embeddings in their released versions \cite{hu2020open, Sen2008CollectiveCora}. For remaining datasets lacking shallow embeddings, we construct these embeddings using \textbf{Node2Vec} \cite{Grover2016node2vecSF} techniques, generating a fixed $300$-dimensional embedding for each node based on a walk length of $30$ and a total of $10$ walks. 

    For the heterophilic GNN, \textbf{H$_2$GCN} \cite{zhu2020beyond}, which we employed on  heterophilic graphs in the fine-grained analysis of LLM-as-Encoder, we perform a grid-search over the following hyperparameters: 

    \texttt{num\_layers} in $[1, 2, 3]$, \texttt{hidden\_dimension} in $[32, 64, 128, 256]$, and \texttt{learning\_rate} in $[0.001, 0.005, 0.01]$. 
    
    \item For \textbf{MLPs} with arbitrary input embeddings, we perform a grid-search on the hyperparameters as follows: 

    \texttt{num\_layers} in $[2, 3, 4]$, \texttt{hidden\_dimension} in $[128, 256, 512]$, and \texttt{dropout} in $[0.5, 0.6, 0.7]$.

    For both GNNs and MLPs across experimental datasets, the \texttt{learning\_rate} is consistently set to $1e-2$, following previous studies \cite{he2023TAPE, Li2024GLBench}. The total number of epochs is set to $500$ with a patience of $100$. 

    \item For \textbf{SenBERT-66M} and \textbf{RoBERTa-355M}, we set the training epochs to $10$ for semi-supervised settings and $4$ for supervised settings. The \texttt{batch\_size} is set to $32$, and the \texttt{learning\_rate} is set to $2e-5$.  

    \item For \textbf{GLEM} \cite{zhao2022GLEM}, we uniformly set the number of EM iterations to $1$ and the pseudo-labeling ratio to $0.5$. For the GNN module within GLEM, we set the \texttt{hidden\_dim} to $256$ and \texttt{num\_layers} to $3$. For the LM module, we use LoRA for optimization with a \texttt{batch\_size} of $32$. Except for the Pubmed, arXiv, and Computer datasets, where the LM is trained first during the EM iteration, all remaining datasets train the GNN first. This choice is based on empirical findings that suggest better performance.

    \item For \textbf{ENGINE} \cite{Zhu2024ENGINE}, we refer to the hyperparameter settings outlined in the original paper to determine the hyperparameter search space as follows:
    
    \texttt{num\_layers} in $[1, 2, 3]$, $\texttt{hidden\_dimension}$ in $[64, 128]$, and \texttt{learning\_rate} in $[5e-4, 1e-3]$. 
    
    The neighborhood sampler is set to ``Random Walk'' for Cora while ``$k$-Hop'' with $k=2$ for the remaining datasets.

    \item For \textbf{TAPE} \cite{he2023TAPE}, we utilize the provided prompt templates to guide Mistral-7B and GPT-4o in conducting reasoning. The LM, RoBERTa-355M, is fine-tuned based on its default parameter settings, while the GNN hyperparameters are explored with \texttt{num\_layers} in $[2,3,4]$, \texttt{hidden\_dimension} in $[128,256]$, and with or without \texttt{batch\_normalization}.

    \item For \textbf{LLM Instruction Tuning}, we use the LoRA \cite{Hu2021LoRALA} techniques to fine-tune LLMs. The \texttt{lora\_r} parameter (dimension for LoRA update matrices) is set to $8$ and the \texttt{lora\_alpha} (scaling factor) to $16$. The \texttt{dropout} ratio is set to $0.1$, the \texttt{batch\_size} to $16$, and the \texttt{learning\_rate} to $1e-5$. For each dataset, the input consists of the node's original text along with a carefully crafted task prompt designed to guide the LLMs in performing the classification task. The expected output is the corresponding label. For small-scale datasets such as Cora, Citeseer, and Instagram, the number of training epochs is $10$ in semi-supervised settings and $4$ in supervised settings. For the remaining datasets, the training epochs are $2$ and $1$ for semi-supervised and supervised settings, respectively. The maximum input and output lengths are determined based on the average token lengths of each dataset.

    \item For \textbf{LLaGA} \cite{chen23llaga}, we empirically find that the HO templates consistently outperform the ND templates. Therefore, we set the HO templates as the default configuration, with \texttt{num\_hop} set to $4$. We use the text encoder as RoBERTa-355M. The linear projection layer $\phi_{\theta}(\cdot)$ consists of a $2-$layer MLP with a \texttt{hidden\_dimension} of $2048$. The \texttt{batch\_size} is set to $64$ and \texttt{learning\_rate} to $1e-4$. The number of training epochs is set to $10$ for semi-supervised settings and $4$ for supervised settings. For Qwen2.5-series, we encounter over-fitting issues in the Photo, Computer, and Books datasets,  leading us to adjust the learning rate to $5e-5$ and reduce the number of epochs to $2$ under supervised settings.

    \item For \textbf{GraphGPT} \cite{tang2023graphgpt}, it includes three distinct stages: (1) text-graph grounding, (2) self-supervised instruction tuning, and (3) task-specific instruction tuning. Our empirical findings indicate that the inclusion of stage (1) does not consistently lead to performance improvements, thereby rendering this stage optional. For stage (2), we construct self-supervised training data for each dataset to perform dataset-specific graph matching tasks, adhering to the provided data format\footnote{\href{https://huggingface.co/datasets/Jiabin99/graph-matching}{https://huggingface.co/datasets/Jiabin99/graph-matching}}. In stage (3), we utilize the training data to create $\langle$instruction, ground-truth label$\rangle$ pairs following the original prompt design. The training parameters for stage (2) include $2$ epochs with a \texttt{learning\_rate} of $1e-4$ and a \texttt{batch\_size} of $16$. For stage (3), we train for $10$ epochs in semi-supervised settings and $6$ epochs in supervised settings, with a \texttt{batch\_size} of $32$. Additionally, we adjust the maximum input and output lengths for each stage based on the dataset's text statistics.

    \item For \textbf{LLM Direct Inference}, we adopt two distinct categories of prompt templates: (1) advanced prompts that enhance the reasoning capabilities of LLMs, and (2) prompts enriched with structural information. These templates are illustrated in Appendix \ref{sec:zeroshot_prompt} and strictly adhere to the zero-shot setting.

    \item For \textbf{ZeroG}, we adhere to its original parameter configurations by setting $k=2$, the number of SGC iterations to $10$, and the \texttt{learning\_rate} to $1e-4$. In experiments involving \textbf{GFMs}, the intra-domain training mode utilizes the following source-target pairs: arXiv $\rightarrow$ Cora, arXiv $\rightarrow$ WikiCS, Reddit $\rightarrow$ Instagram, and Computer $\rightarrow$ Photo. 
\end{itemize}

\subsection{Distinct Features}\label{sec:distinct_llmnodebed}
A fair comparison necessitates a benchmark that evaluates all methods using consistent dataloaders, learning paradigms, backbone architectures, and implementation codebases. Our LLMNodeBed carefully follows these guidelines to support systematic and comprehensive evaluation of LLM-based node classification algorithms. Unlike existing benchmarks \cite{Li2024GLBench}, which primarily rely on each algorithm's official implementation, LLMNodeBed distinguishes itself in the following ways:

\begin{itemize}
    \item \textbf{Systematical Implementation: }We consolidate common components (e.g., DataLoader, Evaluation, Backbones) across algorithms to avoid code redundancy and enable fair comparisons and streamlined deployment. For example, several official implementations involve extensive code snippets, and we have produced cleaner, more streamlined versions that enhance both readability and usability. This systematic approach makes LLMNodeBed easily extendable to new datasets or algorithms.

    \item  \textbf{Flexible Selection of Backbones: } LLMNodeBed incorporates a diverse selection of GNNs, LMs, and LLMs, which can be seamlessly integrated as components in baseline methods. 
    \begin{itemize}
        \item \textbf{GNNs: }Our framework supports a wide range of variants, including GCN \cite{kipf2017GCN}, GraphSAGE \cite{hamilton2017SAGE}, GAT \cite{velickovic2018GAT}, GIN \cite{xu2018GIN},  Graph Transformers \cite{Shi2020GraphTransformer} and H$_2$GCN \cite{zhu2020beyond}. These GNNs can be customized with various layers and embedding dimensions.
        \item \textbf{LMs and LLMs: }Open-source models can be easily loaded via the Transformers library\footnote{\href{https://huggingface.co/docs/transformers}{https://huggingface.co/docs/transformers}}. In our experiments, we primarily utilize SenBERT-66M \cite{reimers-2019-sentence-bert}, RoBERTa-355M \cite{Liu2019roberta}, Qwen2.5-Series \cite{Yang2024Qwen2TR}, Mistral-7B \cite{Jiang2023Mistral7B}, and LLaMA3.1-8B \cite{llama3modelcard}. For close-source LLMs, we have formatted the invocation functions of DeepSeek-V3 \cite{deepseekai2024deepseekv3} and GPT-4o \cite{Achiam2023GPT4TR}. Additionally, LLMNodeBed allows users to specify and invoke any LM or LLM of their choice, providing flexibility for diverse research needs.
    \end{itemize}

    \item \textbf{Robust Evaluation Protocols: }LLMNodeBed incorporates comprehensive hyperparameter tuning and design space exploration to fully leverage the potential of the algorithms. For instance, recent research \cite{luo2024classic} highlights that classic GNNs remain strong baselines for node classification tasks, especially when the design space is expanded through techniques like residual connections, jumping knowledge, and selectable batch normalization. LLMNodeBed supports these enhancements, enabling the full utilization of GNNs. Furthermore, we conduct multiple experimental runs to enhance reliability and account for variability, which was often overlooked in previous studies.

\end{itemize}

\clearpage
\newpage

\section{Supplementary Experiments on Large-scale Dataset}\label{sec:exp_on_product}

\begin{table}[!h]
    \centering
     \caption{\textbf{Performance comparison (in $\%$) on large-scale dataset ogbn-products \cite{hu2020open}.}}
    \resizebox{\linewidth}{!}{
     \begin{tabular}{c|ccccc|ccc|c} 
       \toprule
       \rowcolor{COLOR_MEAN} & \multicolumn{5}{c|}{\textbf{Classic}} & \multicolumn{3}{c|}{\textbf{Encoder}} & \textbf{Predictor} \\ 
       \rowcolor{COLOR_MEAN} \multirow{-2}{*}{\textbf{Metric}} &  GCN\tiny{ShallowEmb} & SAGE\tiny{ShallowEmb} & GAT\tiny{ShallowEmb} & SenBERT & RoBERTa & GCN\tiny{LLMEmb} & SAGE\tiny{LLMEmb} & GAT\tiny{LLMEmb} & LLaGA  \\ \midrule 
        Accuracy & 75.16 & 77.33 & 74.83 & 69.06 & 75.12 & 81.09 & \cellcolor{orange!25}  \textbf{83.34} & 82.20 & 79.07 \\ 
        Macro-F1 & 36.16 & 35.63 & 34.32 & 24.89 & 30.29 & 39.96 & \cellcolor{orange!25}  \textbf{41.14} & 40.49 & 36.36 \\
        
        \bottomrule
    \end{tabular}
    
    }
    \label{tab:exp_products}
\end{table}

We extend our evaluation to the large-scale graph, ogbn-products \cite{hu2020open}, which comprises 2,449,029 nodes and 123,718,152 edges. In this graph, each node represents a unique product sold on Amazon, and edges denote co-purchase relationships. The classification task involves predicting the category of a product across 47 classes. This dataset is significantly larger than any other in LLMNodeBed, posing unique challenges for evaluating LLM-based node classification algorithms.

\textbf{Experimental Setups:} To evaluate LLM-based methods under constrained GPU resources (e.g., a single H100-80GB GPU), we use the \textbf{official data splits}, which include 196,165 nodes (8.03\%) for training, 39,323 nodes (1.61\%) for validation, and a reduced test set of 221,309 nodes (9.04\%), yielding a \textbf{semi-supervised setting}. The original dataset provides 100-dimensional node embeddings, which we directly adopt as shallow embeddings for classic GNNs.

\textbf{Compared Methods:} Given resource constraints, we compare \textbf{classic methods} (GNNs and LMs), \textbf{LLM-as-Encoder} methods (using embeddings derived from Qwen2.5-3B), and the LLM-as-Predictor method, \textbf{LLaGA}. Some other methods, such as TAPE, are excluded due to the prohibitive cost of generating explanatory texts for the entire dataset via APIs, e.g., approximately 500 USD for invoking GPT-4o. 

\textbf{Results: } The results for both Accuracy and Macro-F1 (in $\%$) are summarized in Table \ref{tab:exp_products}. From the table, we observe that LLM-based methods outperform the best classic methods by substantial margins, e.g., 6\% on Accuracy and 5\% on Macro-F1. These results underscore the advantages of LLM-based methods on large-scale graphs.

\clearpage 
\newpage

\section{Supplementary Materials for Comparisons among Algorithm Categories} 
\begin{table*}[!h]
    \centering
    \caption{\textbf{Performance comparison under semi-supervised and supervised settings with Macro-F1 ($\%$) reported.} \\\small{The \colorbox{orange!25}{\textbf{best}} and \colorbox{orange!10}{second-best} results are highlighted. LLM\textsubscript{IT} on the arXiv dataset requires extensive training time, preventing repeated experiments.}}
   \resizebox{\linewidth}{!}{
    \begin{tabular}{cc|cccccccccc}
      \toprule
     \rowcolor{COLOR_MEAN} \multicolumn{2}{c}{\textbf{Semi-supervised}}   & \textbf{Cora} & \textbf{Citeseer} & \textbf{Pubmed} & \textbf{WikiCS} & \textbf{Instagram} & \textbf{Reddit} & \textbf{Books} & \textbf{Photo} & \textbf{Computer} & \textbf{Avg.} \\ \midrule
         \multirow{6}{*}{\textbf{Classic}} & {GCN\tiny{ShallowEmb}} & 80.76$_{\pm \text{0.30}}$ & 66.00$_{\pm \text{0.17}}$ & \cellcolor{orange!10} 79.00$_{\pm \text{0.30}}$ & 77.87$_{\pm \text{0.18}}$ & 52.44$_{\pm \text{1.02}}$ & 61.15$_{\pm \text{0.56}}$ & 22.18$_{\pm \text{1.12}}$ & 62.65$_{\pm \text{1.46}}$ & 61.33$_{\pm \text{2.55}}$ & 62.60 \\ 
          & {SAGE\tiny{ShallowEmb}} & 80.88$_{\pm \text{0.47}}$ & 65.06$_{\pm \text{0.32}}$ & 77.88$_{\pm \text{0.41}}$ & 77.02$_{\pm \text{0.59}}$ & 50.74$_{\pm \text{0.26}}$ & 56.39$_{\pm \text{0.44}}$ & 24.37$_{\pm \text{0.72}}$ & 74.17$_{\pm \text{0.32}}$ & 71.67$_{\pm \text{0.61}}$ & 64.24 \\ 
        & {GAT\tiny{ShallowEmb}} & 79.65$_{\pm \text{1.03}}$ & 64.82$_{\pm \text{1.20}}$ & 78.43$_{\pm \text{0.73}}$ & 77.57$_{\pm \text{1.01}}$ & 40.19$_{\pm \text{1.56}}$ & 60.37$_{\pm \text{1.22}}$ & 28.93$_{\pm \text{3.65}}$ & 75.89$_{\pm \text{0.60}}$ & 77.06$_{\pm \text{2.98}}$ & 64.77 \\ 
        & SenBERT-66M & 64.54$_{\pm \text{1.18}}$ & 56.65$_{\pm \text{1.40}}$ & 32.08$_{\pm \text{3.35}}$ & 74.97$_{\pm \text{0.96}}$ & 55.47$_{\pm \text{0.76}}$ & 55.75$_{\pm \text{0.58}}$ & 43.15$_{\pm \text{1.53}}$ & 66.08$_{\pm \text{0.76}}$ & 59.79$_{\pm \text{1.30}}$  & 56.50 \\
         & {RoBERTa-355M} & 70.41$_{\pm \text{0.83}}$ & 63.36$_{\pm \text{1.75}}$ & 40.82$_{\pm \text{2.05}}$ & 73.98$_{\pm \text{1.72}}$ & \cellcolor{orange!10} 57.43$_{\pm \text{0.42}}$ & 59.23$_{\pm \text{0.36}}$ & \cellcolor{orange!25} \textbf{51.34$_{\pm \text{0.95}}$} & 67.92$_{\pm \text{0.49}}$ & 63.38$_{\pm \text{2.17}}$ & 60.87 \\ 
         & GLEM & 79.82$_{\pm \text{0.95}}$ & 64.66$_{\pm \text{2.39}}$ & \cellcolor{orange!25} \textbf{81.92$_{\pm \text{0.97}}$} & 73.88$_{\pm \text{0.25}}$ & 52.09$_{\pm \text{5.81}}$ & 49.30$_{\pm \text{4.48}}$ & 43.33$_{\pm \text{3.57}}$ & 67.53$_{\pm \text{1.18}}$ & 66.01$_{\pm \text{1.33}}$ & 64.28 \\ 
         
         \midrule
          
        \multirow{2}{*}{\textbf{Encoder}}  
       & $\text{GCN}_{\text{LLMEmb}}$ & 81.19$_{\pm \text{0.59}}$ & \cellcolor{orange!25} \cellcolor{orange!25} \textbf{67.17$_{\pm \text{0.73}}$} & 78.39$_{\pm \text{0.36}}$ & 78.58$_{\pm \text{0.51}}$ & \cellcolor{orange!25} \textbf{58.97$_{\pm \text{0.85}}$} & \cellcolor{orange!10} 68.46$_{\pm \text{0.91}}$ & 39.64$_{\pm \text{0.85}}$ & \cellcolor{orange!10} 79.87$_{\pm \text{0.57}}$ & 77.36$_{\pm \text{0.70}}$ &  \cellcolor{orange!10} 
 69.95 \\ 
       & ENGINE & \cellcolor{orange!25} \textbf{82.52$_{\pm \text{0.45}}$} & \cellcolor{orange!10} \textbf{67.15$_{\pm \text{0.15}}$} & 77.53$_{\pm \text{0.34}}$ & \cellcolor{orange!10} 78.89$_{\pm \text{0.38}}$ & 57.25$_{\pm \text{2.50}}$ & \cellcolor{orange!25} \textbf{69.56$_{\pm \text{0.21}}$} & 34.04$_{\pm \text{1.10}}$ & 78.55$_{\pm \text{1.12}}$ & 75.86$_{\pm \text{0.60}}$ & 69.04 \\  \midrule
       
       \textbf{Explainer} & TAPE & \cellcolor{orange!10} 81.89$_{\pm \text{0.31}}$ & 66.80$_{\pm \text{0.33}}$ & 78.46$_{\pm \text{1.13}}$ & \cellcolor{orange!25} \textbf{80.03$_{\pm \text{0.23}}$} & 50.01$_{\pm \text{1.60}}$ & 61.23$_{\pm \text{0.69}}$ & \cellcolor{orange!10} 47.12$_{\pm \text{3.26}}$ & \cellcolor{orange!25} \textbf{82.31$_{\pm \text{0.19}}$} & \cellcolor{orange!25} \textbf{84.90$_{\pm \text{1.14}}$} & \cellcolor{orange!25} \textbf{70.31} \\  \midrule
       
      \multirow{3}{*}{\textbf{Predictor}} & $\text{LLM}_{\text{IT}}$ & 56.35$_{\pm \text{0.22}}$ & 47.34$_{\pm \text{0.68}}$ & 62.81$_{\pm \text{0.21}}$ & 65.75$_{\pm \text{0.17}}$ & 38.30$_{\pm \text{0.94}}$ & 44.41$_{\pm \text{8.86}}$ & 39.44$_{\pm \text{0.44}}$ & 60.71$_{\pm \text{0.09}}$ & 57.38$_{\pm \text{0.65}}$ & 52.50 \\
       & GraphGPT & 58.33$_{\pm \text{0.81}}$ & 54.21$_{\pm \text{1.11}}$ & 56.09$_{\pm \text{0.88}}$ & 62.04$_{\pm \text{0.62}}$ & 38.78$_{\pm \text{0.60}}$ & 38.88$_{\pm \text{0.28}}$ & 42.85$_{\pm \text{0.94}}$ & 65.77$_{\pm \text{1.34}}$ & 66.69$_{\pm \text{1.49}}$ & 53.74  \\ 
       & LLaGA  & 71.14$_{\pm \text{4.47}}$ & 52.53$_{\pm \text{3.59}}$ & 45.12$_{\pm \text{7.63}}$ & 70.48$_{\pm \text{2.94}}$ & 50.12$_{\pm \text{10.45}}$ & 54.67$_{\pm \text{11.24}}$ & 39.70$_{\pm \text{2.44}}$ & 79.32$_{\pm \text{2.42}}$ & \cellcolor{orange!10} 78.01$_{\pm \text{1.36}}$ & 60.12 \\ \bottomrule
    \end{tabular}
    }

    \vspace*{5pt}
    \resizebox{\linewidth}{!}{
    \begin{tabular}{cc|ccccccccccc}
      \toprule
      \rowcolor{COLOR_MEAN} \multicolumn{2}{c}{\textbf{Supervised}}   & \textbf{Cora} & \textbf{Citeseer} & \textbf{Pubmed} & \textbf{arXiv} & \textbf{WikiCS} & \textbf{Instagram} & \textbf{Reddit} & \textbf{Books} & \textbf{Photo} & \textbf{Computer} & \textbf{Avg.} \\ \midrule
      \multirow{6}{*}{\textbf{Classic}} &{GCN\tiny{ShallowEmb}} & 86.54$_{\pm \text{1.44}}$ & 71.52$_{\pm \text{1.71}}$ & 88.54$_{\pm \text{0.60}}$ & 50.28$_{\pm \text{0.84}}$ & 82.11$_{\pm \text{0.61}}$ & 54.91$_{\pm \text{0.48}}$ & 65.00$_{\pm \text{0.42}}$ & 34.39$_{\pm \text{1.26}}$ & 66.04$_{\pm \text{2.85}}$ & 64.60$_{\pm \text{4.99}}$ & 66.39 \\ 

      & {SAGE\tiny{ShallowEmb}} & 86.37$_{\pm \text{1.26}}$ & 71.87$_{\pm \text{1.34}}$ & 90.16$_{\pm \text{0.27}}$ & 49.73$_{\pm \text{0.49}}$ & 82.78$_{\pm \text{1.53}}$ & 51.37$_{\pm \text{1.67}}$ & 61.39$_{\pm \text{0.54}}$ & 38.29$_{\pm \text{2.54}}$ & 80.37$_{\pm \text{0.34}}$ & 82.93$_{\pm \text{0.49}}$ & 69.53 \\ 
      & {GAT\tiny{ShallowEmb}} & 85.64$_{\pm \text{0.87}}$ & 69.27$_{\pm \text{2.15}}$ & 87.70$_{\pm \text{0.48}}$ & 49.71$_{\pm \text{0.23}}$ & 82.14$_{\pm \text{1.04}}$ & 50.26$_{\pm \text{3.16}}$ & 64.11$_{\pm \text{1.06}}$ & 42.85$_{\pm \text{1.62}}$ & 80.82$_{\pm \text{0.89}}$ & 84.74$_{\pm \text{0.79}}$ & 69.72 \\ 
      & SenBERT-66M & 77.13$_{\pm \text{2.19}}$ & 71.25$_{\pm \text{1.03}}$ & \cellcolor{orange!10} 93.95$_{\pm \text{0.39}}$ & 52.48$_{\pm \text{0.12}}$ & 84.43$_{\pm \text{1.58}}$ & 56.12$_{\pm \text{0.66}}$ & 58.31$_{\pm \text{0.76}}$ & 52.96$_{\pm \text{1.78}}$ & 70.39$_{\pm \text{0.54}}$ & 65.08$_{\pm \text{0.37}}$ & 68.21 \\
      & {RoBERTa-355M} & 81.38$_{\pm \text{1.17}}$ & 72.31$_{\pm \text{1.45}}$ & \cellcolor{orange!25} \textbf{94.33$_{\pm \text{0.14}}$} & 57.25$_{\pm \text{0.53}}$ & \cellcolor{orange!25} \textbf{86.10$_{\pm \text{1.11}}$} & 59.10$_{\pm \text{1.38}}$ & 60.16$_{\pm \text{0.94}}$ & \cellcolor{orange!25} \textbf{57.24$_{\pm \text{1.27}}$} & 72.89$_{\pm \text{0.50}}$ & 70.64$_{\pm \text{0.58}}$  & 71.14 \\ 
      & GLEM & 85.89$_{\pm \text{0.72}}$ & 70.06$_{\pm \text{2.02}}$ & 93.43$_{\pm \text{0.35}}$ & \cellcolor{orange!10} 57.99$_{\pm \text{1.29}}$ & 77.60$_{\pm \text{0.86}}$ & 58.87$_{\pm \text{2.79}}$ & 45.05$_{\pm \text{7.72}}$ & 46.96$_{\pm \text{3.55}}$ & 69.60$_{\pm \text{1.77}}$ & 68.62$_{\pm \text{1.14}}$ & 67.41 \\ 
      
      \midrule
      \multirow{2}{*}{\textbf{Encoder}} & $\text{GCN}_{\text{LLMEmb}}$ & \cellcolor{orange!25} \textbf{87.23$_{\pm \text{1.34}}$} & 
      \cellcolor{orange!10} 72.71$_{\pm \text{0.86}}$ & 87.76$_{\pm \text{0.76}}$ & 55.22$_{\pm \text{0.46}}$ & 82.78$_{\pm \text{1.68}}$ & \cellcolor{orange!25} \textbf{60.38$_{\pm \text{0.53}}$} & \cellcolor{orange!10} 70.64$_{\pm \text{0.75}}$ & 48.18$_{\pm \text{2.29}}$ & 80.51$_{\pm \text{0.65}}$ & 85.10$_{\pm \text{1.00}}$ & 73.05 \\ 
      & ENGINE & 85.72$_{\pm \text{1.58}}$ & 71.22$_{\pm \text{2.17}}$ & 89.63$_{\pm \text{0.14}}$ & 56.32$_{\pm \text{0.60}}$ & 83.80$_{\pm \text{1.06}}$ & \cellcolor{orange!10} 60.02$_{\pm \text{1.26}}$ & \cellcolor{orange!25} \textbf{71.17$_{\pm \text{0.75}}$} & 48.24$_{\pm \text{2.64}}$ & 82.77$_{\pm \text{0.28}}$ & 84.15$_{\pm \text{0.84}}$ & 73.30 \\  \midrule
      \textbf{Explainer} & TAPE & \cellcolor{orange!10} 87.21$_{\pm \text{1.60}}$ & \cellcolor{orange!25} \textbf{73.33$_{\pm \text{1.57}}$} & 92.39$_{\pm \text{0.02}}$ &  57.79$_{\pm \text{0.49}}$ & \cellcolor{orange!10} 86.03$_{\pm \text{1.14}}$ & 58.31$_{\pm \text{1.15}}$ & 65.91$_{\pm \text{0.71}}$ & 54.07$_{\pm \text{2.01}}$ & \cellcolor{orange!10} 83.41$_{\pm \text{0.42}}$ & \cellcolor{orange!10} 86.78$_{\pm \text{0.53}}$  & \cellcolor{orange!25} \textbf{74.52} \\  \midrule
      \multirow{3}{*}{\textbf{Predictor}} & $\text{LLM}_{\text{IT}}$ & 66.93$_{\pm \text{4.54}}$ & 52.22$_{\pm \text{2.71}}$ & 93.45$_{\pm \text{0.25}}$ & 57.48 & 78.39$_{\pm \text{1.20}}$ & 42.15$_{\pm \text{4.54}}$ & 56.65$_{\pm \text{0.85}}$ & 49.86$_{\pm \text{0.71}}$ & 68.74$_{\pm \text{2.54}}$ & 62.78$_{\pm \text{2.83}}$ & 62.86 \\ 
      & GraphGPT & 74.08$_{\pm \text{4.36}}$ & 61.04$_{\pm \text{1.24}}$ & 80.98$_{\pm \text{11.22}}$ & 56.80$_{\pm \text{0.08}}$ & 73.92$_{\pm \text{0.61}}$ & 40.07$_{\pm \text{2.10}}$ & 39.97$_{\pm \text{1.77}}$ & 47.97$_{\pm \text{1.94}}$ & 74.22$_{\pm \text{0.43}}$ & 74.19$_{\pm \text{1.75}}$ & 62.32 \\ 
      & LLaGA & 84.97$_{\pm \text{3.97}}$ & 72.59$_{\pm \text{1.70}}$ & 90.00$_{\pm \text{0.80}}$ & \cellcolor{orange!25} \textbf{58.08$_{\pm \text{0.29}}$} & 82.37$_{\pm \text{1.73}}$ & 57.96$_{\pm \text{2.40}}$ & 62.14$_{\pm \text{15.59}}$ & \cellcolor{orange!10} 54.89$_{\pm \text{2.29}}$ & \cellcolor{orange!25} \textbf{83.56$_{\pm \text{0.40}}$} & \cellcolor{orange!25} \textbf{86.97$_{\pm \text{0.34}}$} & \cellcolor{orange!10} 73.35 \\ \bottomrule
    \end{tabular}
    }
    \label{tab:mainexp_f1}
\end{table*}

\begin{table*}[!h]
    \centering
    \caption{\textbf{Performance comparison under zero-shot setting with both Accuracy ($\%$) and Macro-F1 ($\%$) reported.} }
    \resizebox{0.95\linewidth}{!}{
    \begin{tabular}{cc|cc|cc|cc|cc|cc}
       \toprule
       \rowcolor{COLOR_MEAN} & &  \multicolumn{2}{c|}{\textbf{Cora} (82.52)} & \multicolumn{2}{c|}{\textbf{WikiCS} (68.67)} & \multicolumn{2}{c|}{\textbf{Instagram} (63.35)}  & \multicolumn{2}{c|}{\textbf{Photo} (78.50)} & \multicolumn{2}{c}{\textbf{Avg.}}  \\ 
       \rowcolor{COLOR_MEAN} \multirow{-2}{*}{\textbf{Type \& LLM}} &  \multirow{-2}{*}{\textbf{Method}} & Acc & Macro-F1 & Acc & Macro-F1 & Acc & Macro-F1 &  Acc & Macro-F1  & Acc & Macro-F1 \\  \midrule
       \multirow{6}{*}{\begin{tabular}{c}
            \textbf{LLM} \\ DeepSeek-V3
       \end{tabular}} & Direct  & 68.06 & 60.07 & 71.41 & 65.21 & 42.42 & 39.42 & 65.25 & 56.92 & 61.78 & 55.40 \\
       & CoT  &  68.08 & 60.32 & 71.83 & 60.73 & \textbf{43.47} & 28.22 & 63.97 & 57.61 & 61.84 & 51.72 \\ 
       & ToT & 66.61 & 59.10 & 57.35 & 54.68 & 36.95 & 23.94 & 59.25 & 54.39 & 55.04 & 48.03 \\ 
       & ReAct & 65.68 & 58.68 & 71.24 & 60.97 & 43.30 & 28.42 & 63.66 & 56.23 & 60.97 & 51.08 \\ 
       & w. Neighbor & 68.63 & \textbf{69.21} & 70.26 & 64.26 & 43.34 & \textbf{41.57} & 61.57 & 54.84 & 60.95 & 57.47 \\ 
       & w. Summary & \textbf{73.62} & 64.80 & \textbf{72.53} & \textbf{67.28} & 41.18 & 37.56 & \textbf{72.73} & \textbf{70.58} & \textbf{65.02} & \textbf{60.05} \\ \midrule

       \multirow{6}{*}{\begin{tabular}{c}
            \textbf{LLM} \\ Mistral-7B
       \end{tabular}} & Direct & 59.65 & 58.34  &  70.13 &	67.80 &	44.29 &	42.16 &	\textbf{57.54} & 55.50 & 57.90 & 55.95 \\
       & CoT  & 58.02 & 57.13 & 69.00 &	66.17 &	45.48 & 44.56 & 49.56 &	51.42 & 55.52 & 54.82 \\ 
       & ToT  & 58.78 &	57.20 & 	67.56 &	64.52 &	45.39 &	44.73 &	44.25 &	46.87 & 54.00 & 53.33 \\
       & ReAct & 60.32 & 60.89 & \textbf{71.02} &	67.31 	& 	\textbf{46.26} &	\textbf{46.09} &	52.47 	& 50.92 & 57.52 & 56.30\\ 
       & w. Neighbor & 67.69 & 66.62 & 68.32 & 65.58 & 37.05 & 28.23 &  53.39 & 56.06 & 56.61 & 54.12  \\ 
       & w. Summary & \textbf{68.12} & \textbf{67.45} & 70.52 & \textbf{67.87} & 	41.94 &	38.93 &	56.01 &	\textbf{56.22} & \textbf{59.15} & \textbf{57.62} \\
       \bottomrule
    \end{tabular}
    }
    \label{tab:zeroshot_supple}
\end{table*}

\clearpage
\newpage

\section{Supplementary Materials for Fine-grained Analysis}

\subsection{LLM-as-Encoder: Compared with LMs}
\begin{table*}[!h]
    \centering
    \caption{\textbf{Comparison of LLM- and LM-as-Encoder with Accuracy ($\%$) reported under supervised setting.} The \colorbox{blue!10}{\textbf{best encoder}} within each method on a dataset is highlighted.}
    \resizebox{\linewidth}{!}{

    \begin{tabular}{cc|cccccccccc}
      \toprule
     \rowcolor{COLOR_MEAN} {\textbf{Method}}  & {\textbf{Encoder}}  & \textbf{Computer} &  \textbf{Cora} & \textbf{Pubmed} & \textbf{Photo} & \textbf{Books} & \textbf{Citeseer} & \textbf{WikiCS} & \textbf{arXiv} & \textbf{Instagram} & \textbf{Reddit} \\ \midrule
      \multicolumn{2}{c}{Homophily Ratio (\%)} & 85.28 & 82.52 & 79.24 &  78.50 & 78.05 & 72.93 & 68.67 & 63.53 & 63.35 & 55.22 \\ \midrule

    \multirow{4}{*}{MLP} & SenBERT & 71.75\textsubscript{±0.20} & 75.72\textsubscript{±2.19} & 90.26\textsubscript{±0.37} & 74.59\textsubscript{±0.19} & 83.61\textsubscript{±0.43} & 69.84\textsubscript{±2.21} & 81.13\textsubscript{±1.01} & 68.60\textsubscript{±0.16} & 67.12\textsubscript{±1.01} & 58.40\textsubscript{±0.56} \\
    & RoBERTa & \cellcolor{blue!10} \textbf{72.36\textsubscript{±0.09}} & 80.92\textsubscript{±2.65} & 90.54\textsubscript{±0.37} & 75.50\textsubscript{±0.25} & \cellcolor{blue!10} \textbf{84.06\textsubscript{±0.37}} & \cellcolor{blue!10} \textbf{74.11\textsubscript{±1.33}} & 83.18\textsubscript{±0.96} & 73.65\textsubscript{±0.11} & 68.57\textsubscript{±0.84} & 61.06\textsubscript{±0.31} \\
    & Qwen-3B & 69.25\textsubscript{±0.34} & 81.29\textsubscript{±1.05} & 92.02\textsubscript{±0.38} & 74.35\textsubscript{±0.47} & 83.43\textsubscript{±0.56} & 72.88\textsubscript{±1.66} & 85.46\textsubscript{±0.84} & 74.62\textsubscript{±0.19} & 68.62\textsubscript{±0.54} & 61.39\textsubscript{±0.36}\\ 
    & Mistral-7B & 71.38\textsubscript{±0.17} & \cellcolor{blue!10} \textbf{81.62\textsubscript{±0.63}} & \cellcolor{blue!10} \textbf{92.73\textsubscript{±0.24}} & \cellcolor{blue!10} \textbf{75.83\textsubscript{±0.25}} & 83.96\textsubscript{±0.46} & 73.85\textsubscript{±1.75}  & \cellcolor{blue!10} \textbf{86.92\textsubscript{±0.90}} & \cellcolor{blue!10} \textbf{75.29\textsubscript{±0.16}} & \cellcolor{blue!10} \textbf{69.03\textsubscript{±0.35}}  &  \cellcolor{blue!10}\textbf{62.49\textsubscript{±0.20}} \\ \midrule
    
    \multirow{4}{*}{GCN} & SenBERT & 
    \cellcolor{blue!10}
    \textbf{90.53\textsubscript{±0.18}} & \cellcolor{blue!10} \textbf{88.33\textsubscript{±0.90}} & \cellcolor{blue!10} \textbf{89.26\textsubscript{±0.24}} & \cellcolor{blue!10} \textbf{86.79\textsubscript{±0.18}} & \cellcolor{blue!10} \textbf{84.60\textsubscript{±0.38}} & 75.31\textsubscript{±0.55} & 84.28\textsubscript{±0.26} & 73.29\textsubscript{±0.22} & 68.13\textsubscript{±0.18} & 69.04\textsubscript{±0.46}\\ 
     & RoBERTa & 90.16\textsubscript{±0.13} & 87.96\textsubscript{±1.98} & 89.00\textsubscript{±0.21} & 86.79\textsubscript{±0.48} & 84.42\textsubscript{±0.38} & \cellcolor{blue!10} \textbf{76.57\textsubscript{±0.99}} & 84.64\textsubscript{±0.27} & 74.13\textsubscript{±0.19} & \cellcolor{blue!10} \textbf{68.43\textsubscript{±0.51}} & 69.28\textsubscript{±0.50} \\ 
     & Qwen-3B & 88.06\textsubscript{±0.35} & 88.24\textsubscript{±1.79} & 88.42\textsubscript{±0.51} & 85.20\textsubscript{±0.38} & 84.34\textsubscript{±0.61} & 76.37\textsubscript{±0.97} & 84.05\textsubscript{±0.55} & 73.62\textsubscript{±0.33} & 68.32\textsubscript{±0.73} & \cellcolor{blue!10} \textbf{71.04\textsubscript{±0.42}} \\ 
    & Mistral-7B & 89.52\textsubscript{±0.31} & 88.15\textsubscript{±1.79} & 88.38\textsubscript{±0.68} & 86.07\textsubscript{±0.63} & 84.23\textsubscript{±0.20} & 76.45\textsubscript{±1.19} & \cellcolor{blue!10} \textbf{84.78\textsubscript{±0.86}} & \cellcolor{blue!10} \textbf{74.39\textsubscript{±0.31}} & 68.27\textsubscript{±0.45} & 70.65\textsubscript{±0.75} \\ \midrule
    
    \multirow{4}{*}{SAGE} & SenBERT & \cellcolor{blue!10} \textbf{90.86\textsubscript{±0.18}} & 87.36\textsubscript{±1.79} & \cellcolor{blue!10} \textbf{90.93\textsubscript{±0.13}} & 87.41\textsubscript{±0.33} & \cellcolor{blue!10} \textbf{85.13\textsubscript{±0.27}} & 74.73\textsubscript{±0.67} & 85.94\textsubscript{±0.52} & 73.43\textsubscript{±0.23} & 67.72\textsubscript{±0.43} & 64.13\textsubscript{±0.41} \\ 
     & RoBERTa & 90.70\textsubscript{±0.25} & 87.36\textsubscript{±1.69} & 90.38\textsubscript{±0.09} & \cellcolor{blue!10} \textbf{87.42\textsubscript{±0.51}} & 85.13\textsubscript{±0.41} & \cellcolor{blue!10} \textbf{75.90\textsubscript{±0.41}} & 86.31\textsubscript{±0.68} & 75.28\textsubscript{±0.31} & 68.84\textsubscript{±0.54} & \cellcolor{blue!10}  \textbf{64.85\textsubscript{±0.31}} \\
    & Qwen-3B & 87.44\textsubscript{±0.66} & \cellcolor{blue!10} \textbf{87.36\textsubscript{±1.10}} & 89.98\textsubscript{±0.38} & 85.17\textsubscript{±0.44}  & 84.69\textsubscript{±0.31} & 75.63\textsubscript{±0.94} & 85.58\textsubscript{±0.58} & 75.20\textsubscript{±0.49} & 68.43\textsubscript{±0.57} & 63.98\textsubscript{±0.69} \\ 
    & Mistral-7B & 90.16\textsubscript{±0.26} & 87.22\textsubscript{±1.24} & 90.54\textsubscript{±0.50} & 87.34\textsubscript{±0.43} & 85.01\textsubscript{±0.49} & 75.20\textsubscript{±1.34} & \cellcolor{blue!10} \textbf{87.87\textsubscript{±0.35}} & \cellcolor{blue!10} \textbf{76.18\textsubscript{±0.34}} & \cellcolor{blue!10} \textbf{69.39\textsubscript{±0.52}} & 64.34\textsubscript{±0.23} \\ 

    \bottomrule
    \end{tabular}
    }
    \label{tab:encoder_comp_fullysupervised}
\end{table*}

We supplement the comparison between LLM-as-Encoder and LM-as-Encoder under supervised settings in Table \ref{tab:encoder_comp_fullysupervised}. The key takeaway that \textbf{LLMs outperform LMs as Encoders in less informative graphs, e.g., heterophilic graphs} remains valid. This conclusion is particularly evident on the arXiv dataset, where the performance gap between LM- and LLM-generated embeddings reaches up to $7\%$ on MLP and $3\%$ on GraphSAGE. Additionally, we observe that in supervised settings, the performance gap between LM- and LLM-as-Encoders becomes less pronounced compared to semi-supervised settings. We still consider the theoretical insights in Equation \eqref{eq:mutual_info} for explanation: Increased supervision enhances the mutual information between labels and graph structure, i.e., $I(\mathcal{E}, \mathcal{Y}_{l})$, thereby rendering the second term less significant and diminishing the advantages provided by more powerful encoders like LLMs.

\subsection{LLM-as-Predictor: Sensitivity to LLM Backbones}

\begin{table*}[!h]
    \centering
    \caption{\textbf{Sensitivity of LLaGA to different LLM backbones under semi-supervised Settings}.\\ \small{The best LLM backbone within \colorbox{red!10}{\textbf{each series}} and \colorbox{yellow!20}{\textbf{at similar scales}} is highlighted.}}
    \resizebox{\linewidth}{!}{
    \begin{tabular}{ccc|ccccccccc}
      \toprule
     \rowcolor{COLOR_MEAN} & & \textbf{LLM} & \textbf{Cora} & \textbf{Citeseer} & \textbf{Pubmed} & \textbf{WikiCS} & \textbf{Instagram} & \textbf{Reddit} & \textbf{Books} & \textbf{Photo} & \textbf{Computer} \\ \midrule
      \multirow{8}{*}{\rotatebox[origin=c]{90}{\textbf{Accuracy (\%)}}} & \multirow{4}{*}{ \rotatebox[origin=c]{90}{\small \begin{tabular}{c}
         \textbf{Same}\\ \textbf{series}
      \end{tabular}}} & Qwen-3B  & 74.15$_{\pm \text{2.41}}$ & 62.74$_{\pm \text{10.42}}$ & 54.97$_{\pm \text{10.71}}$ & 71.64$_{\pm \text{1.34}}$ & \cellcolor{red!10}\textbf{61.35$_{\pm \text{2.35}}$} & 65.11$_{\pm \text{1.59}}$ & 82.26$_{\pm \text{0.43}}$ & \cellcolor{red!10} \textbf{83.85$_{\pm \text{0.77}}$} & \cellcolor{red!10} \textbf{85.84$_{\pm \text{1.12}}$} \\ 
     & & Qwen-7B  & 74.23$_{\pm \text{1.58}}$ & 64.79$_{\pm \text{1.77}}$ & \cellcolor{red!10} \textbf{62.58$_{\pm \text{1.36}}$} & 71.40$_{\pm \text{2.28}}$ & 59.09$_{\pm \text{3.32}}$ & 66.07$_{\pm \text{0.32}}$ & 81.38$_{\pm \text{1.58}}$ & 80.75$_{\pm \text{1.60}}$ & 84.47$_{\pm \text{1.73}}$ \\ 
     & & Qwen-14B & 76.63$_{\pm \text{1.79}}$ & \cellcolor{red!10} \textbf{66.06$_{\pm \text{1.86}}$} & 62.17$_{\pm \text{6.86}}$ & \cellcolor{red!10} \textbf{73.76$_{\pm \text{0.42}}$} & 61.14$_{\pm \text{3.84}}$ & \cellcolor{red!10} \textbf{66.59$_{\pm \text{1.23}}$} & 81.40$_{\pm \text{0.20}}$ & 82.47$_{\pm \text{0.80}}$ & 85.08$_{\pm \text{0.36}}$  \\ 
     & & Qwen-32B & \cellcolor{red!10} \textbf{77.01$_{\pm \text{3.62}}$} & 64.01$_{\pm \text{2.59}}$ & 58.60$_{\pm \text{7.93}}$ & 71.31$_{\pm \text{3.05}}$ & 60.24$_{\pm \text{4.03}}$ & 66.19$_{\pm \text{2.11}}$ & \cellcolor{red!10} \textbf{82.34$_{\pm \text{0.44}}$} & 82.85$_{\pm \text{1.52}}$ & 85.74$_{\pm \text{1.65}}$ \\  \cmidrule(r){2-12}
     &  \multirow{3}{*}{\rotatebox[origin=c]{90}{\small \begin{tabular}{c}
          \textbf{Similar}\\ \textbf{scales}
     \end{tabular}}} & Mistral-7B & \cellcolor{yellow!20}\textbf{78.94$_{\pm \text{1.14}}$} & 62.61$_{\pm \text{3.63}}$ & \cellcolor{yellow!20}\textbf{65.91$_{\pm \text{2.09}}$} & \cellcolor{yellow!20}\textbf{76.47$_{\pm \text{2.20}}$} & \cellcolor{yellow!20}\textbf{65.84$_{\pm \text{0.72}}$} & \cellcolor{yellow!20}\textbf{70.10$_{\pm \text{0.38}}$} & \cellcolor{yellow!20}\textbf{83.47$_{\pm \text{0.45}}$} & \cellcolor{yellow!20}\textbf{84.44$_{\pm \text{0.90}}$} & \cellcolor{yellow!20}\textbf{87.82$_{\pm \text{0.53}}$} \\ 
    & & Qwen-7B  & 74.23$_{\pm \text{1.58}}$ & \cellcolor{yellow!20}\textbf{64.79$_{\pm \text{1.77}}$} & 62.58$_{\pm \text{1.36}}$ & 71.40$_{\pm \text{2.28}}$ & 59.09$_{\pm \text{3.32}}$ & 66.07$_{\pm \text{0.32}}$ & 81.38$_{\pm \text{1.58}}$ & 80.75$_{\pm \text{1.60}}$ & 84.47$_{\pm \text{1.73}}$ \\ 
   & & LLaMA-8B  & 75.34$_{\pm \text{1.09}}$ & 61.33$_{\pm \text{2.11}}$ & 61.84$_{\pm \text{3.62}}$ & 72.15$_{\pm \text{3.32}}$ & 55.77$_{\pm \text{3.07}}$ & 65.09$_{\pm \text{1.39}}$ & 81.30$_{\pm \text{0.07}}$ & 82.26$_{\pm \text{1.67}}$ & 86.43$_{\pm \text{0.25}}$ \\ \midrule

    \multirow{8}{*}{\rotatebox[origin=c]{90}{\textbf{Macro-F1 (\%)}}} & \multirow{4}{*}{ \rotatebox[origin=c]{90}{\small \begin{tabular}{c}
         \textbf{Same}\\ \textbf{series}
      \end{tabular}}} & Qwen-3B & 64.48$_{\pm \text{4.32}}$ & 53.25$_{\pm \text{2.21}}$ & 44.35$_{\pm \text{3.64}}$ & 59.09$_{\pm \text{3.00}}$ & 51.92$_{\pm \text{8.02}}$ & 42.42$_{\pm \text{0.50}}$ & \cellcolor{red!10}\textbf{37.56$_{\pm \text{2.56}}$} & \cellcolor{red!10} \textbf{70.94$_{\pm \text{0.97}}$} & \cellcolor{red!10} \textbf{70.65$_{\pm \text{3.31}}$} \\ 
     & & Qwen-7B  & 67.30$_{\pm \text{5.70}}$ & 53.43$_{\pm \text{1.63}}$ & \cellcolor{red!10} \textbf{47.55$_{\pm \text{1.16}}$} & 58.97$_{\pm \text{3.67}}$ & 49.78$_{\pm \text{8.13}}$ & \cellcolor{red!10} \textbf{51.32$_{\pm \text{10.10}}$} & 34.33$_{\pm \text{4.37}}$ & 66.94$_{\pm \text{4.64}}$ & 69.23$_{\pm \text{3.26}}$ \\ 
     & & Qwen-14B  & 67.85$_{\pm \text{4.53}}$ & \cellcolor{red!10} \textbf{55.39$_{\pm \text{3.99}}$} & 45.90$_{\pm \text{7.69}}$ & \cellcolor{red!10} \textbf{64.10$_{\pm \text{0.24}}$} & \cellcolor{red!10} \textbf{55.40$_{\pm \text{1.26}}$} & 44.35$_{\pm \text{0.86}}$ &37.50$_{\pm \text{0.35}}$ & 70.33$_{\pm \text{1.12}}$ & 69.19$_{\pm \text{5.16}}$ \\ 
     & & Qwen-32B & \cellcolor{red!10} \textbf{68.27$_{\pm \text{6.24}}$} & 53.11$_{\pm \text{2.59}}$ & 46.52$_{\pm \text{7.54}}$ & 60.50$_{\pm \text{3.40}}$ & 39.50$_{\pm \text{8.07}}$ & 43.96$_{\pm \text{1.45}}$ & 35.30$_{\pm \text{1.80}}$ & 70.08$_{\pm \text{2.39}}$ & 67.26$_{\pm \text{3.89}}$ \\  \cmidrule(r){2-12}
     &  \multirow{3}{*}{\rotatebox[origin=c]{90}{\small \begin{tabular}{c}
          \textbf{Similar}\\ \textbf{scales}
     \end{tabular}}} & Mistral-7B & \cellcolor{yellow!20}\textbf{71.14$_{\pm \text{4.47}}$} & 52.53$_{\pm \text{3.59}}$ & 45.12$_{\pm \text{7.63}}$ & \cellcolor{yellow!20}\textbf{70.48$_{\pm \text{2.94}}$} & \cellcolor{yellow!20}\textbf{50.12$_{\pm \text{10.45}}$} & \cellcolor{yellow!20}\textbf{54.67$_{\pm \text{11.24}}$} & \cellcolor{yellow!20}\textbf{39.70$_{\pm \text{2.44}}$} & \cellcolor{yellow!20}\textbf{79.32$_{\pm \text{2.42}}$} & \cellcolor{yellow!20}\textbf{78.01$_{\pm \text{1.36}}$}  \\ 
    & & Qwen-7B  & 67.30$_{\pm \text{5.70}}$ & \cellcolor{yellow!20}\textbf{53.43$_{\pm \text{1.63}}$} & 47.55$_{\pm \text{1.16}}$ & 58.97$_{\pm \text{3.67}}$ & 49.78$_{\pm \text{8.13}}$ & 51.32$_{\pm \text{10.10}}$ & 34.33$_{\pm \text{4.37}}$ & 66.94$_{\pm \text{4.64}}$ & 69.23$_{\pm \text{3.26}}$  \\ 
   & & LLaMA-8B & 67.50$_{\pm \text{4.73}}$ & 51.22$_{\pm \text{1.27}}$ & \cellcolor{yellow!20}\textbf{47.80$_{\pm \text{2.78}}$} & 64.17$_{\pm \text{6.00}}$ & 48.56$_{\pm \text{6.92}}$ & 43.31$_{\pm \text{0.94}}$ & 34.49$_{\pm \text{1.48}}$ & 72.45$_{\pm \text{0.35}}$ & 71.43$_{\pm \text{4.43}}$ \\ 

      \bottomrule
    \end{tabular}
    }
    \label{tab:llaga_llm}
\end{table*}

\begin{figure}[!h]
    \centering
    \begin{subfigure}[b]{0.25\textwidth}
        \centering
        \includegraphics[width=\textwidth]{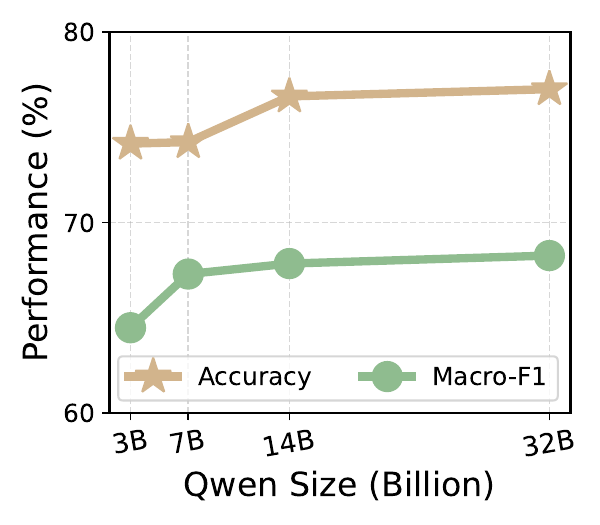}        
        \caption{Cora}
        \label{fig:qwen_cora}
    \end{subfigure}
    \begin{subfigure}[b]{0.25\textwidth}
        \centering
        \includegraphics[width=\textwidth]{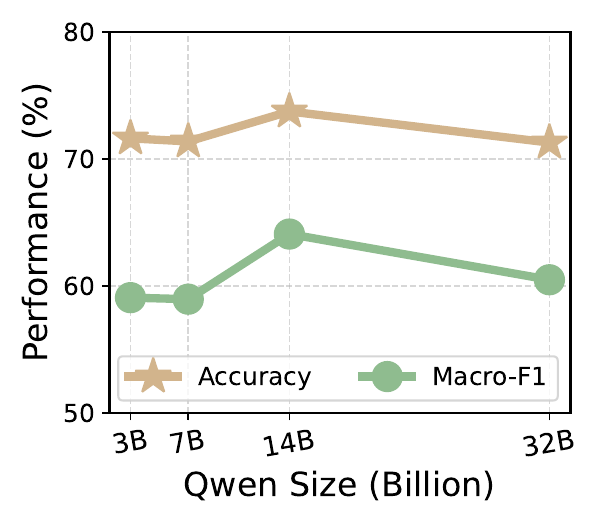}
        \caption{WikiCS}
        \label{fig:qwen_citeseer}
    \end{subfigure}
    \begin{subfigure}[b]{0.25\textwidth}
        \centering
        \includegraphics[width=\textwidth]{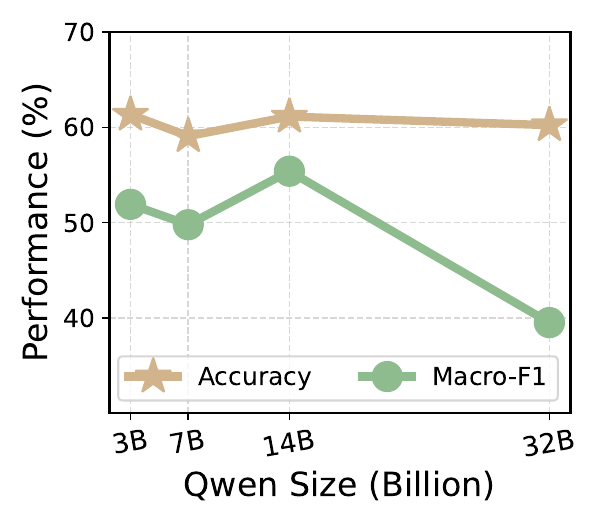}
        \caption{Instagram}
        \label{fig:qwen_instagram}
    \end{subfigure}
    \caption{\textbf{Performance trends within Qwen-series in different scales using LLaGA framework in semi-supervised settings.}}
    \label{fig:llaga_scaling}
\end{figure}

\begin{figure}[!h]
    \centering
    \begin{subfigure}[b]{0.25\textwidth}
        \centering
        \includegraphics[width=\textwidth]{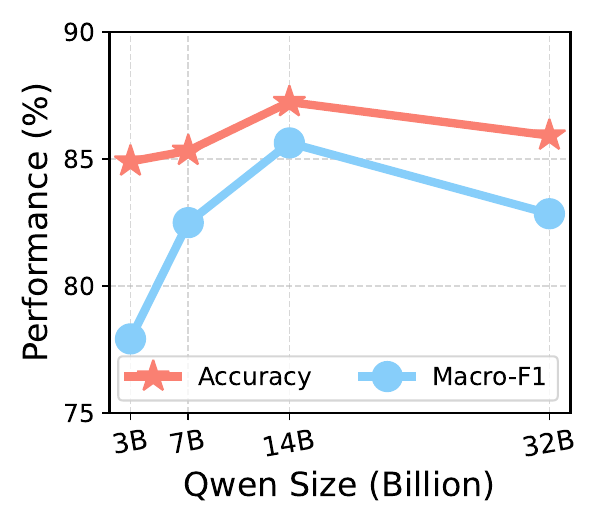}        
        \caption{Cora}
        \label{fig:qwen_cora_s}
    \end{subfigure}
    \begin{subfigure}[b]{0.25\textwidth}
        \centering
        \includegraphics[width=\textwidth]{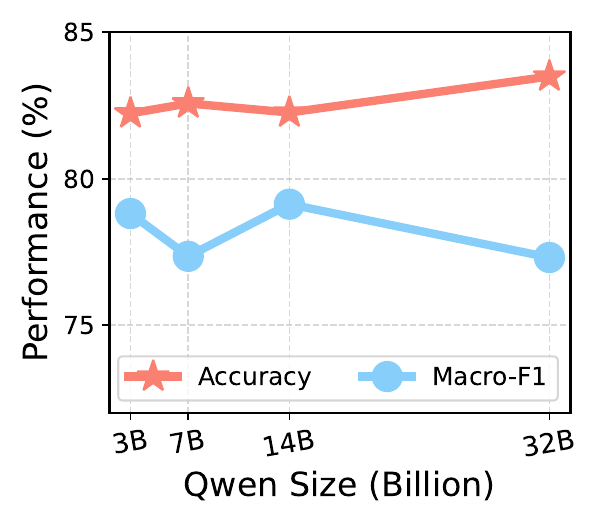}
        \caption{WikiCS}
        \label{fig:qwen_citeseer_s}
    \end{subfigure}
    \begin{subfigure}[b]{0.25\textwidth}
        \centering
        \includegraphics[width=\textwidth]{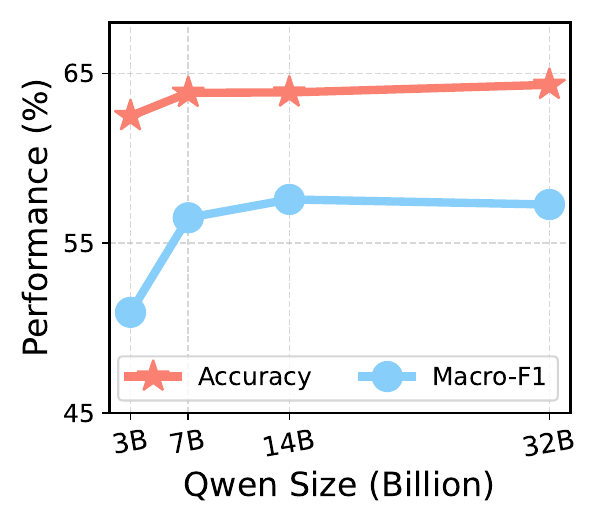}
        \caption{Instagram}
        \label{fig:qwen_instagram_s}
    \end{subfigure}
    \caption{\textbf{Performance trends within Qwen-series in different scales using LLaGA framework in supervised settings.}}
    \label{fig:llaga_scaling_s}
\end{figure}

\begin{table}[!h]
    \centering
    \caption{\textbf{Training and inference times of Qwen-series models ranging from 3B to 32B parameters.}}
    \resizebox{\textwidth}{!}{
    \begin{tabular}{cc|ccc|ccc|ccc}
       \toprule
        \rowcolor{COLOR_MEAN} &  & \multicolumn{3}{c|}{\textbf{Semi-supervised Training Times}} & \multicolumn{3}{c}{\textbf{Supervised Training Times}} & \multicolumn{3}{|c}{\textbf{Avg. Inference Times Per Case}} \\
       \rowcolor{COLOR_MEAN} \multirow{-2}{*}{\textbf{GPU Device}}  & \multirow{-2}{*}{\textbf{LLM}} & Cora & WikiCS & Instagram  &  Cora & WikiCS & Instagram &  Cora & WikiCS & Instagram \\  \midrule 
        &  Qwen-3B & 2.2min & 7.1min & 5.8min & 8.6min & 33.5min & 20.2min & 32.3ms & 37.3ms & 26.9ms \\ 
        \multirow{-2}{*}{1 NVIDIA A6000-48G} & Qwen-7B & 4.7min & 15.3min & 8.2min & 13.3min & 59.4min & 43.2min & 50.9ms & 55.9ms & 43.4ms \\ 
       & Qwen-14B & 8.9min & 25.5min & 15.7min & 30.8min & 2.3h & 1.5h & 97.6ms & 103.0ms & 83.2ms \\ 
        \multirow{-2}{*}{2 NVIDIA A6000-48G} & Qwen-32B & 18.9min & 43.6min & 30.7min & 52.2min & 3.3h & 2.0h & 254.7ms & 262.6ms & 232.4ms \\ 
       \bottomrule
    \end{tabular}
    }
    \label{tab:qwen_cost}
\end{table}

We evaluate LLaGA with different LLM backbones in semi-supervised settings, as detailed in Table \ref{tab:llaga_llm}. We examine two primary trends: (1) \textbf{Scaling within the same series}: Assessing whether scaling laws apply to node classification tasks by using LLMs from the same series, and (2) \textbf{Model selection at similar scales}: Identifying the most suitable LLM for node classification tasks by comparing models of similar scales.

\textbf{Scaling within the same series}: We plot performance trends across several datasets under both semi-supervised and supervised settings to clearly illustrate these dynamics. From Figure \ref{fig:llaga_scaling} and Figure \ref{fig:llaga_scaling_s}, we conclude that scaling laws generally hold: as the Qwen model size increases from 3B to 32B parameters, performance improves, indicating that larger model sizes enhance task performance. However, the 7B and 14B models are sufficiently large, typically representing the point beyond which further increases in model size yield only marginal improvements but introducing huge computational costs (Table \ref{tab:qwen_cost}). Unexpectedly, in the Instagram dataset under semi-supervised settings, the Qwen-32B model experiences a performance drop. This may be because 32B models require extensive data to train effectively, making them less robust and stable compared to smaller models. Based on these findings, we recommend the 7B or 14B models as they offer an optimal balance between performance and computational costs.

\textbf{Model selection at similar scales:} By comparing the performance of Mistral-7B, Qwen-7B, and LLaMA-8B in Table \ref{tab:llaga_llm}, we conclude that Mistral-7B outperforms the other two similarly scaled LLMs in most cases. This makes Mistral-7B the optimal choice as a backbone LLM for node classification tasks.

\clearpage
\newpage
\subsection{LLM-as-Predictor: Biased and Hallucinated Predictions}\label{sec:llm_bias_pred}

During our experiments, we found that LLM-as-Predictor methods are vulnerable to limited supervision. In addition to standard metrics such as Accuracy (Table \ref{tab:mainexp}) and Macro-F1 scores (Table \ref{tab:mainexp_f1}), their predictions also exhibit significant  \textbf{biases} and \textbf{hallucinations}.

\begin{figure}[!t]
    \centering
    \begin{subfigure}[b]{0.4\textwidth}
        \centering
        \includegraphics[width=\textwidth]{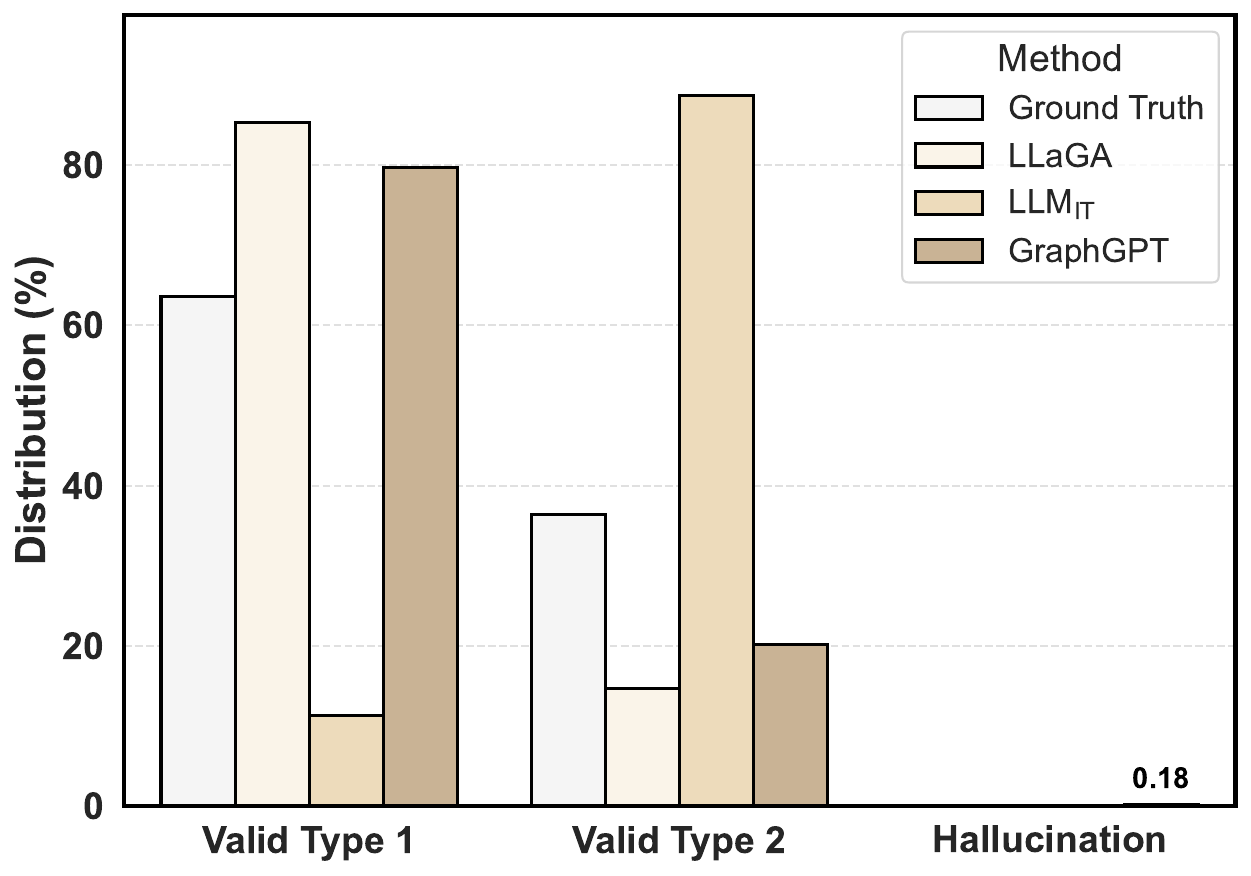}
        \caption{Different LLM-as-Predictor Methods}
        \label{fig:instagram_predictor}
    \end{subfigure}
    \begin{subfigure}[b]{0.4\textwidth}
        \centering
        \includegraphics[width=\textwidth]{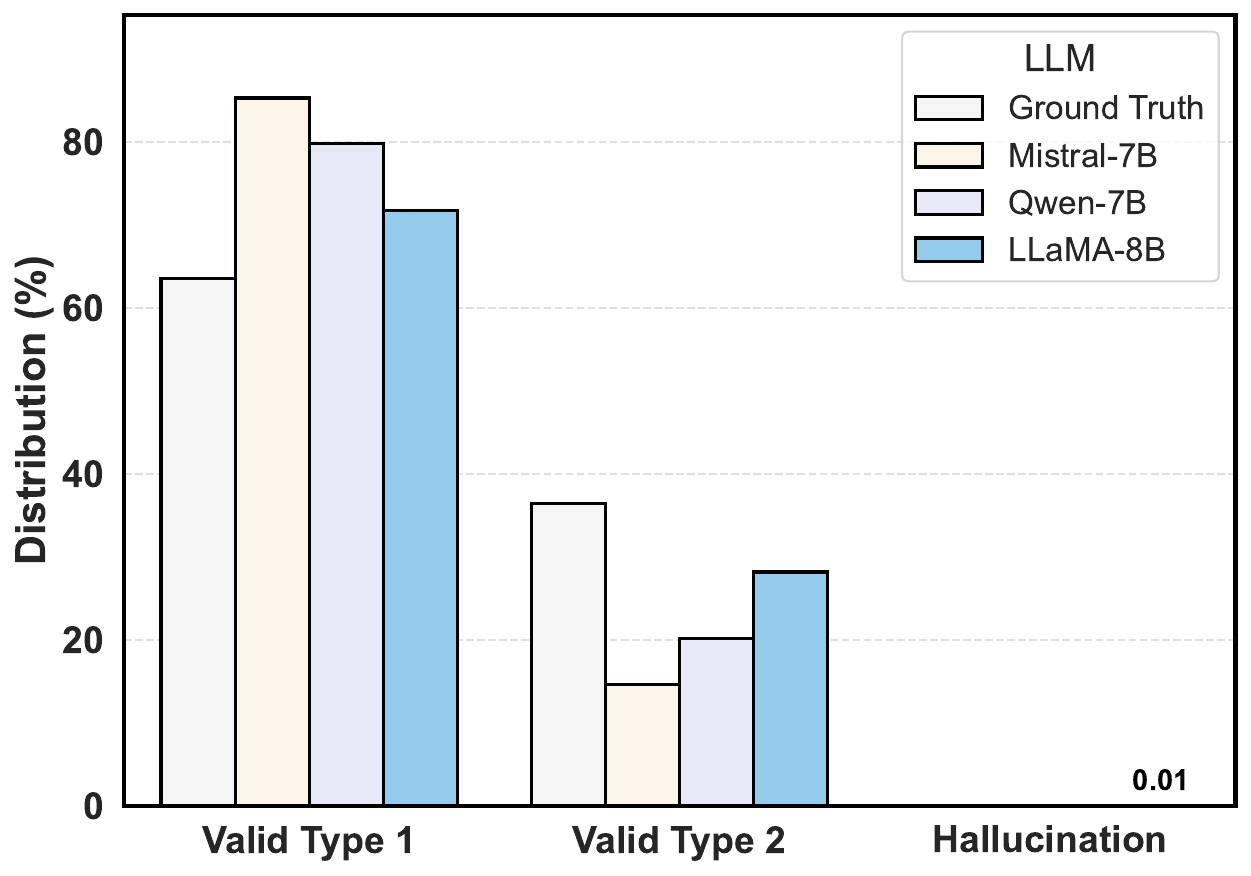}
        \caption{Different LLM Backbones within LLaGA}
        \label{fig:instagram_llm}
    \end{subfigure}
    \caption{\textbf{Biased predictions by LLM-as-Predictor methods on the Instagram dataset:} Comparison of ground-truth label distributions with predictor-generated label distributions.}
    \label{fig:distribution_instagram}
\end{figure}

\textbf{Biased Predictions:} For datasets with fewer labels, LLM-as-Predictor methods tend to disproportionately predict certain labels while under-predicting others. To illustrate this phenomenon, we compare the ground-truth label distributions with the predicted label distributions. Specifically, we present different LLM-as-Predictor methods, LLM\textsubscript{IT}, GraphGPT, and LLaGA, in Figure \ref{fig:instagram_predictor}, and the LLaGA method with various LLM backbones in Figure \ref{fig:instagram_llm}, using the Instagram dataset in semi-supervised settings, which has two labels.

\begin{figure}[!t]
    \centering
    \begin{subfigure}[b]{0.38\textwidth}
        \centering
        \includegraphics[width=\textwidth]{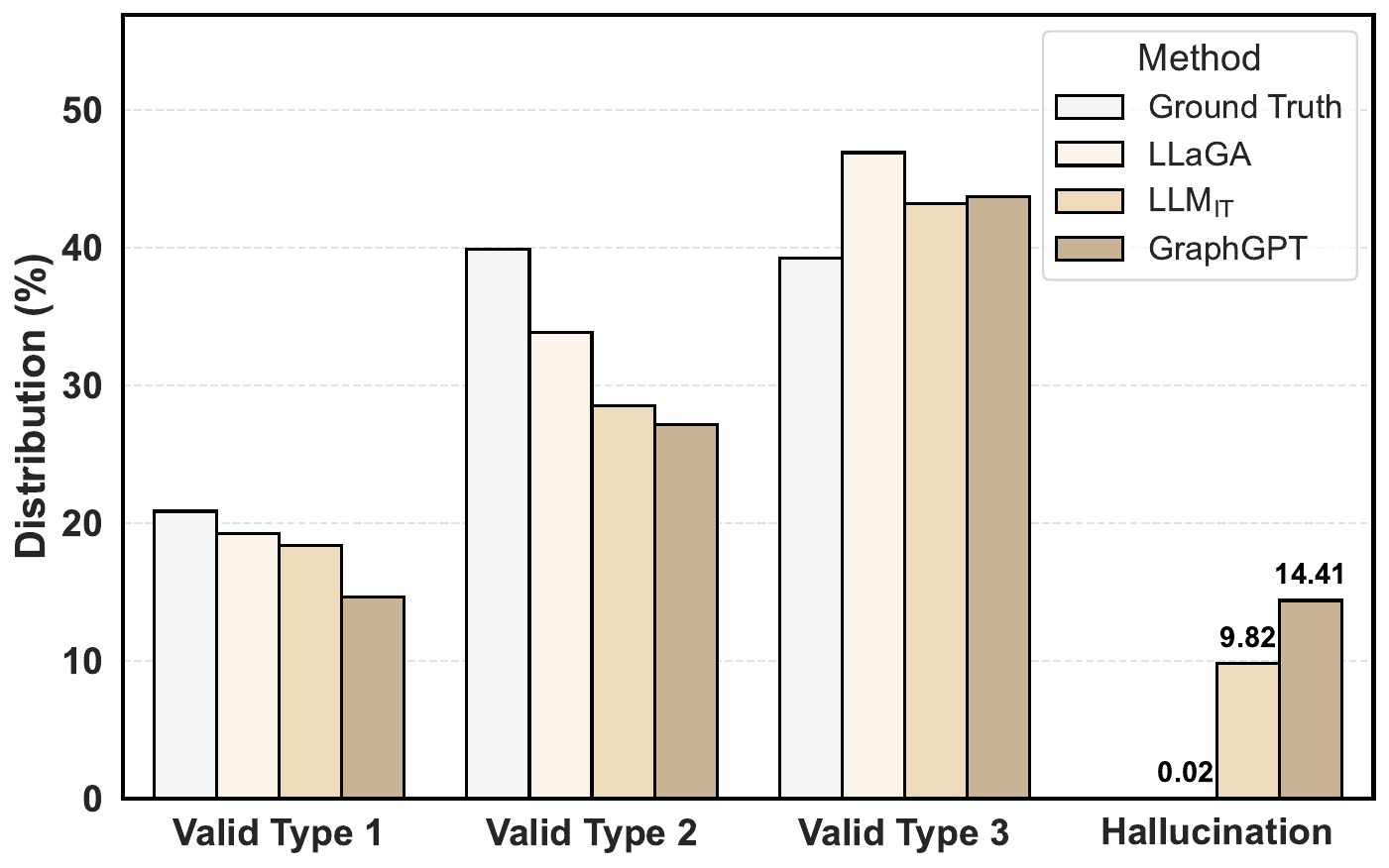}
        \caption{Different LLM-as-Predictor Methods}
        \label{fig:pubmed_predictor}
    \end{subfigure}
    \begin{subfigure}[b]{0.38\textwidth}
        \centering
        \includegraphics[width=\textwidth]{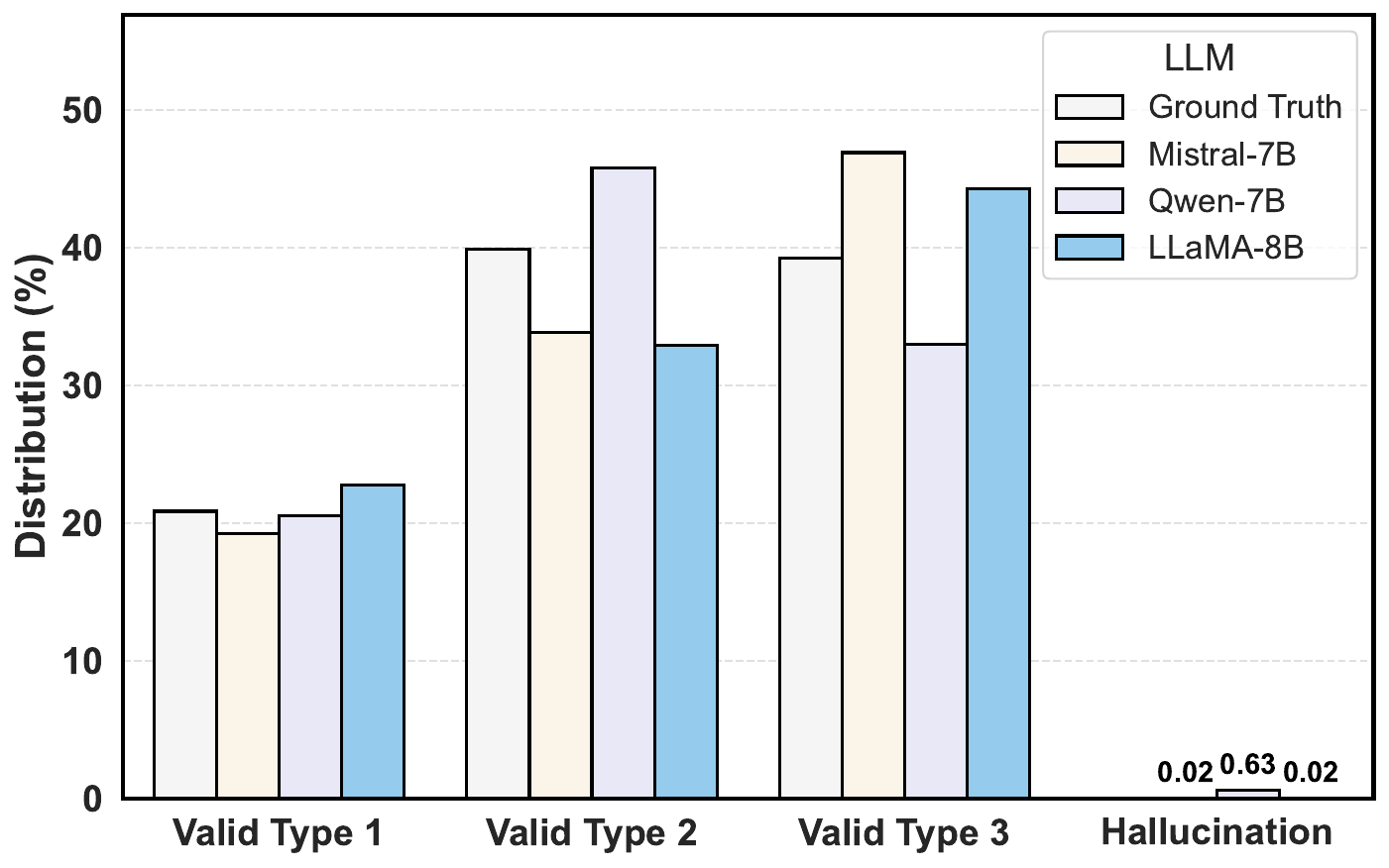}
        \caption{Different LLM Backbones within LLaGA}
        \label{fig:pubmed_llm}
    \end{subfigure}
    \caption{\textbf{Biased predictions by LLM-as-Predictor methods on the Pubmed dataset.}}
    \label{fig:distribution_pubmed}
\end{figure}

From Figure \ref{fig:instagram_predictor}, we can directly observe that LLaGA and GraphGPT predominantly bias towards the first class, while LLM\textsubscript{IT} tends to predict the second class more frequently. The predicted label distributions of LLMs are more \textbf{polarized} compared to the ground-truth distributions, where the two labels are roughly in a $6:4$ proportion. In contrast, LLMs tend to predict in ratios such as $8:2$ or $1:9$. This observation also holds across different LLMs, as shown in Figure \ref{fig:instagram_llm}, where both Qwen-7B and LLaMA-8B tend to bias towards the first label. A similar example on the Pubmed dataset, which contains three classes in a semi-supervised setting, is shown in Figure \ref{fig:distribution_pubmed}. Here, the predictor methods tend to bias towards the third class, while LLaGA with Qwen-7B tends to predict the second class. Additionally, in the semi-supervised setting for Pubmed, the training data consists of only $60$ samples, which is insufficient to train a robust predictor model, leading to high levels of hallucinations across all methods.

\textbf{Hallucinations: }In addition to biased predictions, we observed that a certain portion of the LLMs' outputs \textbf{fall outside the valid label space} or \textbf{contain unexpected content that cannot be parsed}. In semi-supervised settings, the limited training data restricts these predictor methods from developing effective models, resulting in failures to follow instructions and understand dataset-specific classification requirements. To illustrate that, we provide both quantitative and qualitative analyses as follows:

\begin{itemize}
    \item \textbf{Quantitative Analysis: }Table \ref{tab:predictor_hall} presents the hallucination rates of each LLM-as-Predictor method across various experimental datasets in both semi-supervised and supervised settings. The hallucination rate is calculated as the proportion of outputs containing invalid labels or unexpected content among all test cases, where higher values indicate poorer classification performance. Hallucinations are most severe on the Pubmed and Citeseer datasets within semi-supervised settings, where the number of training samples does not exceed hundreds, making effective model training challenging. \textbf{This demonstrates that the number of training samples significantly impacts the mitigation of hallucinations}: even in semi-supervised settings, larger datasets like Books and Photo provide thousands of training samples, resulting in hallucination ratios consistently below $1\%$. Therefore, this further verifies that LLM-as-Predictor methods require extensive labeled data for effective model training.
    
    \item \textbf{Qualitative Analysis: }We provide several examples to facilitate the comprehension of hallucinated predictions, which we categorize into three types: (1) misspellings of existing labels, (2) generation of non-existent types, and (3) unexpected content that cannot be parsed. Illustrative examples of these types from GraphGPT's outputs on the Citeseer and Pubmed datasets are presented in Table \ref{tab:predictor_hall_example}.
    
\end{itemize}

\begin{table}[!t]
    \centering
    \caption{\textbf{Average hallucination ratios ($\%$) of LLM-as-Predictor methods.} The hallucination rate is calculated as the proportion of outputs containing invalid labels or unexpected content across all test cases, where higher values indicate poorer classification ability. Hallucinations $>1\%$ are \textcolor{brown}{\textbf{highlighted}}.}
    \vspace*{-8pt}
    \resizebox{0.9\linewidth}{!}{
      \begin{tabular}{cc|ccccccccc}
        \toprule
        \rowcolor{COLOR_MEAN} \textbf{Setting} &  \textbf{Method} & \textbf{Cora} & \textbf{Citeseer} & \textbf{Pubmed} & \textbf{WikiCS} & \textbf{Instagram} & \textbf{Reddit} & \textbf{Books} & \textbf{Photo} & \textbf{Computer}  \\ \midrule
       \multirow{4}{*}{\textbf{Semi-supervised}} &  \# Train Samples & 140 & 120 & 60 & 580 & 1,160 & 3,344 & 4,155 & 4,836 & 8,722 \\ 
       & {LLM\textsubscript{IT}} & 0.43 & \textcolor{brown}{\textbf{13.08}} & \textcolor{brown}{\textbf{9.24}} & 0.06 & 0.00 & 0.00 & 0.01 & 0.02 & 0.02  \\ 
       & {GraphGPT} & \textcolor{brown}{\textbf{7.56}} & \textcolor{brown}{\textbf{2.51}} & \textcolor{brown}{\textbf{15.97}} & \textcolor{brown}{\textbf{7.76}} & 0.28 & \textcolor{brown}{\textbf{1.78}} & 0.51 & 0.72 & 0.27 \\
       &  LLaGA & 0.35 & 0.20 & 0.29 & 0.00 & 0.01 & 0.02 & 0.00 & 0.00 & 0.00 \\  \midrule

       \multirow{4}{*}{\textbf{Supervised}} & \# Train Samples & 1,624 & 1,911 & 11,830 &  7,020 & 6,803 & 20,060 & 24,930 & 29,017 & 52,337 \\ 
       & LLM\textsubscript{IT} & 0.06 & \textcolor{brown}{\textbf{13.61}} & 0.00 & 0.00 & 0.00 & 0.00 & 0.00 & 0.01 & 0.02 \\ 
       & GraphGPT & \textcolor{brown}{\textbf{1.29}} & 0.63 & 0.08 & \textcolor{brown}{\textbf{1.64}} & 0.13 & 0.50 & 0.11 & 0.11 & 0.10 \\ 
       & LLaGA & 0.03 & 0.00 & 0.00 & 0.00 & 0.00 & 0.01 & 0.00 & 0.00 & 0.00 \\
       
         \bottomrule
    \end{tabular}
    }
    \label{tab:predictor_hall}
\end{table}

\begin{table}[!t]
    \centering
    \caption{\textbf{Examples of hallucinations in GraphGPT's outputs on the Citeseer and Pubmed datasets.}}
     \vspace*{-8pt}
    \resizebox{\linewidth}{!}{
    \begin{tabular}{c|cc|cc}
      \toprule
      \rowcolor{COLOR_MEAN}  &  \multicolumn{2}{c|}{\textbf{Citeseer}} & \multicolumn{2}{c}{\textbf{Pubmed}} \\ 
        \rowcolor{COLOR_MEAN} \multirow{-2}{*}{\textbf{Error Type}} & \textbf{Prediction} & \textbf{Ground-truth} &  \textbf{Prediction} & \textbf{Ground-truth} \\ \midrule
        \textbf{Misspelling} & AGents & Agents & Type II diabetes & Type 2 diabetes \\ \midrule

        \multirow{4}{*}{\begin{tabular}{c}
             \textbf{Non-existent} \\ \textbf{Types}
        \end{tabular}} & Logic and Mathematics & ML (Machine Learning) & Type 3 diabetes of the young (MODY) & Type 2 diabetes \\
        & Information Extraction &	IR (Information Retrieval) & Genetic Studies of Wolfram Syndrome & Type 2 diabetes\\ 
        & Pattern Recognition & ML (Machine Learning) & Experimentally induced insulin resistance & Type 2 diabetes \\ 
        & Multiagent Systems & Agents & Experimentally induced oxidative stress & Experimentally induced diabetes \\ \midrule

        \multirow{2}{*}{\begin{tabular}{c}
             \textbf{Unexpected} \\ \textbf{Contents}
        \end{tabular}} & \multicolumn{2}{c|}{H.4.1 Office Automation: Workflow Management} & \multicolumn{2}{c}{membrane is not altered by diabetes.} \\ 

        & \multicolumn{2}{c|}{The citation graph is given by the following: ...} & \multicolumn{2}{c}{What is the sensitivity and specificity of the IgA-EMA test ...} \\

        \bottomrule
    \end{tabular}
    }
    \label{tab:predictor_hall_example}
\end{table}

\clearpage 
\newpage

\section{Supplementary Materials for Computational Cost Analysis}\label{sec:detail_cost}

\begin{table}[!h]
    \centering
    \caption{\textbf{Total training times of different methods in supervised settings.} All recorded experiment times are based on a single NVIDIA H100-80G GPU.}
    \resizebox{\linewidth}{!}{

      \begin{tabular}{cc|cccccccccc}
       \toprule
       \rowcolor{COLOR_MEAN}  \textbf{Type} & \textbf{Method} & \textbf{Cora} & \textbf{Citeseer} & \textbf{Pubmed} & \textbf{WikiCS} & \textbf{arXiv} & \textbf{Instagram} & \textbf{Reddit} & \textbf{Books} & \textbf{Photo} & \textbf{Computer} \\ \midrule
       \multicolumn{2}{c|}{\# Training Samples} & 1,624 & 1,911 & 11,830 & 7,020 &  90,941  & 6,803 & 20,060 & 24,930 & 29,017 & 52,337  \\ \midrule
       \multirow{5}{*}{\textbf{Classic}} & {GCN$_{\text{ShallowEmb}}$} & 1.8s & 1.7s & 5.2s & 5.1s & 51.2s & 19.5s & 8.5s & 14.9s & 19.7s & 25.8s \\ 
       & {GAT$_{\text{ShallowEmb}}$} & 2.1s & 1.9s & 7.9s & 5.7s & 1.5m & 2.7s & 6.9s & 16.6s & 28.0s & 44.6s \\ 
       & {SAGE$_{\text{ShallowEmb}}$} & 1.7s & 3.0s & 7.6s & 4.0s & 1.3m & 2.0s & 7.2s & 19.6s & 20.1s & 43.2s \\ 
       & {SenBERT-66M} &  35s & 41s & 2.6m & 2.5m & 7.4m & 1.2m & 4.4m & 1.8m & 2.2m & 4.1m \\ 
       & {RoBERTa-355M} & 1.3m & 1.6m & 9.2m & 5.5m & 40.8m & 5.3m & 15.9m & 9.7m & 11.9m & 22.4m \\ \midrule 

      \multirow{2}{*}{\textbf{Encoder}} & GCN$_{\text{LLMEmb}}$ & 1.2m & 1.4m & 13.4m & 7.5m & 1.4h & 4.5m & 16.1m & 23.6m & 26.8m & 44.8m  \\ 
      & ENGINE & 2.2m & 2.4m & 16.1m & 19.4m & 2.6h & 8.9m & 24.2m & 35.2m & 44.2m & 1.2h \\ \midrule

      \textbf{Explainer} & TAPE & 27.4m & 30.3m & 5.9h & 2.8h & 37.4h &  2.1h & 8.3h & 10.0h & 12.0h & 15.0h \\ \midrule

      \multirow{3}{*}{\textbf{Predictor}} & {LLM$_{\text{IT}}$} & 1.0h & 1.3h & 9.9h & 4.2h & 36.3h & 2.7h & 3.4h & 5.7h & 7.4h & 12.4h \\ 
      & GraphGPT & 26.4m & 29.5m & 2.7h & 1.7h & 7.8h & 49.1m & 3.4h & 3.8h & 3.6h & 7.8h \\ 
       & LLaGA &  5.6m & 7.7m & 25.6m & 18.8m & 7.7h & 10.6m & 32.2m & 1.0h & 1.4h & 2.5h \\    
        \bottomrule
      \end{tabular}
    }
    \label{tab:timecost_supervised}
\end{table}

\begin{table}[!h]
    \centering
    \caption{\textbf{Inference times of different methods.} Values in brackets denote the average inference time per case in milliseconds (ms). All recorded experiment times are based on a single NVIDIA H100-80G GPU.}
    \resizebox{0.8\linewidth}{!}{
     \begin{tabular}{cc|ccccc}
     \toprule
        \rowcolor{COLOR_MEAN} \multicolumn{2}{c|}{\textbf{Method}} & \textbf{Cora} & \textbf{arXiv}  & \textbf{Instagram} & \textbf{Photo} & \textbf{WikiCS} \\ \midrule
        \multicolumn{2}{c|}{\# Test Samples} &  542 & 48,603 & 5,847 & 2,268 & 9,673 \\ \midrule
        \textbf{Classic} & GCN & 0.9ms & 21.8ms & 2.0ms & 7.5ms & 4.4ms \\ \midrule 
        \textbf{Encoder} & GCN$_{\text{LLMEmb}}$ & 14.0s (26ms) & 23.8m (29ms) & 53.6s (24ms) & 5.3m (33ms) & 3.7m (38ms) \\  \midrule
        \textbf{Explainer} & TAPE & 5.0m (551ms) & 10.4h (767ms) & 23.7m (627ms) & 2.3h (863ms) & 1.3h (813ms) \\ \midrule
        \multirow{3}{*}{\textbf{Predictor}} & LLM$_{\text{IT}}$ & 1.2m (129ms) & 3.3h (243ms) & 2.7m (71ms) & 24.1m (149ms) & 5.8m (60ms) \\ 
        & GraphGPT & 1m (104ms) & 1.2h (87ms) & 2.0m (52ms) & 10.4m (64ms) & 11.0m (112ms) \\ 
        & LLaGA & 11.2s (21ms) & 57.1m (70ms) & 1.3m (35ms) & 4.4m (27ms) & 2.4m (25ms) \\
        \bottomrule
    \end{tabular}
    }
    \label{tab:inference_cost}
\end{table}

\begin{table}[!h]
    \centering
    \caption{\textbf{Memory costs (in GB) of different methods during training and inference stages.}}
    \resizebox{0.65\linewidth}{!}{

      \begin{tabular}{cc|ccc|ccc}
       \toprule
       \rowcolor{COLOR_MEAN} & & \multicolumn{3}{c|}{\textbf{Training}}  & \multicolumn{3}{c}{\textbf{Inference}}   \\
       \rowcolor{COLOR_MEAN} \multirow{-2}{*}{\textbf{Type}} & \multirow{-2}{*}{\textbf{Method}} & Cora & WikiCS & arXiv &  Cora & WikiCS & arXiv \\ \midrule 
       \multirow{3}{*}{\textbf{Classic}} &GCN$_{\text{ShallowEmb}}$ & 0.76 & 1.43 & 4.79 & 0.37 & 0.84 & 3.32 \\ 
       & SenBERT-66M & 9.70 & 9.70 & 9.98 & 3.25 & 3.26 & 3.38 \\ 
       & RoBERTa-355M &  46.00 & 46.00 & 46.10 & 7.44 & 7.44 & 7.55 \\ \midrule
      \multirow{2}{*}{\textbf{Encoder}} & GCN$_{\text{LLMEmb}}$ & 0.85 & 1.62 & 7.36 & 0.40 & 1.02 & 5.79 \\ 
      & ENGINE & 1.10 & 4.52 & 7.08 & 1.29 & 1.81 & 2.12 \\ \midrule
      \textbf{Explainer} & TAPE & 46.00 & 46.00 & 46.10 & 7.44 & 7.44 & 8.88 \\  \midrule 
      \multirow{3}{*}{\textbf{Predictor}} & LLM$_{\text{IT}}$ & 72.12 & 67.74 & 69.13 & 32.08 & 31.53 & 33.12  \\ 
      & GraphGPT & 41.42 & 53.10 & 60.41  & 36.02 & 36.07 & 36.97  \\ 
      & LLaGA & 34.60 & 35.54 & 41.94 & 20.45 & 20.94 & 29.00 \\ 
        
       \bottomrule
      \end{tabular}
    
    }
    \label{tab:memory_usage}
\end{table}

\end{document}